\documentclass{article} 
\usepackage{iclr2027_conference,times}

\usepackage{amsmath,amsfonts,bm}

\def\eqref#1{equation~\ref{#1}}

\def\1{\bm{1}}

\DeclareMathAlphabet{\mathsfit}{\encodingdefault}{\sfdefault}{m}{sl}
\SetMathAlphabet{\mathsfit}{bold}{\encodingdefault}{\sfdefault}{bx}{n}

\usepackage{hyperref}
\usepackage{url}

\usepackage{microtype}
\usepackage{booktabs} 
\usepackage{graphicx}
\usepackage{multirow}
\usepackage{subcaption}
\usepackage{wrapfig}
\usepackage{xcolor}
\usepackage{placeins}
\usepackage{tikz}
\usetikzlibrary{arrows.meta,calc}

\title{
What Makes Position Zero Special? \\
A Mechanistic Study of \\
Position Zero Attention Sinks in LLMs
}

\author{%
    \textbf{Runyu Peng}${}^{1}$\thanks{Partial works are done at Shanghai AI Laboratory. Works are done at NYU. \quad ${}^{\dag}$ Equal advising.}, 
    \textbf{Ruixiao Li}${}^{2}$, 
    \textbf{Mingshu Chen}${}^{2}$, 
    \textbf{Yunhua Zhou}${}^{3}$, \\ 
    \textbf{Qipeng Guo}${}^{3}$,
    \textbf{Xipeng Qiu}${}^{2}$,
    \textbf{Yucheng Lu}${}^{1,4 \dag}$,
    \textbf{Chen Zhao}${}^{1,4 \dag}$\\
    \\
    \textsuperscript{1}New York University \\
    \textsuperscript{2}Fudan University \\
    \textsuperscript{3}Shanghai AI Laboratory \\
    \textsuperscript{4}NYU Shanghai \\
    Correspondence to: \texttt{\{rp4671, cz1285, lu.yucheng\}@nyu.edu}
}%

\iclrfinalcopy 
\begin{document}

\maketitle

\begin{abstract}
Transformers frequently allocate disproportionate attention to specific tokens, a phenomenon known as \textbf{attention sinks}. Causal large language models reliably form one at \textbf{position zero}, though its role remains debated. We approach this question from a mechanistic perspective, tracing how the position-zero sink arises from the model's internal computation. 
We identify a two-block subnetwork responsible for this behavior, which we term the \textbf{P0-Sink Circuit}, and show it arises purely from the structural properties of causal attention, requiring no semantic content.
We further validate through from-scratch pre-training experiments that two proposed parameter-free methods effectively accelerate P0 sink formation, and find that earlier sink formation benefits pre-training and improves downstream performance. Both methods outperform the Transformer baseline and achieve performance comparable to Gated Attention across comprehensive settings. Code is available now at https://github.com/Pryest/flash-linear-attention.
\end{abstract}

\section{Introduction}
Large-scale auto-regressive language models tend to allocate disproportionately large attention to a small subset of tokens, a phenomenon known as the \textbf{attention sink}. While sinks can emerge at any position, those away from the start of the input are generally harmful~\citep{zhang2025attentionsinkscatchtag,yu2024unveilingharnessinghiddenattention}: excessive focus on semantically vacuous tokens such as punctuation interferes with reasoning, and suppressing them has been shown to improve downstream performance. The \textbf{position-zero (P0) sink}, however, correlates with improved predictions and underlies several downstream applications~\citep{xiao2024efficientstreaminglanguagemodels,han-etal-2024-lm,chen2024magicpiglshsamplingefficient}. Nevertheless, recent work questions whether the P0 sink is strictly necessary, proposing architectural alternatives such as Gated Attention~\citep{qiu2025gatedattentionlargelanguage} that claim improved performance without it.

To assess the P0 sink's role, we first investigate its underlying mechanism. Prior work suggests that removing \texttt{[BOS]} leaves the sink largely intact~\citep{gu2025attentionsinkemergeslanguage}, but this view is incomplete: removing \texttt{[BOS]} affects only the earliest-layer attention patterns, suggesting the sink arises from a more fundamental mechanism rooted in model architecture rather than token semantics.

From the perspective of mechanistic interpretability~\citep{elhage2021mathematical}, we reverse-engineer the Transformer and propose the \textbf{P0-Sink Circuit}, a two-block subnetwork that exploits the asymmetry of the causal attention mask. Under causal masking, position zero attends only to itself, producing an unmixed and directionally stable attention output. Subsequent MLP sublayers detect and amplify this signal into a high-norm, fixed-direction representation at position zero, which reliably triggers the P0 sink throughout the network. This explains the sink's consistent emergence across models and inputs.

Building on this mechanistic understanding, we propose two parameter-free pre-training methods that help models form the P0 sink more effectively. Our view is that the Transformer naturally favors routing attention through a consistent sink, and position zero is selected for this role precisely because it is always in context. Besides, the two functions of the P0-Sink Circuit, identifying position zero and suppressing semantic content, can be fulfilled by alternative mechanisms.

From the first perspective, we propose \textbf{Transformer-Native LM (TNLM) Loss}, which removes the cross-entropy term at the \texttt{[BOS]} position, eliminating the semantic component of the P0 representation and freeing it to act as a dedicated attention register, inspired by register theory in vision transformer research~\citep{darcet2024visiontransformersneedregisters,jiang2025visiontransformersdontneed}. From the second perspective, we propose \textbf{Force-sink Masking}, which applies a fixed binary mask to the residual stream at every layer, hard-partitioning the hidden dimensions between position zero and all other positions. This structural partition simultaneously realizes both functions of the P0-Sink Circuit without additional parameters, and alleviates the pressure that drives standard models toward the \textbf{outlier-driven rescaling} issue~\citep{qiu2026unifiedviewattentionresidual}.

In this work, we make the following contributions:
\begin{enumerate}
    \item We formalize the \textbf{P0-Sink Circuit}, a two-block subnetwork that exploits the asymmetry of the causal attention mask to generate a high-norm, fixed-direction representation at position zero, providing a consistent reference point for attention heads throughout the network.
    \item We propose two parameter-free pre-training methods that promote P0 sink formation from complementary mechanistic perspectives. 
    \item We conduct comprehensive experiments across diverse learning rate and data scale settings, demonstrating that both methods consistently improve over the Transformer baseline and achieve performance comparable to Gated Attention.
\end{enumerate}

\begin{figure}[t]
    \centering
    \includegraphics[width=0.8\linewidth]{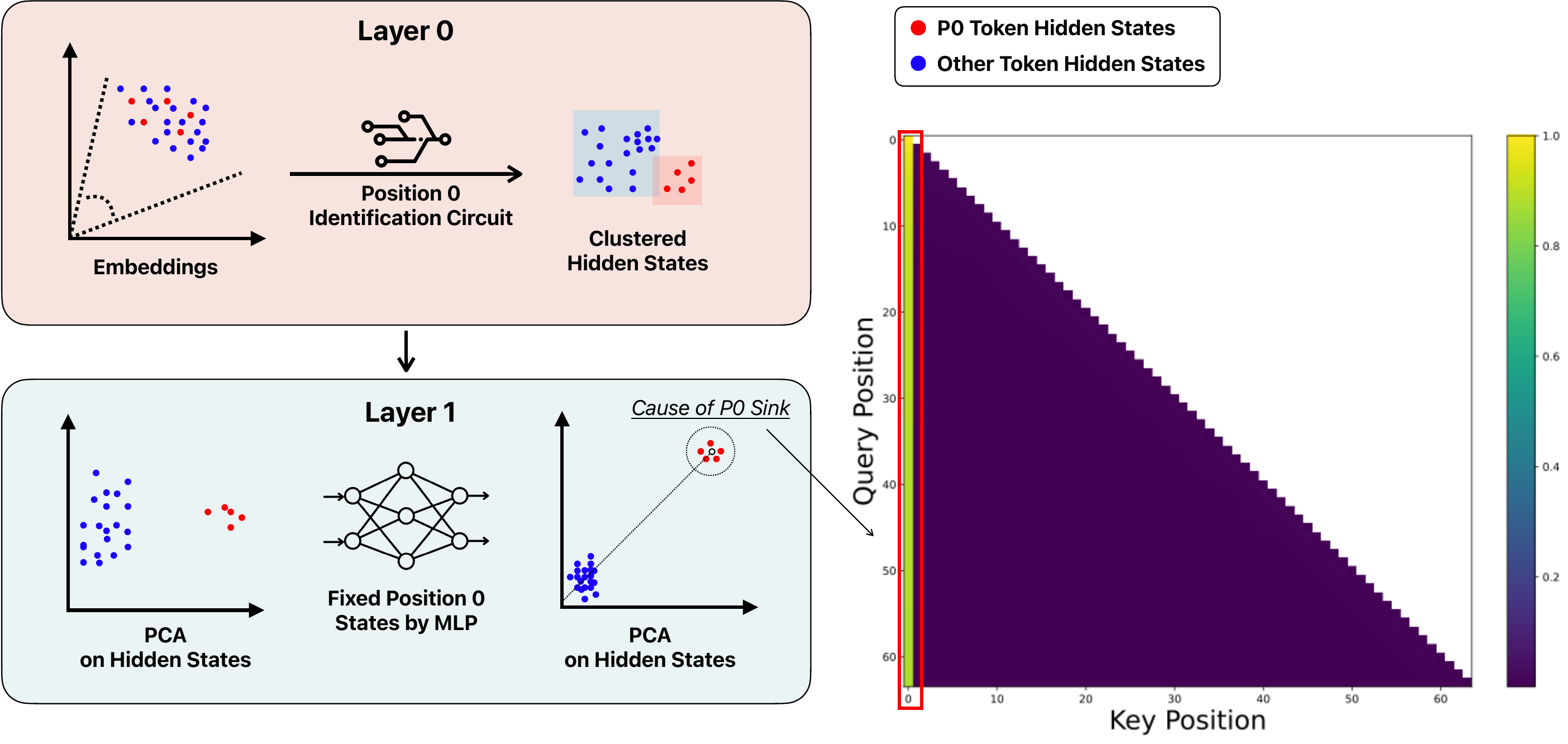}
    \caption{Overview of the proposed \textbf{P0-Sink Circuit}. Within just two transformer blocks, the model learns to identify the position-zero (P0) token and amplify it into a fixed high-norm representation, which gives rise to the attention sink effect.}
    \label{fig-P0SC}
    \vspace{-1em}
\end{figure}

\section{How \texttt{[BOS]} Token Affects P0 Sink}
\label{sec:nobos}
\citet{barbero2025llmsattendtoken} remove \texttt{[BOS]} and report a weakened sink effect, measured by the \textbf{Sink Rate} metric proposed by \citet{gu2025attentionsinkemergeslanguage}:
\begin{equation}
    \text{Sink}_{k}^{\epsilon} = \frac{1}{L} \sum_{l=0}^{L-1} \frac{1}{H}\sum_{h=1}^{H}\mathbb{I}(\alpha_{k}^{l,h} > \epsilon),
\end{equation}
where $\alpha_k^{l,h}$ denotes the post-softmax attention score assigned to position $k$ at layer $l$ and head $h$ of an $L$-layer, $H$-head Transformer, and $\epsilon$ is a predefined threshold. However, averaging over all layers obscures layer-wise differences: removing \texttt{[BOS]} may primarily affect early layers while leaving the sink intact elsewhere, a distinction the Sink Rate cannot capture.

\citet{gu2025attentionsinkemergeslanguage} show that the attention sink persists regardless of dataset choice, even on random sequences. However, when the input consists of repeated first tokens, the sink disappears in Mistral and LLaMA models\footnote{Mistral and LLaMA models use \texttt{[BOS]} as the first token of any sequence, while other LLMs, e.g., Qwen and Pythia, do not.}. They attribute this to identical hidden states produced by repeated first tokens under RoPE~\citep{su2023roformerenhancedtransformerrotary}. This raises a question: does the P0 sink depend on the learned embedding of \texttt{[BOS]}, or on its structural role as the token at position zero?

To answer this question, we examine LLMs that prepend a \texttt{[BOS]} token, namely LLaMA and Mistral, which differ in their training regimes. LLaMA uses full-context attention, so \texttt{[BOS]} is always visible as the first token throughout training. Mistral uses sliding window attention, causing the first token to fall out of scope after the first slide, making both the \texttt{[BOS]} hidden states invisible to later tokens.

This architectural difference has a functional consequence. Prior work shows that the P0 sink acts as a head-level gating mechanism, where attention is routed to a semantically vacuous position-zero representation to deactivate certain heads~\citep{sandovalsegura2026identifyingevaluatinginactiveheads}, and that this role can be fulfilled by alternative designs such as gated attention~\citep{qiu2026unifiedviewattentionresidual,qiu2025gatedattentionlargelanguage}. \textbf{Sliding window models like Mistral develop different strategies to fulfill the same functional role.} The same consideration extends to hybrid architectures that interleave sliding window and full attention layers: sliding window layers exhibit no P0 sink while full attention layers do, regardless of whether RoPE is employed~\citep{yang2025ropenopeagainnew}. We leave a detailed investigation of these cases for future work.

\begin{table*}[t]
\centering
\vspace{-1em}
\begin{subtable}[c]{0.48\linewidth}
\centering
\caption{$\text{Sink}_{0}^{\epsilon}[l_{start}:]$ with ablation on \texttt{[BOS]} ($\epsilon=0.3$), following \citet{gu2025attentionsinkemergeslanguage}.}
\label{sink-rate-rm-bos}
\resizebox{\linewidth}{!}{
\begin{tabular}{cl|cccc}
\toprule
\multicolumn{2}{l|}{\textbf{Model}}                                                                & $\text{Sink}_{0}^{\epsilon}[0:]$     & $\text{Sink}_{0}^{\epsilon}[1:]$     & $\text{Sink}_{0}^{\epsilon}[2:]$     & $\text{Sink}_{0}^{\epsilon}[3:]$     \\ \cline{1-6}
\multicolumn{1}{l|}{\multirow{2}{*}{\textbf{Llama3.1-8B}}}   & \multicolumn{1}{l|}{w/  \texttt{[BOS]}}  & 93.69 & 95.52 & 95.39 & 95.23 \\ 
\multicolumn{1}{l|}{}                               & \multicolumn{1}{l|}{w/o  \texttt{[BOS]}} & 89.40 & 92.28 & 95.35 & 95.19 \\ \cline{1-6} 
\multicolumn{1}{l|}{\multirow{2}{*}{\textbf{Llama3.2-1B}}}   & w/ \texttt{[BOS]}                       & 90.32 & 92.07 & 91.53 & 90.89 \\ 
\multicolumn{1}{l|}{}                               & w/o \texttt{[BOS]}                      & 79.49 & 84.79 & 90.85 & 90.15 \\ \cline{1-6} 
\multicolumn{1}{l|}{\multirow{2}{*}{\textbf{Llama3.2-3B}}}   & \multicolumn{1}{l|}{w/  \texttt{[BOS]}}  & 96.14 & 96.83 & 96.71 & 96.58 \\  
\multicolumn{1}{l|}{}                               & w/o \texttt{[BOS]}                      & 89.78 & 93.11 & 96.69 & 96.55 \\ \cline{1-6} 
\multicolumn{1}{l|}{\multirow{2}{*}{\textbf{Mistral-7B-v0.3}}} & w/ \texttt{[BOS]}                       & 87.45 & 88.27 & 87.91 & 87.50 \\ 
\multicolumn{1}{l|}{}                               & w/o \texttt{[BOS]}                      & 1.87  & 1.93  & 1.92  & 1.90 \\
\bottomrule
\end{tabular}}
\end{subtable}%
\hfill
\begin{subtable}[c]{0.48\linewidth}
\centering
\caption{Cross-Entropy loss under different first-token repeat counts. \textbf{Bold}: smaller loss.}
\label{bos-vs-non-bos-tab1}
\resizebox{\linewidth}{!}{
\begin{tabular}{l|l|cccccccc|c}
\toprule
\multicolumn{2}{l|}{\textbf{Repeat (n)}}        & n=1           & n=2           & n=3           & n=4           & n=5           & n=6           & n=7           & n=8   & $\Delta_{1\text{-}8}$        \\
\midrule
\multirow{2}{*}{\textbf{Llama-3.1-8B}} & w/ \texttt{[BOS]}  & \textbf{2.66} & \textbf{2.68} & \textbf{2.72} & \textbf{2.76} & \textbf{2.80} & \textbf{2.84} & \textbf{2.89} & 2.93   & 0.27       \\
                                                & w/o \texttt{[BOS]} & 2.79          & 2.81          & 2.83          & 2.85          & 2.86          & 2.88          & 2.89          & \textbf{2.90} & \textbf{0.21}\\
\midrule
\multirow{2}{*}{\textbf{Llama-3.2-1B}} & w/ \texttt{[BOS]}  & \textbf{3.22} & \textbf{3.27} & \textbf{3.36} & \textbf{3.46} & 3.56          & 3.68          & 3.80          & 3.92      & 0.70    \\
                                                & w/o \texttt{[BOS]} & 3.32          & 3.36          & 3.42          & 3.48          & \textbf{3.53} & \textbf{3.58} & \textbf{3.62}          & \textbf{3.66} & \textbf{0.34}          \\
\midrule
\multirow{2}{*}{\textbf{Llama-3.2-3B}} & w/ \texttt{[BOS]}  & \textbf{2.97} & \textbf{3.01} & \textbf{3.07} & \textbf{3.15} & 3.23          & 3.33          & 3.43          & 3.53     & 0.56     \\
                                                & w/o \texttt{[BOS]} & 3.06          & 3.10          & 3.15          & 3.19          & \textbf{3.22} & \textbf{3.26} & \textbf{3.29} & \textbf{3.32} & \textbf{0.26} \\
\midrule
\multirow{2}{*}{\textbf{Mistral-7B-v0.3}}   & w/ \texttt{[BOS]}  & 2.44          & 2.46          & 2.49          & 2.54          & 2.60          & 2.69          & 2.79          & 2.91     & 0.47     \\
                                                & w/o \texttt{[BOS]} & \textbf{2.39} & \textbf{2.39} & \textbf{2.40} & \textbf{2.41} & \textbf{2.41} & \textbf{2.42} & \textbf{2.43} & \textbf{2.44} & \textbf{0.05} \\
\bottomrule
\end{tabular}}
\end{subtable}
\caption{Ablation study on \texttt{[BOS]} token across different LLMs.}
\label{tab:bos-ablation-combined}
\vspace{-1em}
\end{table*}

Since LLMs without a \texttt{[BOS]} token also exhibit P0 sinks, there must exist at least one mechanism that gives rise to the P0 sink independently of the \texttt{[BOS]} embedding. For LLMs where \texttt{[BOS]} remains visible throughout training, the model re-establishes P0 sinks that do not rely on the \texttt{[BOS]} embedding within the shallowest layers. To demonstrate this, we introduce a modification to $\text{Sink}_{k}^{\epsilon}$ \citep{gu2025attentionsinkemergeslanguage}:
\begin{equation}
    \text{Sink}_{k}^{\epsilon}[l_{\text{start}}:] = \frac{1}{L - l_{\text{start}}} \sum_{l=l_{\text{start}}}^{L-1} \frac{1}{H}\sum_{h=1}^{H}\mathbb{I}(\alpha_{k}^{l,h} > \epsilon),
\end{equation}
where $l_{\text{start}} \in [0, L)$. This metric shows directly that removing \texttt{[BOS]} affects only the shallowest layers, as shown in Table~\ref{sink-rate-rm-bos}.

This raises a further question: between a P0 representation shaped purely by positional structure and one grounded in the \texttt{[BOS]} embedding, which better fulfills the vacuous role that the P0 sink requires? To answer this, we repeat the first token multiple times and measure the resulting change in loss. As shown in Table~\ref{bos-vs-non-bos-tab1}, experiments are conducted on a subset of the FineWeb-Edu 10B dataset\footnote{Unless specified, we use the first 1024 samples from the first \texttt{.parquet} file of FineWeb-Edu 10B dataset as validation set, with a maximum sequence length of 64. These samples are excluded from training.}. In the w/ \texttt{[BOS]} setting, the repeated token is \texttt{[BOS]}; in the w/o \texttt{[BOS]} setting, the first content token is repeated. Losses on repeated tokens and the original \texttt{[BOS]} position are excluded for fair comparison.

The results in Tables~\ref{sink-rate-rm-bos} and~\ref{bos-vs-non-bos-tab1} together support a clear conclusion. For Mistral, \texttt{[BOS]} is the sole driver of the sink: \textbf{removing it collapses the sink rate greatly and directly reduces loss}. This is consistent with our earlier discussion: sliding window attention causes \texttt{[BOS]} to fall out of scope, so Mistral never develops a positional P0 sink and relies entirely on the \texttt{[BOS]} embedding instead. For the LLaMA family, the picture is different. Removing \texttt{[BOS]} affects only the first two layers, as shown in Table~\ref{sink-rate-rm-bos}, where $\text{Sink}_{0}^{\epsilon}[2:]$ recovers to near the w/ \texttt{[BOS]} level. Correspondingly, repeating a non-\texttt{[BOS]} token yields a much smaller difference than repeating \texttt{[BOS]} itself, confirming that the positional P0 representation is a more stable semantic vacuum than the \texttt{[BOS]} embedding.

\section{How LLMs Sink the Position-Zero Token}
\label{sec:how_sink}

\subsection{Fixed Representation at Position Zero}
\label{subsec:amplified_l2}

Prior work has identified two key observations underlying the P0 sink. \citet{cancedda2024spectralfiltersdarksignals} show that the hidden state at the \texttt{[BOS]} position develops significantly amplified $\ell_2$ norms after passing through certain Transformer layers. \citet{gu2025attentionsinkemergeslanguage} further demonstrate that this norm inflation correlates with disproportionate concentration of attention scores at position zero. Consistent findings are reported by \citet{an2025systematic} and \citet{sun2024massiveactivationslargelanguage}, who observe similar outlier activation patterns from different analytical perspectives.

\begin{figure*}[t]
    \centering
    \begin{minipage}[b]{0.32\linewidth} 
        \centering
        \begin{subfigure}[b]{0.95\linewidth} 
            \centering
            \caption{\tiny{$\ell_2$ Norm of Activations, w/ \texttt{[BOS]}}}
            \includegraphics[width=0.95\columnwidth]{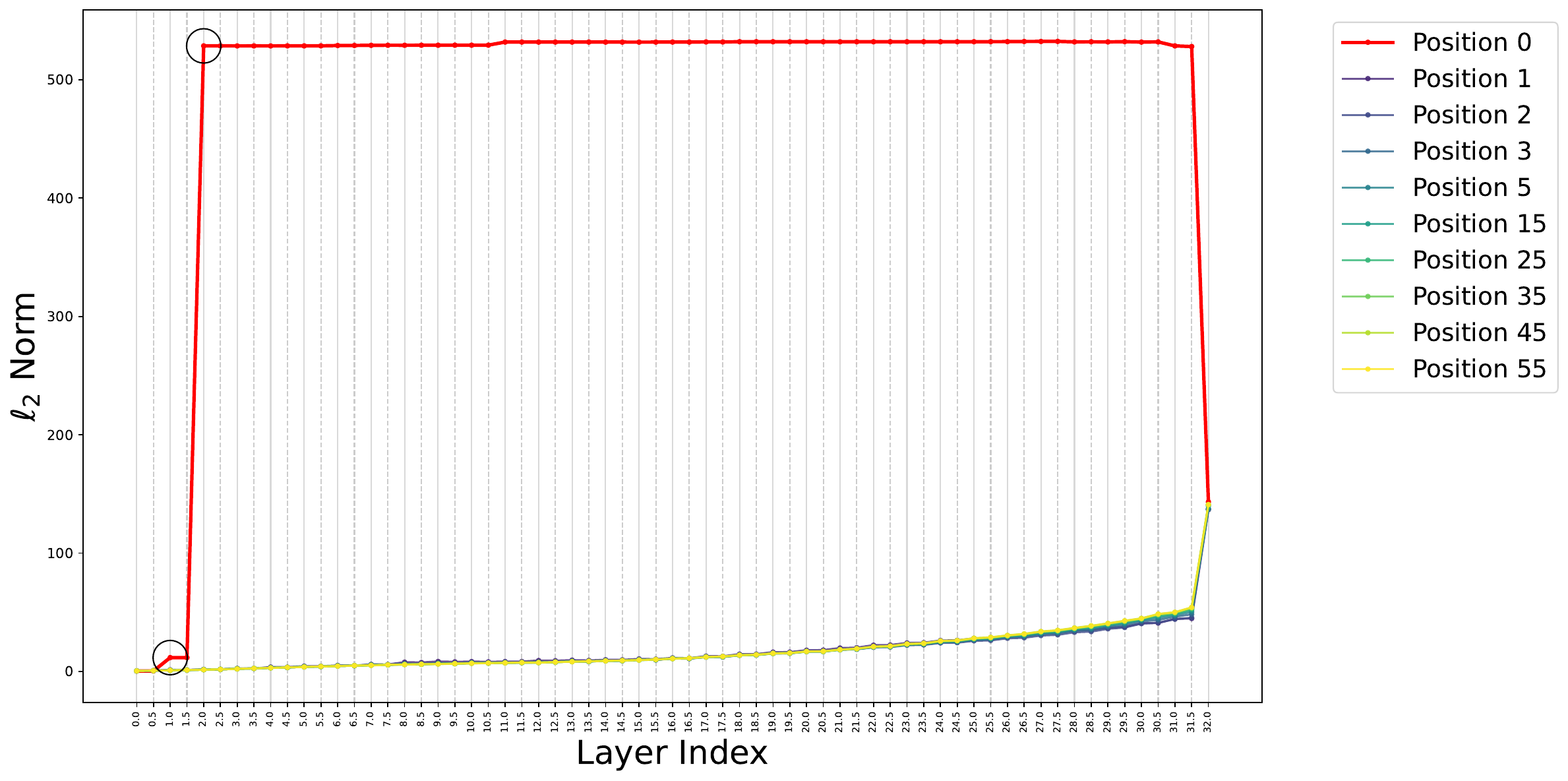}
        \end{subfigure}
    \end{minipage}
    \begin{minipage}[b]{0.32\linewidth} 
        \centering
        \begin{subfigure}[b]{0.95\linewidth} 
            \centering
            \caption{\tiny{$\ell_2$ Norm of Activations, w/o \texttt{[BOS]}}}
            \includegraphics[width=0.95\columnwidth]{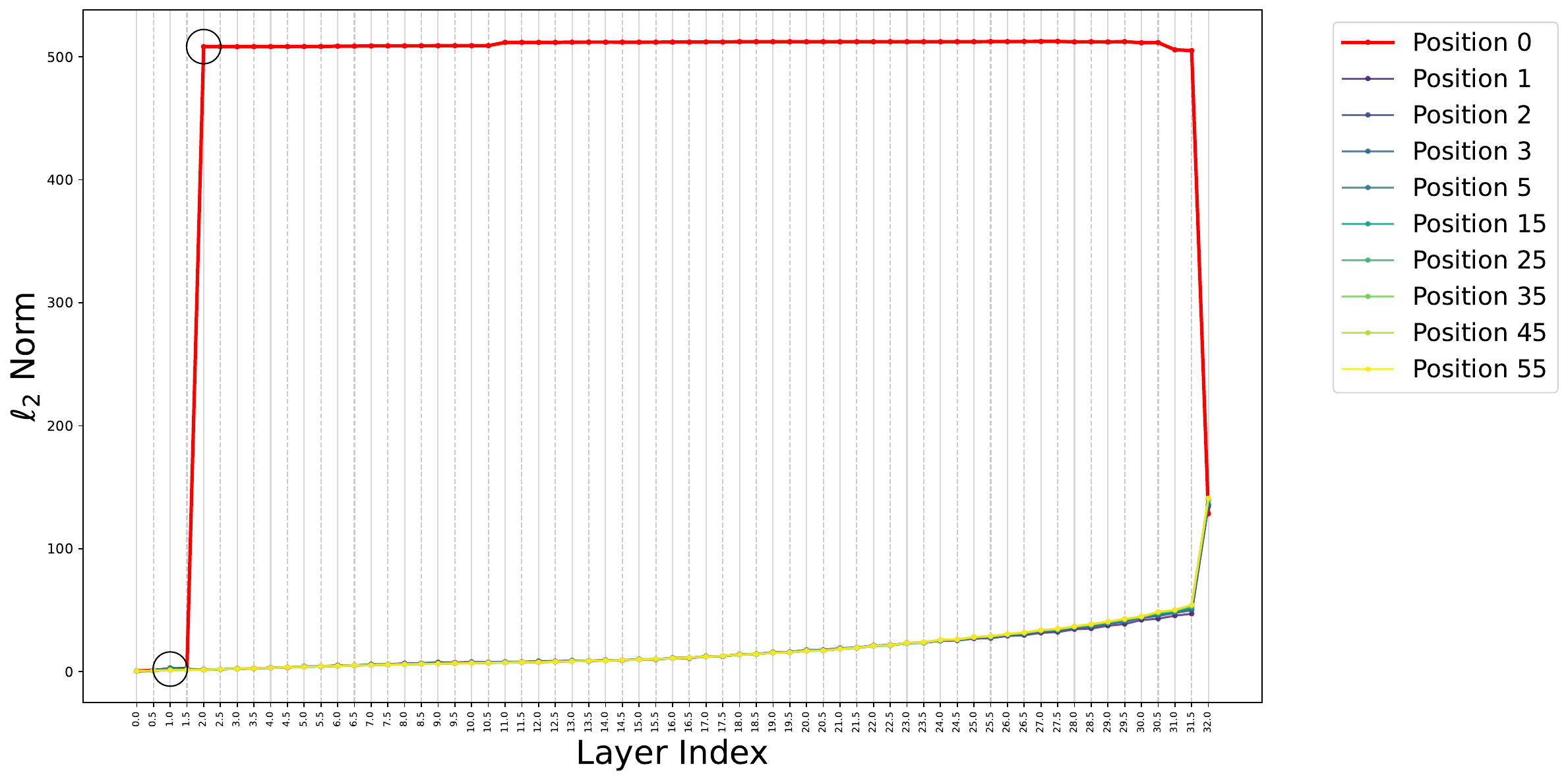}
        \end{subfigure}
    \end{minipage}
    \begin{minipage}[b]{0.32\linewidth} 
        \centering
        \begin{subfigure}[b]{0.95\linewidth} 
            \centering
            \caption{\tiny{Cosine Similarity of Activations, w/o \texttt{[BOS]}}}
            \label{cos-sim-fig}
            \includegraphics[width=0.95\columnwidth]{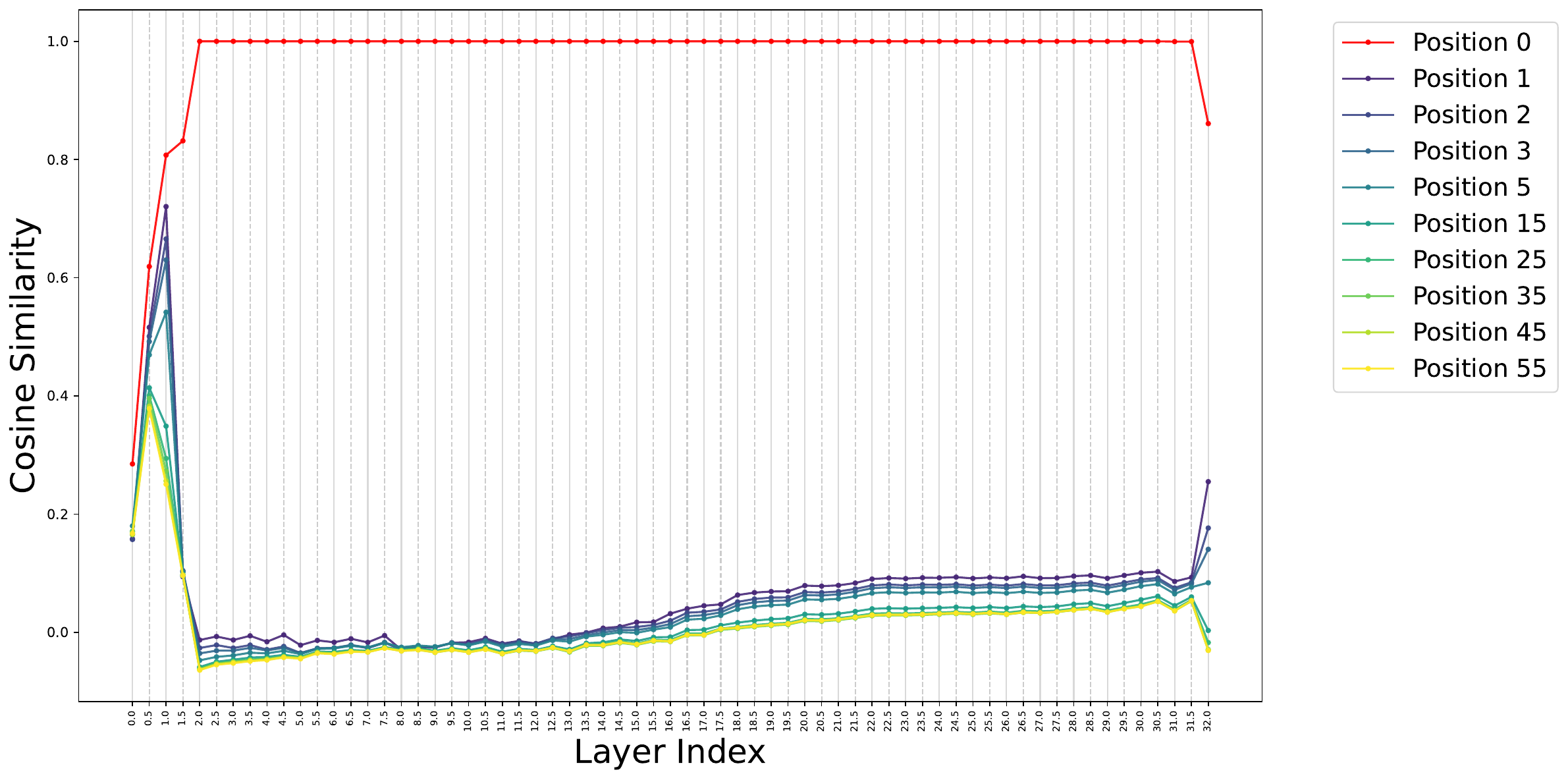}
        \end{subfigure}
    \end{minipage}
    \caption{Layer-wise $\ell_2$ norm and average cosine similarity of hidden states in LLaMA3.1-8B. Half-integer indices correspond to attention module outputs after the residual connection. Cosine similarity is computed per layer and position as the average cosine similarity between each sample activation and the mean activation across all samples. Both metrics are computed over sequences of up to 64 tokens across 1024 validation samples.}
    \label{bos-vs-non-bos-fig1}
    \vspace{-1em}
\end{figure*}

We build on these observations to characterize the mechanism more precisely. As shown in Table~\ref{sink-rate-rm-bos} and Figure~\ref{bos-vs-non-bos-fig1}, even after removing \texttt{[BOS]}, the model localizes the position-zero token and forms a P0 sink within just two Transformer blocks. This is not specific to \texttt{[BOS]}: \citet{li2026structuraloriginattentionsink} show that blocking any intermediate token from attending to prior positions is sufficient to induce an attention sink on it, and that artificially amplifying the variance of an arbitrary token's attention output produces the same effect, whereas merely scaling its representation norm does not. The common condition is therefore elevated variance in the attention output, not token identity or representation magnitude.

The MLP sublayer is central to this process. It not only inflates the $\ell_2$ norm of the P0 hidden state, but also projects it toward a consistent direction in representation space. Figure~\ref{bos-vs-non-bos-fig1} illustrates both effects. We quantify the directional convergence via cosine similarity in Figure~\ref{cos-sim-fig}: values near 1 confirm that, across varying inputs, the P0 representation converges to the same direction after just two Transformer blocks. Since pre-layer RMS normalization (pre-norm) removes magnitude information before each attention and MLP sublayer~\citep{zhang2019rootmeansquarelayer,ba2016layernormalization}, direction is precisely what downstream computation sees, making this convergence functionally significant. We omit the w/ \texttt{[BOS]} condition from Figure~\ref{cos-sim-fig}, as the repeated presence of the same token embedding at position zero makes directional convergence uninformative as a test of the positional mechanism.

This $\ell_2$ amplification serves two functional roles. First, it renders the P0 representation semantically vacuous. Because the MLP output is added to the residual stream, the amplified hidden state at P0 comes to dominate over early layer contributions. Second, it stabilizes the P0 representation across training. Under pre-norm, a high-magnitude vector is less sensitive to gradient updates and more likely to preserve its direction across optimization steps. This stability allows the model to maintain a consistent, distinguishable representation at position zero throughout training, one that attention heads across the network can reliably target.

To formalize the stability of the P0 representation under training, we analyze how pre-norm interacts with gradient updates. The pre-norm transformation applied to a hidden state $\mathbf{x} \in \mathbb{R}^d$ is:
\begin{equation}
    \text{Norm}(\mathbf{x}) = \frac{\mathbf{x}}{r(\mathbf{x})}, \quad r(\mathbf{x}) := \sqrt{\frac{1}{d} \mathbf{x}^\top \mathbf{x}}.
\end{equation}
Its Jacobian with respect to $\mathbf{x}$ is:
\begin{equation}
    \mathbf{J}_{\text{Norm}}(\mathbf{x}) = \frac{1}{r(\mathbf{x})} \mathbf{I} - \frac{1}{d} \frac{1}{r^3(\mathbf{x})} \mathbf{x} \mathbf{x}^\top.
\end{equation}
Consider a perturbation $\delta \mathbf{x} \ll \mathbf{x}$ induced by a gradient update during training, decomposed into components parallel and orthogonal to $\mathbf{x}$:
\begin{equation}
    \delta \mathbf{x} = \delta \mathbf{x}_\parallel + \delta \mathbf{x}_\perp, \quad \delta \mathbf{x}_\parallel = \alpha \mathbf{x}, \quad \delta \mathbf{x}_\perp \cdot \mathbf{x} = 0.
\end{equation}
The resulting change in the normalized vector is approximately:
\begin{equation}
    \text{Norm}(\mathbf{x} + \delta \mathbf{x}) - \text{Norm}(\mathbf{x}) \approx \mathbf{J}_{\text{Norm}}(\mathbf{x})\,\delta\mathbf{x} = \frac{\delta \mathbf{x}_\perp}{r(\mathbf{x})}.
    \label{eq:norm}
\end{equation}
The parallel component vanishes entirely, since rescaling $\mathbf{x}$ does not change its normalized direction. The orthogonal component, which does alter the direction, is attenuated by $1/r(\mathbf{x})$. Under the standard i.i.d. assumption on training data, gradient noise can be treated as approximately independent of the input~\citep{martens2020optimizingneuralnetworkskroneckerfactored}, so $\delta \mathbf{x}_\perp$ represents generic directional perturbation. A larger $r(\mathbf{x})$ therefore directly reduces the sensitivity of the normalized direction to gradient updates, providing a formal basis for the directional stability of high-norm representations observed at position zero.

\begin{figure*}[t]
    \centering
    \begin{minipage}[b]{0.32\linewidth} 
        \centering
        \begin{subfigure}[b]{0.95\linewidth} 
            \centering
            \includegraphics[width=\linewidth]{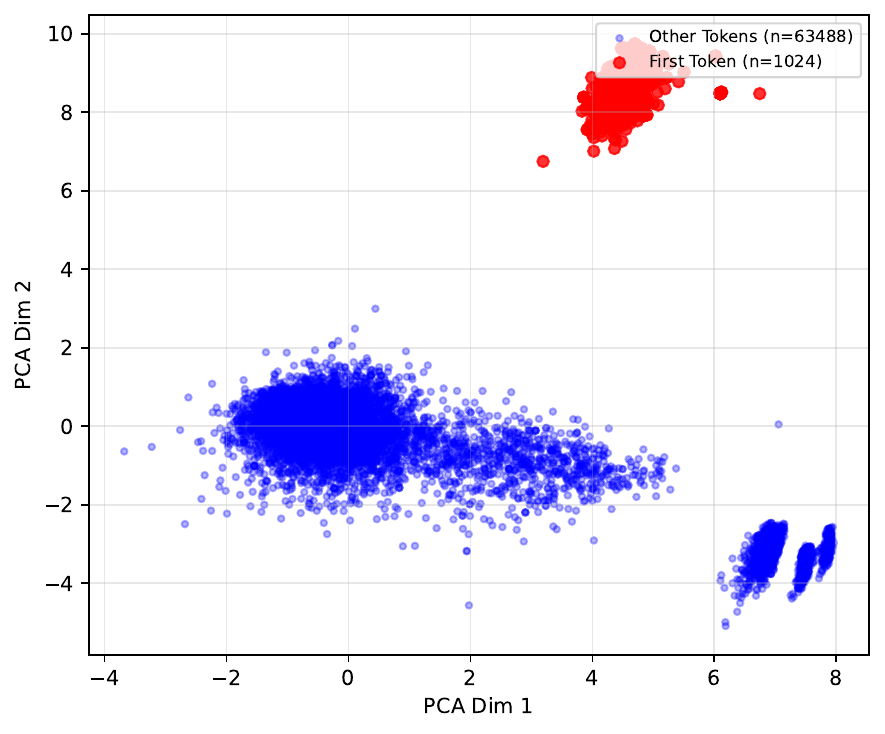}
            \caption{\scriptsize{PCA of Layer 1 MLP Input}}
        \end{subfigure}
    \end{minipage}
    \begin{minipage}[b]{0.32\linewidth} 
        \centering
        \begin{subfigure}[b]{0.95\linewidth} 
            \centering
            \includegraphics[width=\linewidth]{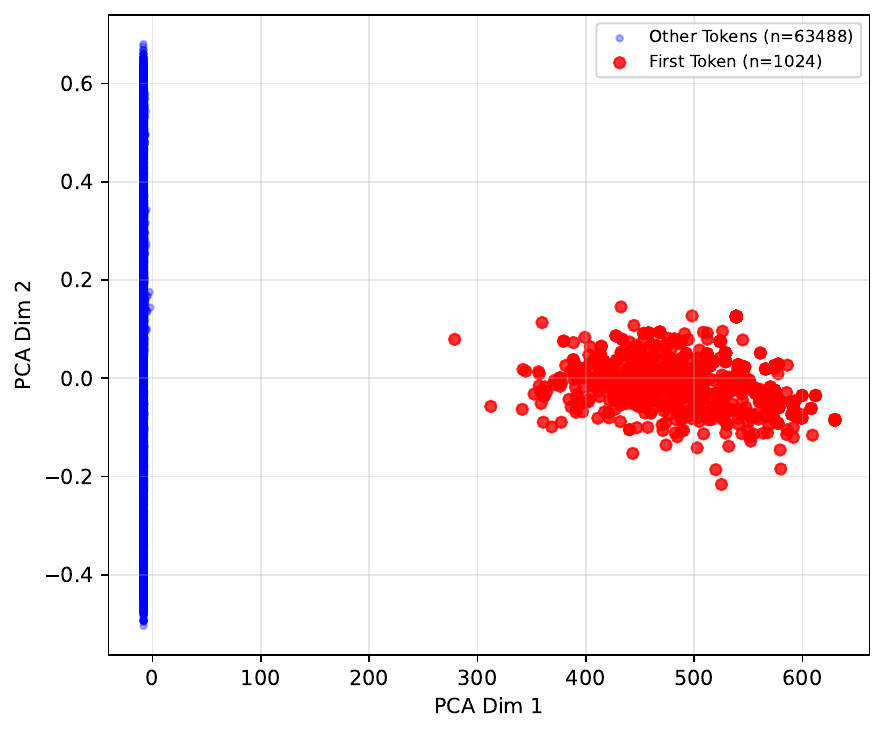}
            \caption{\scriptsize{PCA of Layer 2 Attn Input}} 
        \end{subfigure}
    \end{minipage}
    \begin{minipage}[b]{0.32\linewidth} 
        \centering
        \begin{subfigure}[b]{0.95\linewidth} 
            \centering
            \includegraphics[width=0.95\columnwidth]{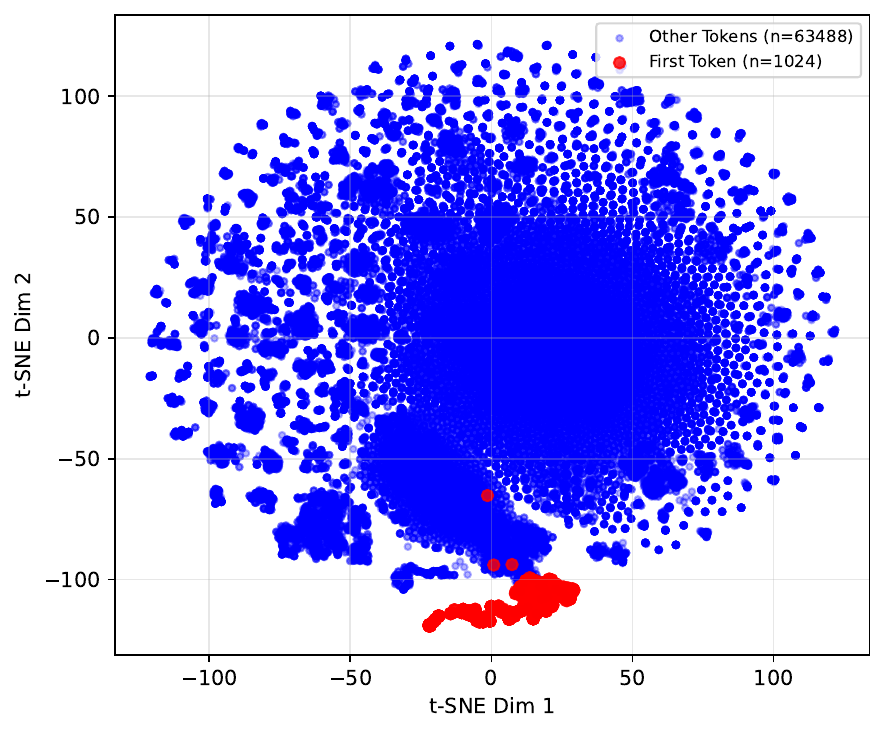}
            \caption{\scriptsize{t-SNE of Layer 0 MLP Intermediate States, w/o Head 29}}
            \label{fig-cone}
        \end{subfigure}
    \end{minipage}
    \caption{2D dimensionality-reduction visualizations of activations in LLaMA3.1-8B with the \texttt{[BOS]} token removed. Each point represents the hidden state of one position, over sequences of up to 64 tokens across 1024 samples from validation dataset.}
    \label{bos-vs-non-bos-fig2}
    \vspace{-1.5em}
\end{figure*}

\subsection{Position Zero Identification Circuit}

We now turn to how the model establishes a positional signature at position zero prior to MLP amplification. Empirically, the position-zero representation is already pushed toward a relatively stable direction across inputs after layer 0, before any strong MLP amplification takes place. \citet{li2026structuraloriginattentionsink} identify value aggregation within the attention module as the key cause, but do not provide a mechanistic account of why and how the model exploits this positional asymmetry. Since activations pass through linear layers before any nonlinearity within a Transformer block, PCA provides a natural tool for capturing this linear structure. Figure~\ref{bos-vs-non-bos-fig2} confirms that positional identification is established before the layer-1 MLP, that is, before the model's second Transformer block. Once certain confounding signals are removed, for example the output of head 29 in layer 0 of LLaMA3.1-8B, the up-projection of the layer-0 MLP already clusters the position-zero states. This subtraction is exact: since attention head outputs are summed in the residual stream, individual head contributions can be isolated and removed precisely. t-SNE offers a more detailed view of this structure than PCA, as it models pairwise distances between activations~\citep{cai2022theoreticalfoundationstsnevisualizing}.

The mechanism underlying this asymmetry is the causal attention mask. Under causal masking, all positions other than zero aggregate diverse context vectors, which reduces the consistency of any shared directional component in their attention outputs. Position zero, by contrast, attends only to itself, so its attention output remains unmixed and preserves its direction more reliably across inputs. This clean signal is what subsequent MLP sublayers can detect and amplify, ultimately producing the stable P0 representation and the corresponding sink pattern.

Motivated by the observation that attention heads' weighted-averaging behaviors play a key role in forming the P0 sink, we develop a simplified theoretical model to estimate the norm of the attention output. We assume that input vectors $\mathbf{x}_0, \dots, \mathbf{x}_{l-1}$ have unit norm due to pre-layer normalization, i.e., $\|\mathbf{x}_i\|_2 = 1$ for all $i$, and model attention as a weighted sum over value vectors $\mathbf{v}_0, \dots, \mathbf{v}_{l-1}$ with weights $\mathbf{p} = (p_0, \dots, p_{l-1})$.

\paragraph{Cone-based model of value vectors.}
As shown in Figure~\ref{cos-sim-fig}, hidden states in LLMs exhibit clear directional bias and are far from isotropic. We capture this by modeling the value vectors $\mathbf{v}_i$ as lying on a fixed-angle cone centered around a unit vector $\mathbf{u} \in \mathbb{R}^d$, each with unit norm and constant cosine similarity $\alpha$ with the cone axis. In practice, the per-channel learnable scaling in RMSNorm can push the realized norm away from 1, so this is an idealization rather than an exact constraint. A convenient construction is:
\begin{equation}
\mathbf{v}_i = \alpha \mathbf{u} + \sqrt{1 - \alpha^2} \, \mathbf{s}_i,
\end{equation}
where each $\mathbf{s}_i$ is a unit vector orthogonal to $\mathbf{u}$, sampled uniformly from the $(d-1)$-dimensional unit sphere and independently across $i$. Under this construction:
\begin{equation}
\mathbb{E}[\mathbf{v}_i^\top \mathbf{v}_j] = 
\begin{cases}
1, & i = j, \\
\alpha^2, & i \neq j.
\end{cases}
\end{equation}

\paragraph{Attention output norm.}
Define the attention output at position $l-1$ as:
\begin{equation}
\mathbf{c} := \sum_{i=0}^{l-1} p_i \mathbf{v}_i.
\end{equation}
Expanding the squared norm and taking expectation over $\mathbf{v}$ conditioned on $\mathbf{p}$:
\begin{equation}
\mathbb{E}_{\mathbf{v}}[\|\mathbf{c}\|_2^2 \mid \mathbf{p}] = \sum_i p_i^2 \cdot 1 + \sum_{i \ne j} p_i p_j \cdot \alpha^2 = \alpha^2 + (1 - \alpha^2) \sum_i p_i^2.
\end{equation}
Taking expectation over $\mathbf{p}$ then gives:
\begin{equation}
\mathbb{E}[\|\mathbf{c}\|_2^2]
= \alpha^2 + (1 - \alpha^2)\, \mathbb{E}\left[\sum_i p_i^2\right].
\label{eq:mixing}
\end{equation}

\begin{wraptable}{r}{0.5\textwidth}
\centering
\resizebox{\linewidth}{!}{
\begin{tabular}{lcccccccc}
\toprule
\textbf{Token Index}           & 0    & 1    & 2    & 4    & 8    & 16   & 32   & 64    \\
\midrule
\textbf{Llama-3.1-8B} & 0.87 & 0.82 & 0.79 & 0.73 & 0.66 & 0.63 & 0.61 & 0.59  \\
\textbf{Llama-3.2-1B} & 1.04 & 1.03 & 0.98 & 0.96 & 0.91 & 0.90 & 0.88 & 0.87  \\
\textbf{Llama-3.2-3B} & 1.37 & 1.29 & 1.21 & 1.17 & 1.11 & 1.10 & 1.07 & 1.05  \\
\bottomrule
\end{tabular}
}
\caption{$\ell_2$ Norm of Layer 0 Attn Out, w/o \texttt{[BOS]}}
\label{non-bos-l2norm-attnout-tab}
\vspace{-1em}
\end{wraptable}

While the full distribution of $\mathbf{p}$ is difficult to characterize analytically, empirical evidence suggests that attention tends to be sparse across positions~\citep{zucchet2025the}, implying that $\mathbb{E}\left[\sum_i p_i^2\right]$ decreases monotonically as sequence length $l$ increases. This is consistent with our empirical finding that the $\ell_2$ norm of the attention output decreases with sequence length, as shown in Table~\ref{non-bos-l2norm-attnout-tab}. Position zero is the limiting case: with $p_0 = 1$ enforced by causal masking, $\sum_i p_i^2 = 1$ regardless of sequence length, yielding a strictly higher expected output norm than any other position. This norm asymmetry is the signal that MLP sublayers detect and amplify into the stable P0 representation.

\section{Free Lunch for Improving Transformers as LLM Backbones}

For years, researchers have sought improved attention mechanisms to strengthen the Transformer decoder as a backbone for auto-regressive LLMs~\citep{peng2026explicitmultiheadattentioninterhead, shazeer2020talkingheadsattention}. Few such efforts have been adopted in modern practice, with the recently notable exceptions of Gated Attention~\citep{qiu2026unifiedviewattentionresidual}. We propose two simple methods that accelerate the formation of a stable P0 sink representation and achieve performance comparable to Gated Attention, while introducing no additional parameters. This makes them particularly attractive during pretraining and pre-filling, the two stages where Gated Attention incurs the most overhead.

\begin{figure*}[t]
    \centering
    \begin{minipage}[b]{0.5\linewidth} 
        \centering
        \begin{subfigure}[b]{0.95\linewidth} 
            \centering
            \caption{Transformer-native LM Training Loss}
            \includegraphics[width=0.95\columnwidth]{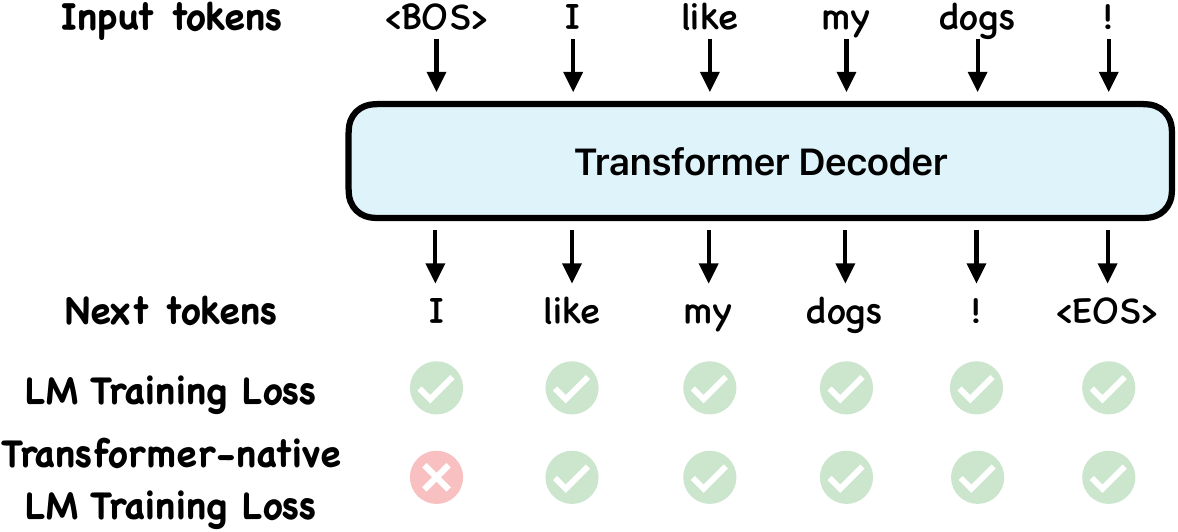}
            \label{fig:tnl}
        \end{subfigure}
    \end{minipage}
    \begin{minipage}[b]{0.48\linewidth} 
        \centering
        \begin{subfigure}[b]{0.95\linewidth} 
            \centering
            \caption{Force-sink Masking}
            \includegraphics[width=0.95\columnwidth]{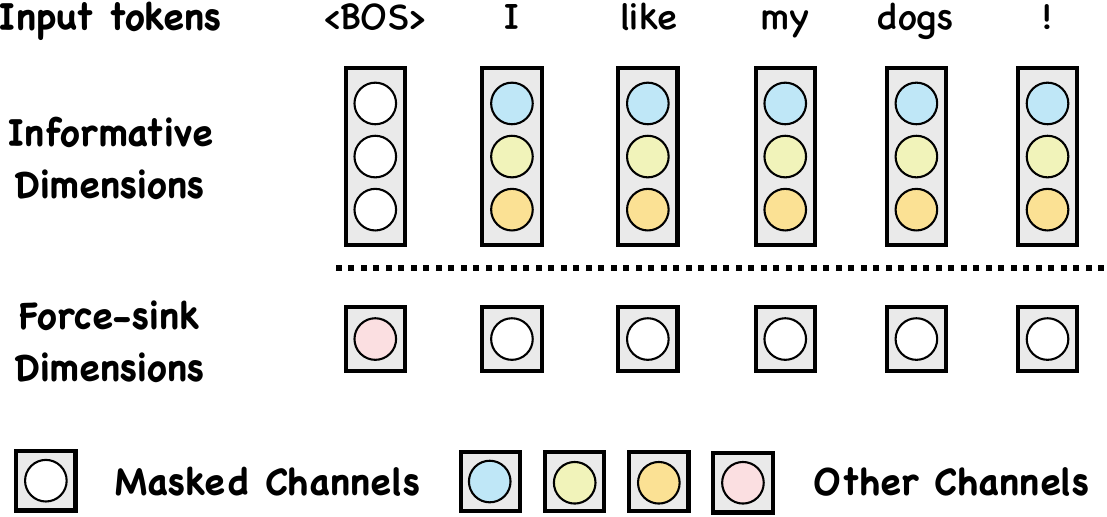}
            \label{fig:fsm}
        \end{subfigure}
    \end{minipage}
    \caption{Illustration of the two variants that modify only how the P0 representation is trained, each designed to accelerate the formation process of the P0-Sink Circuit.}
    \label{two-variant-fig}
    \vspace{-1em}
\end{figure*}

\subsection{Preliminary}
\label{sec:preliminary}

As discussed in Section~\ref{sec:nobos}, modern open-source LLMs show no consistent pattern in their handling of the \texttt{<BOS>} token, and no consensus has emerged on whether to include it during pretraining. Our experimental results are consistent with this observation: pretraining under either choice yields little difference in downstream performance, with detailed results reported in Appendix~\ref{Appendix:performance-bos-ablation}.

Two additional observations inform our experimental setup. First, training on FineWeb-Edu appears to make WikiText~\citep{merity2016pointersentinelmixturemodels} an unreliable benchmark: as training progresses, WikiText perplexity reported by LM-Eval~\citep{eval-harness} degrades, in contrast to the consistent improvement seen on every other benchmark. We therefore omit WikiText results from all tables in the main paper.

Second, under the Chinchilla-optimal setting~\citep{hoffmann2022trainingcomputeoptimallargelanguage}, a 1B model trained on 20B tokens has not yet fully formed its sink, as shown in Table~\ref{tab:sink-rate-bos-vs-non-bos}. Models trained with \texttt{<BOS>} form the sink faster, whereas models trained without \texttt{<BOS>} eventually form a stronger sink in the shallow layers. 

\begin{table}[ht]
\centering
\resizebox{\linewidth}{!}{
\begin{tabular}{l|l|cccccc}
\toprule
\multicolumn{2}{l|}{\textbf{Model}}                                                                                    & \textbf{$\text{Sink}_{0}^{\epsilon}[0:]$} & \textbf{$\text{Sink}_{0}^{\epsilon}[1:]$} & \textbf{$\text{Sink}_{0}^{\epsilon}[2:]$} & \textbf{$\text{Sink}_{0}^{\epsilon}[4:]$} & \textbf{$\text{Sink}_{0}^{\epsilon}[8:]$} & \textbf{$\text{Sink}_{0}^{\epsilon}[16:]$} \\
\midrule
\multirow{2}{*}{\textbf{Trained W/} \texttt{<BOS>}}  & \textbf{Baseline-1B}, Chinchilla-optimal           & 36.33                                     & 37.91                                     & 39.63                                     & 43.53                                     & 47.47                                     & 49.68                                      \\
                                                                                     & \textbf{Baseline-1B}, $5\times$ Chinchilla-optimal           & 76.58                                     & 79.90                                     & 83.53                                     & 87.37                                     & 89.26                                     & 94.01                                      \\
\midrule
\multirow{2}{*}{\textbf{Trained W/O} \texttt{<BOS>}} & \textbf{Baseline-1B}, Chinchilla-optimal           & 10.34                                     & 10.79                                     & 11.28                                     & 12.37                                     & 13.59                                     & 15.98  \\
                                                                                     & \textbf{Baseline-1B}, $5\times$ Chinchilla-optimal           & 79.89                                     & 83.36                                     & 87.15                                     & 90.93                                     & 91.38                                     & 93.28  \\
\bottomrule
\end{tabular}}
\caption{Our modified version of $\text{Sink}^\epsilon_{k}$ under \texttt{<BOS>} ablation shows that the sink builds up gradually from bottom layers. Under the popular Chinchilla-optimal setting, a 1B model trained on 20B tokens has not yet finished forming the sink. At $5\times$ this scale, 100B tokens following the FLAME setup~\citep{yang2025flame}, the sink is clearly visible in both the W/ and W/O \texttt{<BOS>} settings.}
\label{tab:sink-rate-bos-vs-non-bos}
\end{table}

This difference follows from the distinct learning targets the two choices induce. Causal masking restricts the first token to attend only to its own value vector, making the QK interaction irrelevant there; both models reduce to a pointwise function, but with different targets. With \texttt{<BOS>}, the target is the marginal distribution over the first real token, a single fixed distribution. Without \texttt{<BOS>}, the target is a bigram distribution indexed by whichever token occupies the first position, a family of distributions.

To see why this affects sink formation, consider the gradient at position zero decomposed into two components:
\begin{equation}
    \nabla \mathcal{L}_0 = \nabla \mathcal{L}_0^{\text{sink}} + \nabla \mathcal{L}_0^{\text{pred}},
\end{equation}
where $\nabla \mathcal{L}_0^{\text{sink}}$ drives the model toward forming the sink circuit and $\nabla \mathcal{L}_0^{\text{pred}}$ drives it toward capturing the prediction target. With \texttt{<BOS>}, the prediction target is a fixed unigram distribution independent of the input, so $\nabla \mathcal{L}_0^{\text{pred}}$ converges faster and exerts diminishing interference on $\nabla \mathcal{L}_0^{\text{sink}}$. Without \texttt{<BOS>}, the prediction target is a bigram distribution conditioned on the first content token, a strictly harder objective that varies across inputs. The corresponding $\nabla \mathcal{L}_0^{\text{pred}}$ therefore remains large throughout training and persistently interferes with $\nabla \mathcal{L}_0^{\text{sink}}$, slowing sink circuit formation. This accounts for the slower sink development we observe in non-\texttt{<BOS>} models under the Chinchilla-optimal setting.

Based on this analysis, although the two choices are nearly indistinguishable in downstream performance, the \texttt{<BOS>} configuration gives every sequence the same fixed starting token, keeping the setup simple and consistent with common practice. We therefore adopt the \texttt{<BOS>}-in-vocabulary setting for all subsequent experiments.

\subsection{Transformer-Native LM Training Loss}

As established above, the prediction made at the \texttt{<BOS>} position only recovers the marginal distribution over the first real token, a degenerate target that carries no genuine sequence-modeling signal. Moreover, as the gradient decomposition in Section~\ref{sec:preliminary} shows, the corresponding gradient component $\nabla \mathcal{L}_0^{\text{pred}}$ actively interferes with sink circuit formation throughout training. We therefore exclude the cross-entropy term at the \texttt{<BOS>} position from the training objective, as shown in Figure~\ref{fig:tnl}. This simultaneously removes the interfering gradient component and frees the position to act as a dedicated register that holds the sink, motivated by work on register tokens in vision transformers~\citep{darcet2024visiontransformersneedregisters, jiang2025visiontransformersdontneed}: a constant, content-free token can absorb the attention sink without degrading the model's representations. We refer to the resulting objective as the \textbf{Transformer-Native LM Training Loss}, since it removes a prediction target that does not reflect a genuine sequence-modeling task and aligns the training signal with the sink structure the architecture naturally develops.

\subsection{Force-Sink Masking}
\label{sec:fsm}
\begin{figure*}[ht]
    \centering
    \begin{minipage}[b]{0.48\linewidth} 
        \centering
        \begin{subfigure}[b]{0.95\linewidth} 
            \centering
            \caption{Channel values before pre-layer RMS Norm}
            \includegraphics[width=0.95\columnwidth]{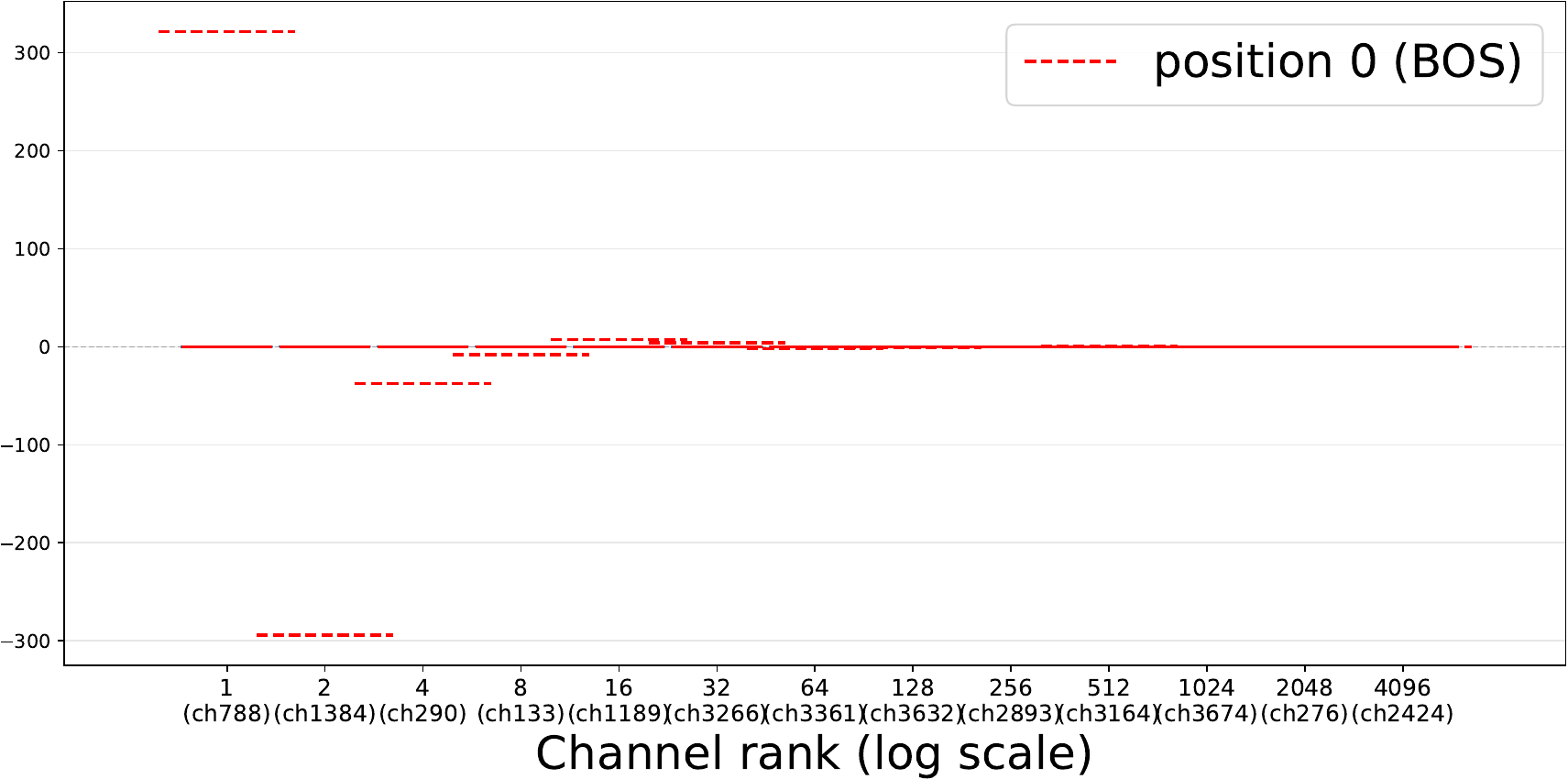}
            \label{fig:raw-box}
        \end{subfigure}
    \end{minipage}
    \begin{minipage}[b]{0.48\linewidth} 
        \centering
        \begin{subfigure}[b]{0.95\linewidth} 
            \centering
            \caption{Channel values after pre-layer RMS Norm}
            \includegraphics[width=0.95\columnwidth]{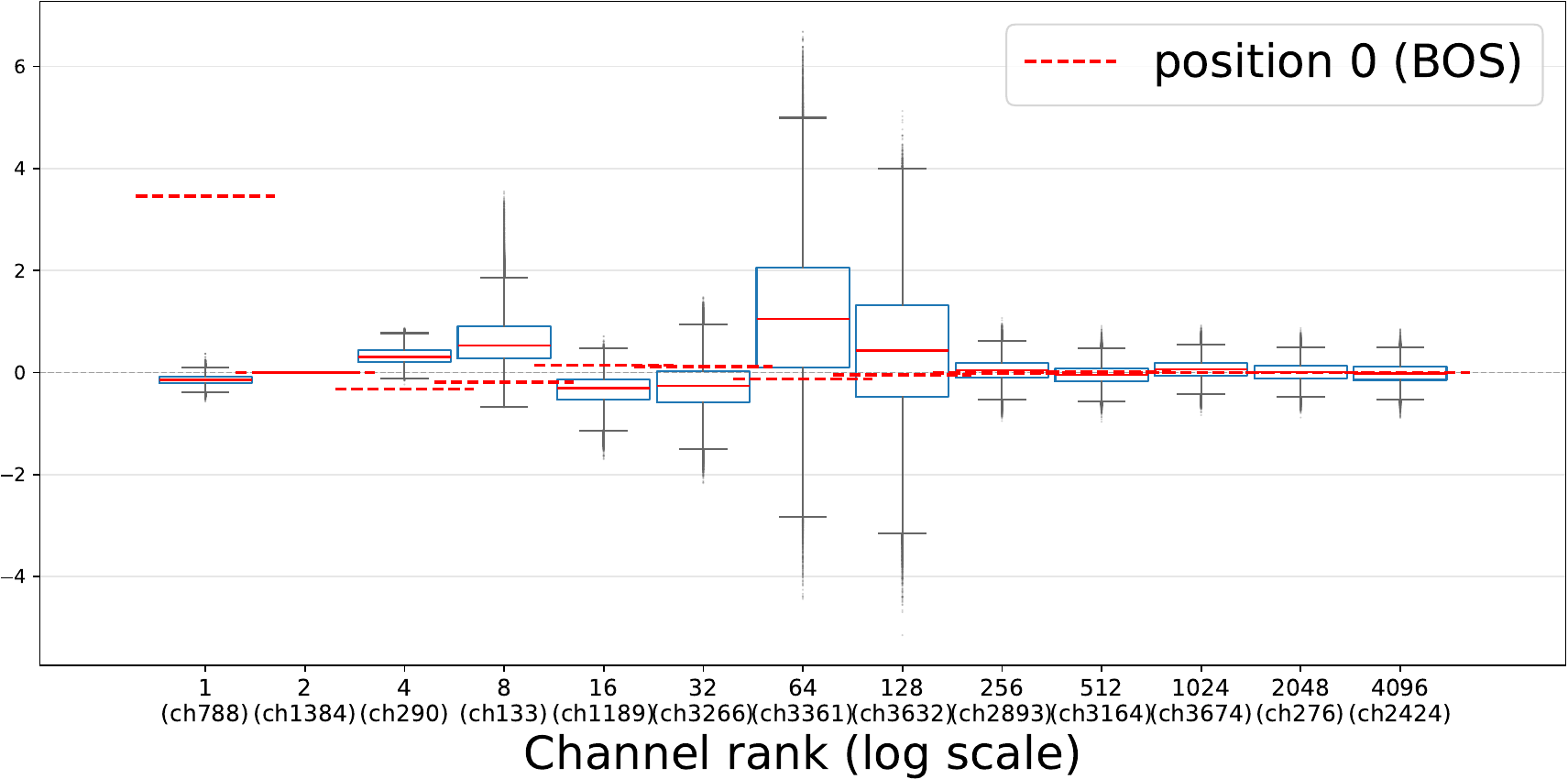}
            \label{fig:rms-box}
        \end{subfigure}
    \end{minipage}
    \caption{Box plots of hidden-state values across channels in LLaMA3.1-8B at layer 2, with channels ranked by their absolute mean value of \texttt{[BOS]}. Since the \texttt{[BOS]} hidden state is fixed across samples, it is shown as a red dashed line for each channel. Layers beyond layer 2 exhibit similar patterns. We count over sequences of up to 64 tokens across 1024 samples from validation dataset.}
    \label{fig:box-plot}
    \vspace{-1.5em}
\end{figure*}

Returning to Section~\ref{sec:how_sink}, the P0-Sink Circuit performs two essential functions: identifying the P0 position, and suppressing non-sink channels by amplifying a subset of them to large magnitude. The latter is also a core issue that recent work has sought to address~\citep{qiu2026unifiedviewattentionresidual}, where this behavior is termed \textbf{outlier-driven rescaling}.

As shown in Figure~\ref{fig:box-plot}, approximately 16--32 channels exhibit statistically significant differences in hidden-state values at the \texttt{[BOS]} position compared to other positions. After applying the pre-layer RMS Norm with learnable scaling factors, these outlier values are sufficiently suppressed that their statistics become comparable to those of other channels. Notably, however, for the \texttt{[BOS]} token, the remaining channels outside this top set are more tightly concentrated around zero relative to other positions.

Based on this observation, we design a single, parameter-free mechanism that realizes both functions simultaneously, as shown in Figure~\ref{fig:fsm}. Let $h_t \in \mathbb{R}^d$ denote the residual-stream hidden state at position $t$. We fix a binary mask $m \in \{0,1\}^d$ selecting $k \ll d$ channels, and apply it at every layer before pre-norm:
\begin{equation}
    h_t \leftarrow
    \begin{cases}
        m \odot h_t, & t = 0, \\
        (\mathbf{1} - m) \odot h_t, & t \neq 0,
    \end{cases}
\end{equation}
where $\odot$ denotes elementwise multiplication and $\mathbf{1} - m$ is the complementary mask over the remaining $d-k$ channels. Applied at every layer, this keeps the P0 representation confined to the same $k$-dimensional subspace throughout the network, while every other token's representation stays confined to the complementary $(d-k)$-dimensional subspace. No channel is ever active for both P0 and non-P0 tokens at any depth.

This hard partition realizes both functions of the P0-Sink Circuit without any additional parameters. First, it directly achieves identification: since the linear layers, which account for nearly all learnable parameters in the Transformer, receive P0 and non-P0 inputs on disjoint subspaces, each layer naturally learns a distinct sub-mapping for the two cases. Second, it eliminates the pressure toward outlier-driven rescaling: since the $d-k$ non-sink channels are already zeroed at P0, suppression is structural rather than learned, and the model never needs to develop large-magnitude outliers to achieve it.

\subsection{Experimental Results}

\paragraph{Experimental Settings}
We follow the Chinchilla-optimal setting~\citep{hoffmann2022trainingcomputeoptimallargelanguage} adopted in prior work~\citep{qiu2025gatedattentionlargelanguage}, training on 20B tokens for a 1B-parameter model.
All other hyperparameters are aligned with the \href{https://huggingface.co/fla-hub/transformer-1.3B-100B}{transformer-1.3B-100B} baseline from the \textsc{Flame} pre-training framework~\citep{yang2025flame}.
Full training details are provided in Appendix~\ref{Appendix:trainging-settings}.

\begin{table}[t]
\resizebox{\linewidth}{!}{
\begin{tabular}{l|c|l|l|lllll|l}
\toprule
\textbf{Method} & \textbf{LR} & \textbf{Sink Rate} & \textbf{Lambada}                       & \textbf{PIQA}                          & \textbf{HellaSwag}                      & \textbf{WinoGrande}                     & \textbf{ARC\_e}                        & \textbf{ARC\_c}                         & \textbf{Avg.}                           \\
\midrule
Baseline*                     & \multirow{5}{*}{\textbf{$3\times10^{-4}$}} & 54.11 & 38.73                                   & 69.10                                   & 49.53                                    & 53.12                                    & 58.16                                   & 31.40                                    & 52.26                                    \\
Gated Attention              &                                             & 0.51 & \textbf{41.78} (\textcolor{red}{+3.05})          & 70.84 (\textcolor{red}{+1.74})          & \textbf{51.41} (\textcolor{red}{+1.88})  & 55.56 (\textcolor{red}{+2.44})           & 61.32 (\textcolor{red}{+3.16})          & 33.62 (\textcolor{red}{+2.22})           & 54.55 (\textcolor{red}{+2.29}) \\
Force-sink Masking           &                                             & 71.27 (\textcolor{red}{+17.16}) & 41.24 (\textcolor{red}{+2.51})          & \textbf{70.95} (\textcolor{red}{+1.85})          & 50.46 (\textcolor{red}{+0.93})           & \textbf{56.04} (\textcolor{red}{+2.92})  & \textbf{62.21} (\textcolor{red}{+4.05})          & 33.36 (\textcolor{red}{+1.96})           & \textbf{54.60} (\textcolor{red}{+2.34})          \\
TNLM Loss                    &                                             & 50.56 (\textcolor{blue}{-3.55}) & 41.72 (\textcolor{red}{+2.99})          & 70.78 (\textcolor{red}{+1.68})          & 50.48 (\textcolor{red}{+0.95})           & 55.01 (\textcolor{red}{+1.89})           & 59.60 (\textcolor{red}{+1.44})          & \textbf{33.79} (\textcolor{red}{+2.39})           & 53.93 (\textcolor{red}{+1.67})          \\
Force-sink Masking+TNLM Loss &                                             & \textbf{71.77} (\textcolor{red}{+17.66}) & 41.18 (\textcolor{red}{+2.45})          & 70.51 (\textcolor{red}{+1.41})          & 50.83 (\textcolor{red}{+1.30})           & 54.78 (\textcolor{red}{+1.66})           & 61.74 (\textcolor{red}{+3.58})          & 33.02 (\textcolor{red}{+1.62})           & 54.18 (\textcolor{red}{+1.91})          \\
\midrule
Baseline*                     & \multirow{5}{*}{\textbf{$4\times10^{-4}$}} & 67.97 & 40.38                                   & 69.97                                   & 50.81                                    & 54.78                                    & 57.62                                   & 33.70                                    & 53.38                                    \\
Gated Attention              &                                             & 0.84 & 42.77 (\textcolor{red}{+2.39})          & 71.11 (\textcolor{red}{+1.14})          & \textbf{52.90} (\textcolor{red}{+2.09})  & 55.09 (\textcolor{red}{+0.31})           & 62.12 (\textcolor{red}{+4.50})          & 35.58 (\textcolor{red}{+1.88})           & \textbf{55.36} (\textcolor{red}{+1.98})          \\
Force-sink Masking           &                                             & 74.68 (\textcolor{red}{+6.71}) & \textbf{44.71} (\textcolor{red}{+4.33})          & 71.22 (\textcolor{red}{+1.25})          & 51.40 (\textcolor{red}{+0.59})           & 55.01 (\textcolor{red}{+0.23})           & \textbf{63.38} (\textcolor{red}{+5.76})          & \textbf{35.75} (\textcolor{red}{+2.05})           & 55.35 (\textcolor{red}{+1.98}) \\
TNLM Loss                    &                                             & 54.92 (\textcolor{blue}{-13.05}) & 42.15 (\textcolor{red}{+1.77})          & \textbf{71.71} (\textcolor{red}{+1.74})          & 51.40 (\textcolor{red}{+0.59})           & 54.30 (\textcolor{blue}{-0.48})          & 58.67 (\textcolor{red}{+1.05})          & 33.53 (\textcolor{blue}{-0.17})          & 53.92 (\textcolor{red}{+0.55})          \\
Force-sink Masking+TNLM Loss &                                             & \textbf{74.88} (\textcolor{red}{+6.91}) & 42.67 (\textcolor{red}{+2.29})          & 71.27 (\textcolor{red}{+1.30})          & 51.55 (\textcolor{red}{+0.74})           & \textbf{56.20} (\textcolor{red}{+1.42})  & 61.83 (\textcolor{red}{+4.21})          & 33.70 (\textcolor{red}{+0.00})           & 54.91 (\textcolor{red}{+1.53})          \\
\midrule
Baseline*                     & \multirow{5}{*}{\textbf{$5\times10^{-4}$}} & 33.24 & 39.38                                   & 70.51                                   & 50.12                                    & 55.17                                    & 60.90                                   & 33.45                                    & 54.03                                    \\
Gated Attention              &                                             & - & 43.78 (\textcolor{red}{+4.40})          & 71.76 (\textcolor{red}{+1.25})          & \textbf{53.26} (\textcolor{red}{+3.14})  & 57.22 (\textcolor{red}{+2.05})           & 62.12 (\textcolor{red}{+1.22})          & \textbf{35.49} (\textcolor{red}{+2.04})           & 55.97 (\textcolor{red}{+1.94})          \\
Force-sink Masking           &                                             & 76.38 (\textcolor{red}{+43.14}) & \textbf{44.25} (\textcolor{red}{+4.87})          & 71.38 (\textcolor{red}{+0.87})          & 52.51 (\textcolor{red}{+2.39})           & 57.54 (\textcolor{red}{+2.37})           & 62.88 (\textcolor{red}{+1.98})          & 35.41 (\textcolor{red}{+1.96})           & 55.94 (\textcolor{red}{+1.91}) \\
TNLM Loss                    &                                             & 57.31 (\textcolor{red}{+24.07}) & 41.49 (\textcolor{red}{+2.11})          & \textbf{72.47} (\textcolor{red}{+1.96})          & 52.42 (\textcolor{red}{+2.30})           & 57.38 (\textcolor{red}{+2.21})           & 61.20 (\textcolor{red}{+0.30})          & 33.70 (\textcolor{red}{+0.25})           & 55.43 (\textcolor{red}{+1.40})          \\
Force-sink Masking+TNLM Loss &                                             & \textbf{76.97} (\textcolor{red}{+43.73}) & 42.48 (\textcolor{red}{+3.10})          & 71.16 (\textcolor{red}{+0.65})          & 52.64 (\textcolor{red}{+2.52})           & \textbf{58.33} (\textcolor{red}{+3.16})  & \textbf{63.17} (\textcolor{red}{+2.27})          & 34.73 (\textcolor{red}{+1.28})           & \textbf{56.01} (\textcolor{red}{+1.98})          \\
\bottomrule
\end{tabular}
}
\caption{Ablations under the Chinchilla-optimal setting~\citep{hoffmann2022trainingcomputeoptimallargelanguage}, using a 1B model trained on 20B tokens across different learning rates. \textbf{Bold} indicates the best performance within the same setting. The star (*) indicates a different model checkpoint from the one used in the previous section. For a fair comparison, we add the same number of parameters to the MLP layer of both the baseline model and our method, since Gated Attention introduces an entire additional linear layer to implement its gating function.}
\label{tab:main-exp}
\end{table}

Table~\ref{tab:main-exp} presents an ablation study across three learning rate settings. Both proposed methods consistently improve downstream performance over the baseline. Force-sink Masking outperforms Gated Attention on most tasks, while TNLM Loss alone also yields competitive results. Their combination achieves solid performance as well. Besides, as shown in Appendix~\ref{appendix:l2-norm}, Force-sink Masking most effectively suppresses outlier values in the hidden states.

One exception is the drop in Sink Rate for TNLM Loss at $4\times10^{-4}$ (marked in blue). Under this setting, the model does not form a P0 sink until layer 6, resembling the behavior of Qwen3-4B, where the sink emerges in deeper layers with more severely concentrated attention on the P0 token. Further discussion is provided in Appendix~\ref{appendix:sink-rate-on-main-exp}.

More broadly, the results suggest that earlier P0 sink formation is beneficial during pre-training: across comparable settings, a higher Sink Rate consistently corresponds to better downstream performance. Beyond promoting earlier sink formation, both methods also improve task performance directly, supporting the view that a stable P0 sink representation helps stabilize the QK optimization landscape during pre-training. We additionally evaluate both methods under data-saturated settings at $5\times$ the Chinchilla-optimal data scale; results are reported in Appendix~\ref{appendix:5xchinchilla}, where Gated Attention shows diminished gains while our methods remain effective.

\section{Conclusion}
We proposed the \textbf{P0-Sink Circuit}, a two-block subnetwork that explains the consistent emergence of the position-zero attention sink in causal LLMs. Position zero attends only to itself under causal masking, producing a directionally stable output that MLP sublayers amplify into a high-norm, fixed-direction representation, without relying on any semantic content.

We further proposed \textbf{TNLM Loss}, which removes the cross-entropy term at \texttt{[BOS]} to free it as a dedicated attention register, and \textbf{Force-sink Masking}, which hard-partitions hidden dimensions between position zero and all other positions, structurally realizing the P0-Sink Circuit and alleviating outlier-driven rescaling. Both are parameter-free and require no architectural changes.

Experiments across diverse learning rate and data scale settings confirm that earlier P0 sink formation correlates with better downstream performance, both methods consistently improve over the Transformer baseline, and Force-sink Masking achieves performance comparable to Gated Attention.

\bibliography{iclr2027_conference}
\bibliographystyle{iclr2027_conference}

\newpage

\appendix
\section{Ablation Study on the \texttt{[BOS]} Token}
\label{Appendix:performance-bos-ablation}

\begin{table}[ht]
\centering
\resizebox{\linewidth}{!}{
\begin{tabular}{l|l|cc|c|ccccc}
\toprule
\multicolumn{2}{l|}{\multirow{2}{*}{\textbf{Model}}}                                        & \multicolumn{2}{c|}{\textbf{Wikitext}} & \textbf{Lambada} & \textbf{PIQA} & \textbf{HellaSwag} & \textbf{WinoGrande} & \textbf{ARC\_e} & \textbf{ARC\_c} \\
\multicolumn{2}{l|}{}                                                                       & bits per byte$\downarrow$       & byte ppl$\downarrow$       & acc$\uparrow$              & acc$\uparrow$           & acc$\uparrow$                & acc$\uparrow$                 & acc$\uparrow$             & acc$\uparrow$             \\
\midrule
\multirow{2}{*}{\textbf{Trained W/ \textless{}BOS\textgreater{}}}  & \textbf{Baseline}, Chinchilla-optimal    & 1.9132               & 3.7664         & 39.96            & 70.57         & 49.70              & 54.62               & 59.22           & 33.19           \\
                                                                   & \textbf{Baseline}, 5x Chinchilla-optimal & 2.2953               & 4.9084         & 47.80            & 74.16         & 59.13              & 59.43               & 68.69           & 39.16           \\
\midrule
\multirow{2}{*}{\textbf{Trained W/O \textless{}BOS\textgreater{}}} & \textbf{Baseline}, Chinchilla-optimal    & 1.8124               & 3.5122         & 39.57            & 71.44         & 50.11              & 52.09               & 61.32           & 34.04           \\
                                                                   & \textbf{Baseline}, 5x Chinchilla-optimal & 2.3440               & 5.0772         & 47.88            & 74.10         & 59.03              & 59.27               & 68.56           & 40.27 \\
\bottomrule
\end{tabular}}
\caption{Downstream Performance for the \texttt{[BOS]} ablation.}
\label{tab:performance-bos-vs-nobos-pretrain}
\end{table}

As discussed in Section~\ref{sec:nobos}, modern open-source LLMs show no consistent pattern in their handling of the \texttt{[BOS]} token, and no consensus has emerged on whether to include it during pre-training. Our experimental results are consistent with this observation: pre-training under either choice yields little difference in downstream performance, as shown in Table~\ref{tab:performance-bos-vs-nobos-pretrain}.

We also note that training on FineWeb-Edu makes WikiText~\citep{merity2016pointersentinelmixturemodels} an unreliable benchmark. As training progresses, WikiText perplexity reported by LM-Eval~\citep{eval-harness} degrades, in contrast to the consistent improvement observed on every other benchmark. We therefore omit WikiText results from all tables in the main paper.

\section{Experimental Details}
\subsection{Experimental Setups}
\label{Appendix:trainging-settings}

\begin{table}[ht]
\centering
\resizebox{\linewidth}{!}{
\begin{tabular}{lccc}
\toprule
 & \textbf{Baseline for \texttt{[BOS]} Ablation} & \textbf{Gated Attention} & \textbf{Baseline for Main Exp.} \\
\midrule
\multicolumn{4}{l}{\textbf{Model architecture}} \\
\midrule
Parameters                & $\approx$1.36B ($\approx$1.23B non-emb) & $\approx$1.46B ($\approx$1.33B non-emb) & $\approx$1.44B ($\approx$1.31B non-emb) \\
Hidden size $d_\text{model}$ & \multicolumn{3}{c}{2048} \\
Layers                    & \multicolumn{3}{c}{24} \\
Attention heads           & \multicolumn{3}{c}{32} \\
KV heads                  & \multicolumn{3}{c}{32} \\
Attention linear layers   & 4 (WQ, WK, WV, WO) & 5 ($+$ output gate $W_\theta$) & 4 (WQ, WK, WV, WO) \\
FFN activation            & \multicolumn{3}{c}{SwiGLU} \\
FFN intermediate dim      & 5632 & 5632 & 6144 \\
FFN ratio                 & 8 & 8 & 9 \\
Normalization             & \multicolumn{3}{c}{RMSNorm, $\epsilon=10^{-6}$} \\
Positional encoding       & \multicolumn{3}{c}{RoPE, $\theta=10^4$} \\
Context length            & \multicolumn{3}{c}{2048} \\
Vocabulary                & \multicolumn{3}{c}{32{,}000} \\
Tied embeddings           & \multicolumn{3}{c}{No} \\
Init.\ std                & \multicolumn{3}{c}{0.02} \\
\midrule
\multicolumn{4}{l}{\textbf{Training / optimization}} \\
\midrule
Global batch              & \multicolumn{3}{c}{256 seqs $\times$ 2048 = 0.5M tokens/step} \\
Sequence length           & \multicolumn{3}{c}{2048} \\
Optimizer                 & \multicolumn{3}{c}{AdamW ($\beta_1{=}0.9,\ \beta_2{=}0.95,\ \epsilon{=}10^{-15}$)} \\
Weight decay              & \multicolumn{3}{c}{0.1} \\
Peak learning rate        & \multicolumn{3}{c}{$3\times10^{-4}$} \\
LR schedule               & \multicolumn{3}{c}{cosine decay to $0.1\times$ peak learning rate} \\
Warmup                    & \multicolumn{3}{c}{1024 steps ($\approx$0.54B tokens)} \\
Gradient clipping         & \multicolumn{3}{c}{1.0 (global norm)} \\
Precision                 & \multicolumn{3}{c}{bfloat16 mixed precision (fp32 reduce)} \\
Data                      & \multicolumn{3}{c}{FineWeb-Edu (streaming, online tokenization)} \\
Seed                      & \multicolumn{3}{c}{42} \\
\bottomrule
\end{tabular}}
\caption{Model and training configuration for the three 1B runs. All hyperparameters, data, and seed are shared; the models differ only in the attention output gate (Gated Attention) and the FFN ratio. The FFN ratio is defined relative to a single attention linear layer in Flame's documentation; note that the Flame repository incorrectly reports half this value for LLaMA-style FFN, and our configurations reflect the corrected definition.}
\label{tab:train-config}
\end{table}

We run experiments at peak learning rates of $3\times10^{-4}$, $4\times10^{-4}$, and $5\times10^{-4}$ under the Chinchilla-optimal setting for our main experiments, training for 40{,}960 steps ($\sim$20B token). For the $5\times$ Chinchilla-optimal (data-saturated) setting in Appendix~\ref{appendix:5xchinchilla}, we train for 204{,}800 steps ($\sim$100B token). For a fair parameter count comparison, we allocate the additional parameters to the FFN rather than introducing an attention output gate as in Gated Attention; \textbf{this also reduces the number of additional activations introduced during training.}

\paragraph{Experiment-Specific Settings} For \textbf{TNLM Loss}, no additional configuration is required beyond excluding the \texttt{[BOS]} position from the cross-entropy loss. For \textbf{Force-sink Masking}, we use $k=16$ channels as the Force-sink dimensions (Figure~\ref{fig:fsm}). In the data-saturated experiments reported in Appendix~\ref{appendix:5xchinchilla}, Force-sink Masking$^+$ uses $k=32$. As discussed in Section~\ref{sec:fsm}, there is no principled criterion for choosing $k$; we adopt $k=16$ for the Chinchilla-optimal setting as it performs well, and extend to $k=32$ for the data-saturated setting where $k=16$ yields suboptimal results. In practice, we reduce the model's hidden dimensions by $k$ via masking rather than appending extra dimensions, ensuring a fair parameter comparison with the baseline. The baseline model is shared across other experiments and thus it does not use an output normalization layer right before lm head layer; we find no evidence that this affects the validity of our comparisons. Following \citet{yang2025ropenopeagainnew}, QK-norm seems to bring no advantages, so we do not apply QK-norm to any of our models.

\subsection{Experimental Visualization}

\begin{figure*}[ht]
    \centering
    \includegraphics[width=\columnwidth]{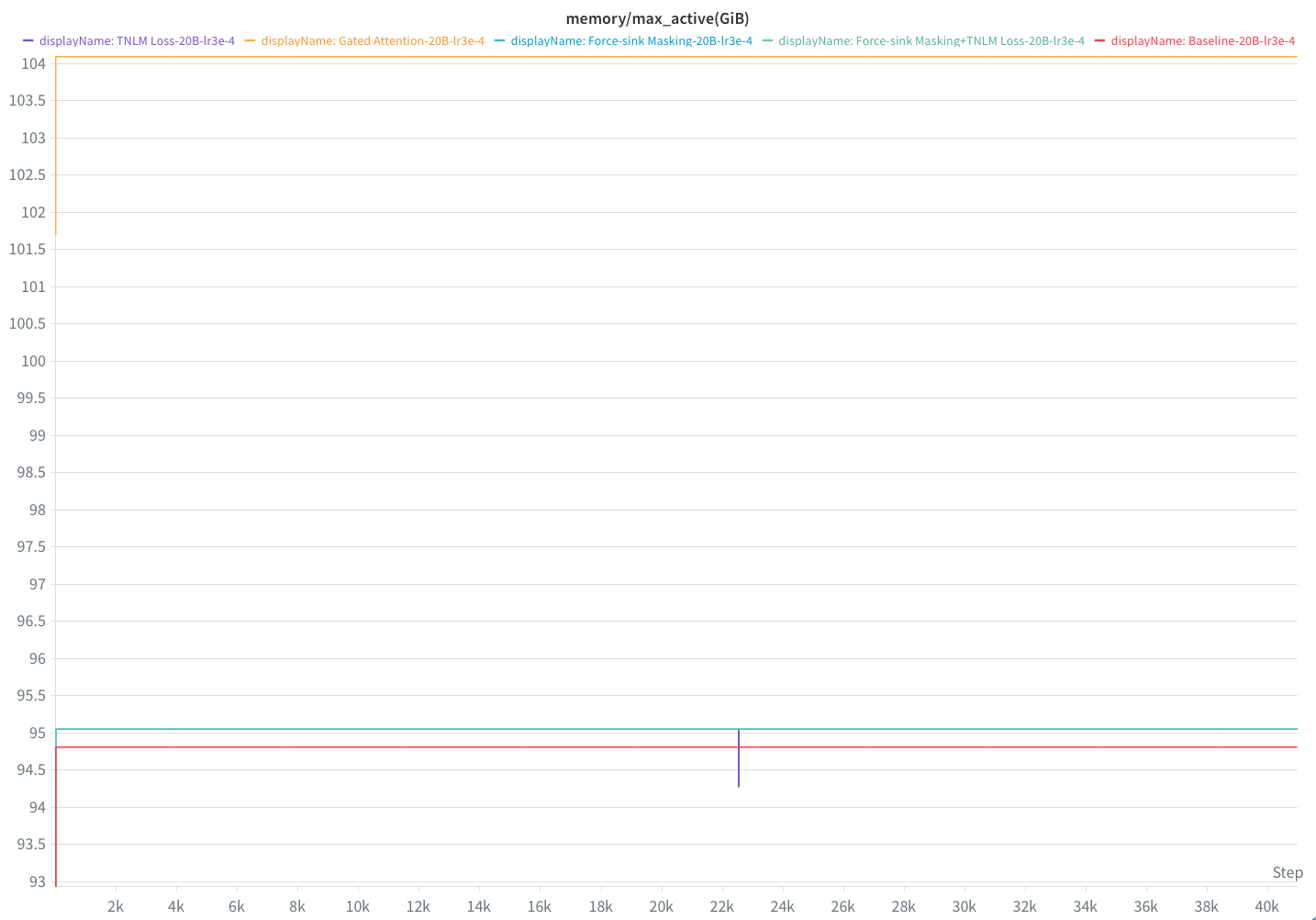}
    \caption{Memory occupation exported from Weights \& Biases. A single representative run is shown; all other configurations exhibit identical behavior.}
    \label{fig:memory}
    \vspace{-1em}
\end{figure*}

As noted above, Gated Attention introduces an entire additional linear layer in the attention module, resulting in increased activation memory overhead during pre-training, as shown in Figure~\ref{fig:memory}.

Despite this overhead, our methods achieve comparable loss to Gated Attention during pre-training.

\begin{figure*}[ht]
    \centering
    \begin{minipage}[b]{0.48\linewidth} 
        \centering
        \begin{subfigure}[b]{0.95\linewidth} 
            \centering
            \caption{\tiny{Training loss, $lr=3\times10^4$ }}
            \includegraphics[width=0.95\columnwidth]{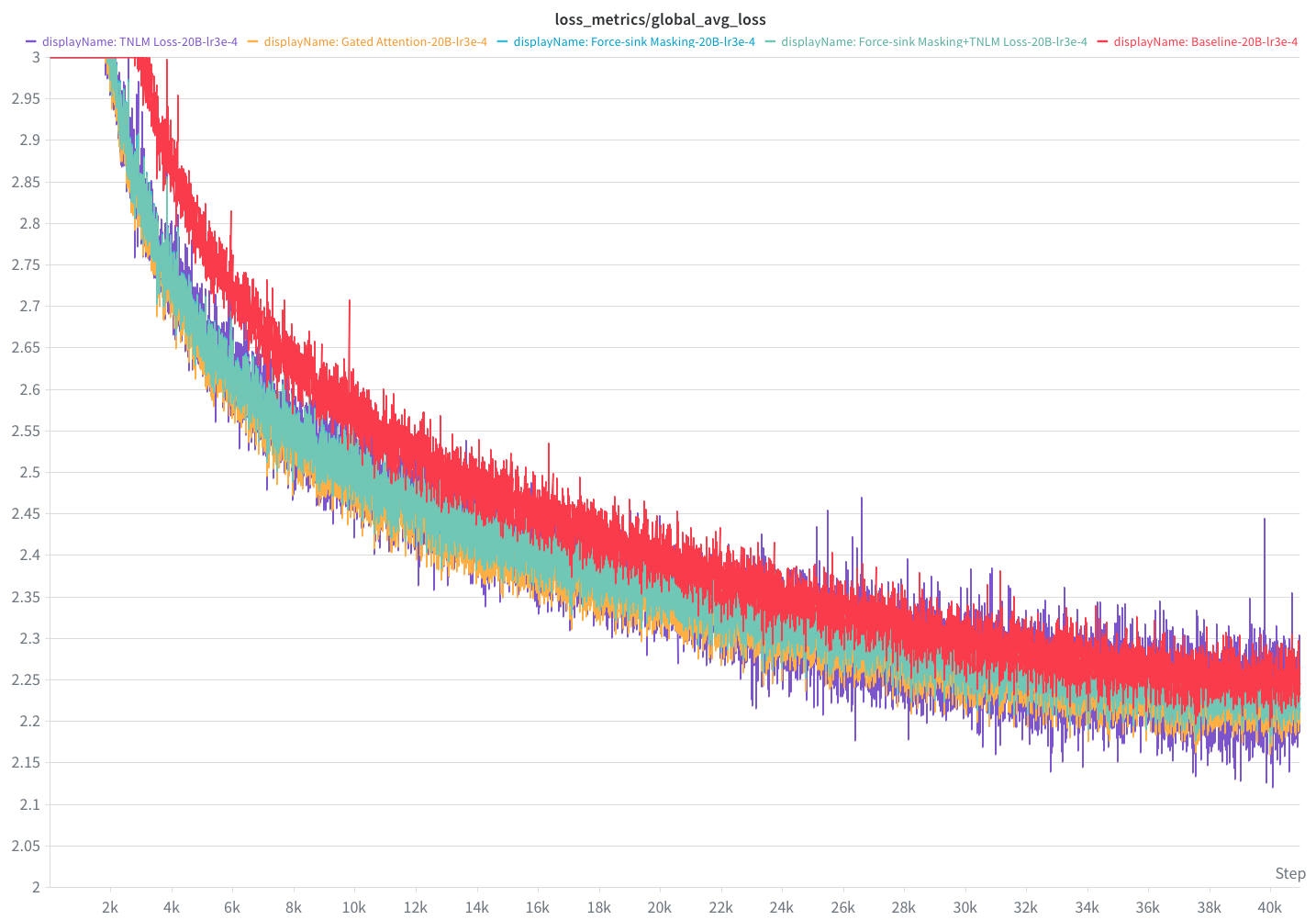}
        \end{subfigure}
    \end{minipage}
    \begin{minipage}[b]{0.48\linewidth} 
        \centering
        \begin{subfigure}[b]{0.95\linewidth} 
            \centering
            \caption{\tiny{Training loss, $lr=4\times10^4$}}
            \includegraphics[width=0.95\columnwidth]{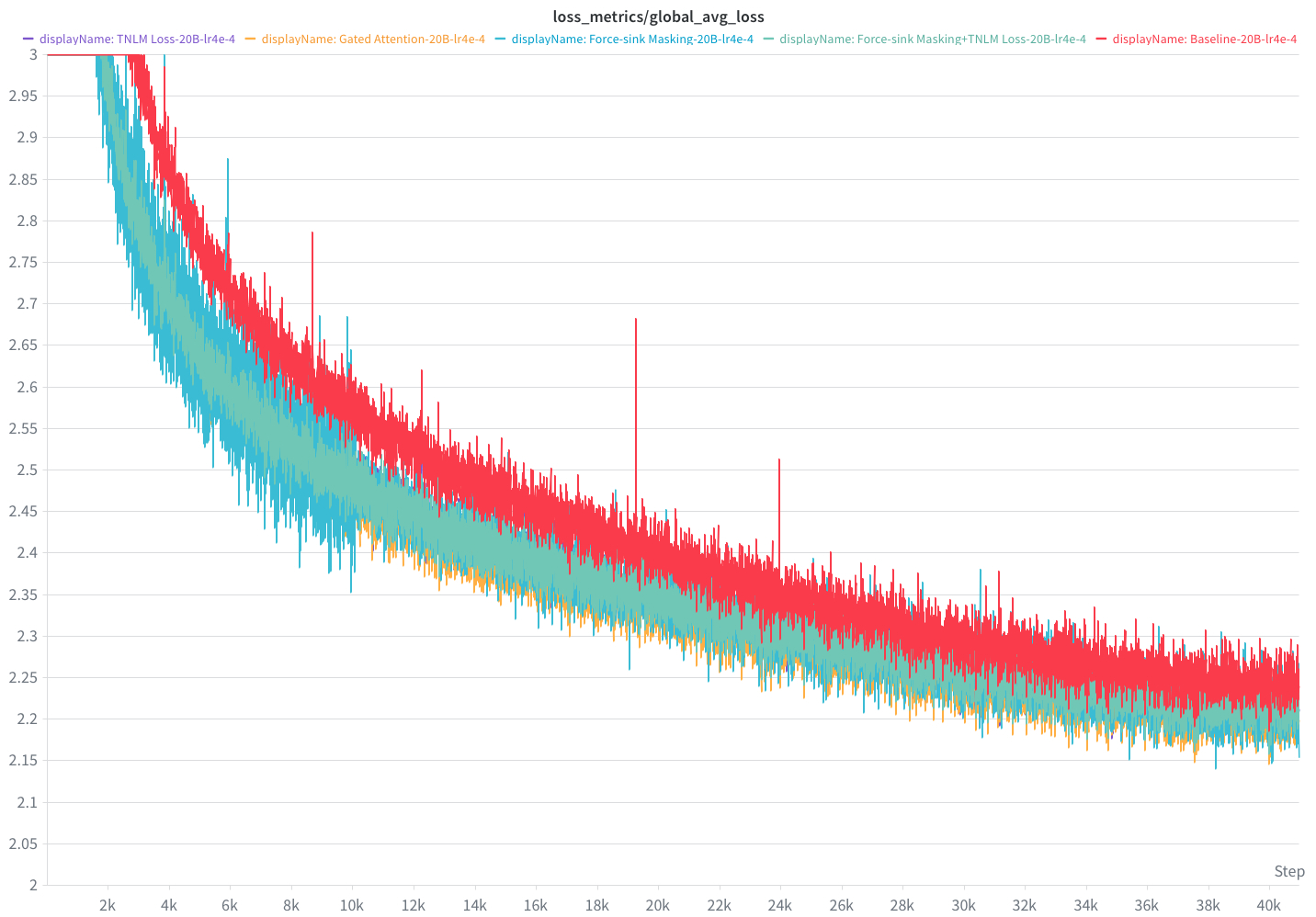}
        \end{subfigure}
    \end{minipage}
    \begin{minipage}[b]{0.48\linewidth} 
        \centering
        \begin{subfigure}[b]{0.95\linewidth} 
            \centering
            \caption{\tiny{Training loss, $lr=5\times10^4$}}
            \includegraphics[width=0.95\columnwidth]{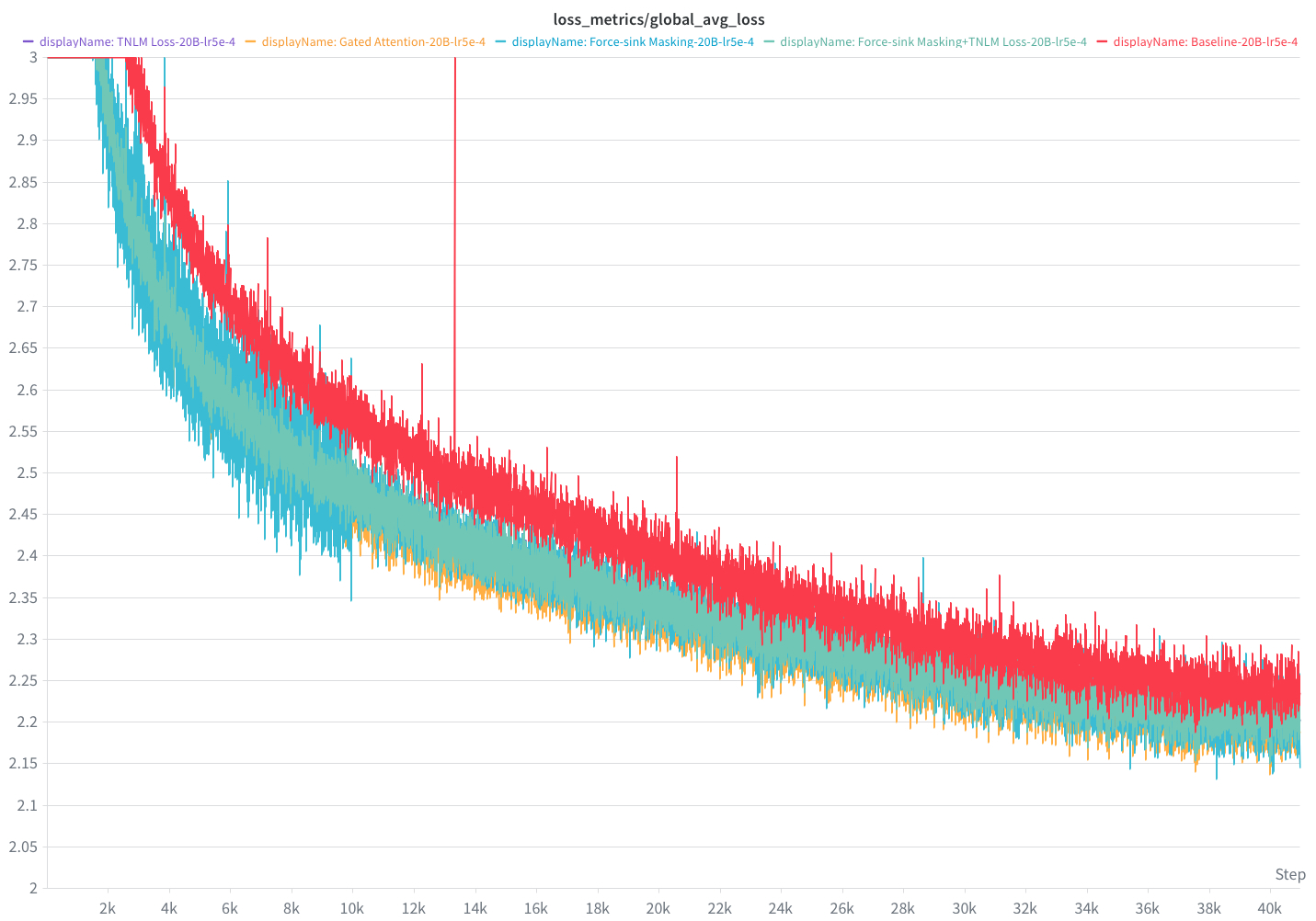}
        \end{subfigure}
    \end{minipage}
    \begin{minipage}[b]{0.48\linewidth} 
        \centering
        \begin{subfigure}[b]{0.95\linewidth} 
            \centering
            \caption{\tiny{Training loss, data-saturated}}
            \includegraphics[width=0.95\columnwidth]{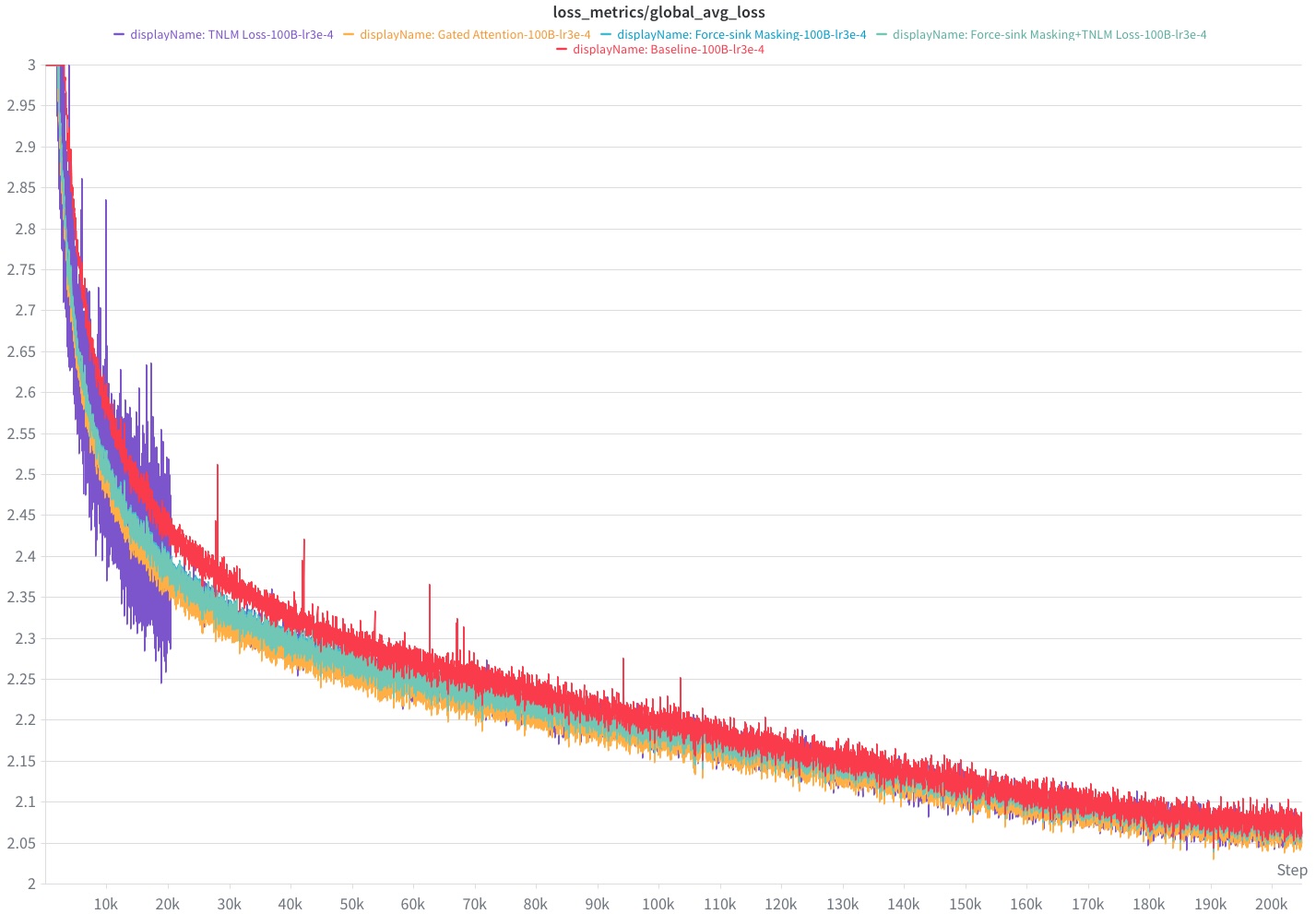}
        \end{subfigure}
    \end{minipage}
    \caption{Training loss of all the models in different training settings.}
    \label{fig:Training loss-baseline}
\end{figure*}

\FloatBarrier

\subsection{Code and Weights}

The checkpoints are available at https://huggingface.co/Anonymous2027. All checkpoints can be loaded with \href{https://github.com/Pryest/flash-linear-attention}{our modified FLA repository}.

\newpage

\section{Outliers Values in Activations}
\label{appendix:l2-norm}

As discussed in Section~\ref{subsec:amplified_l2}, the co-occurrence of attention sinks and outlier activation values is well documented~\citep{an2025systematic,gu2025attentionsinkemergeslanguage,jiang2025visiontransformersdontneed,sun2024massiveactivationslargelanguage}, and our theory provides a mechanistic account of the \textbf{outlier-driven rescaling} phenomenon~\citep{qiu2026unifiedviewattentionresidual}. However, as we show in this section, outlier values are not a necessary condition for the attention sink: the two phenomena can be decoupled.

As shown in Table~\ref{tab:main-exp}, Force-sink Masking yields a higher Sink Rate than the baseline, indicating a stronger P0 sink. Yet the corresponding hidden-state $\ell_2$ norms are substantially smaller, demonstrating that a strong attention sink can emerge without large activation outliers.

\begin{figure*}[ht]
    \centering
    \begin{minipage}[b]{0.48\linewidth} 
        \centering
        \begin{subfigure}[b]{0.95\linewidth} 
            \centering
            \caption{\tiny{$\ell_2$ Norm of Activations, $lr=3\times10^4$ }}
            \includegraphics[width=0.95\columnwidth]{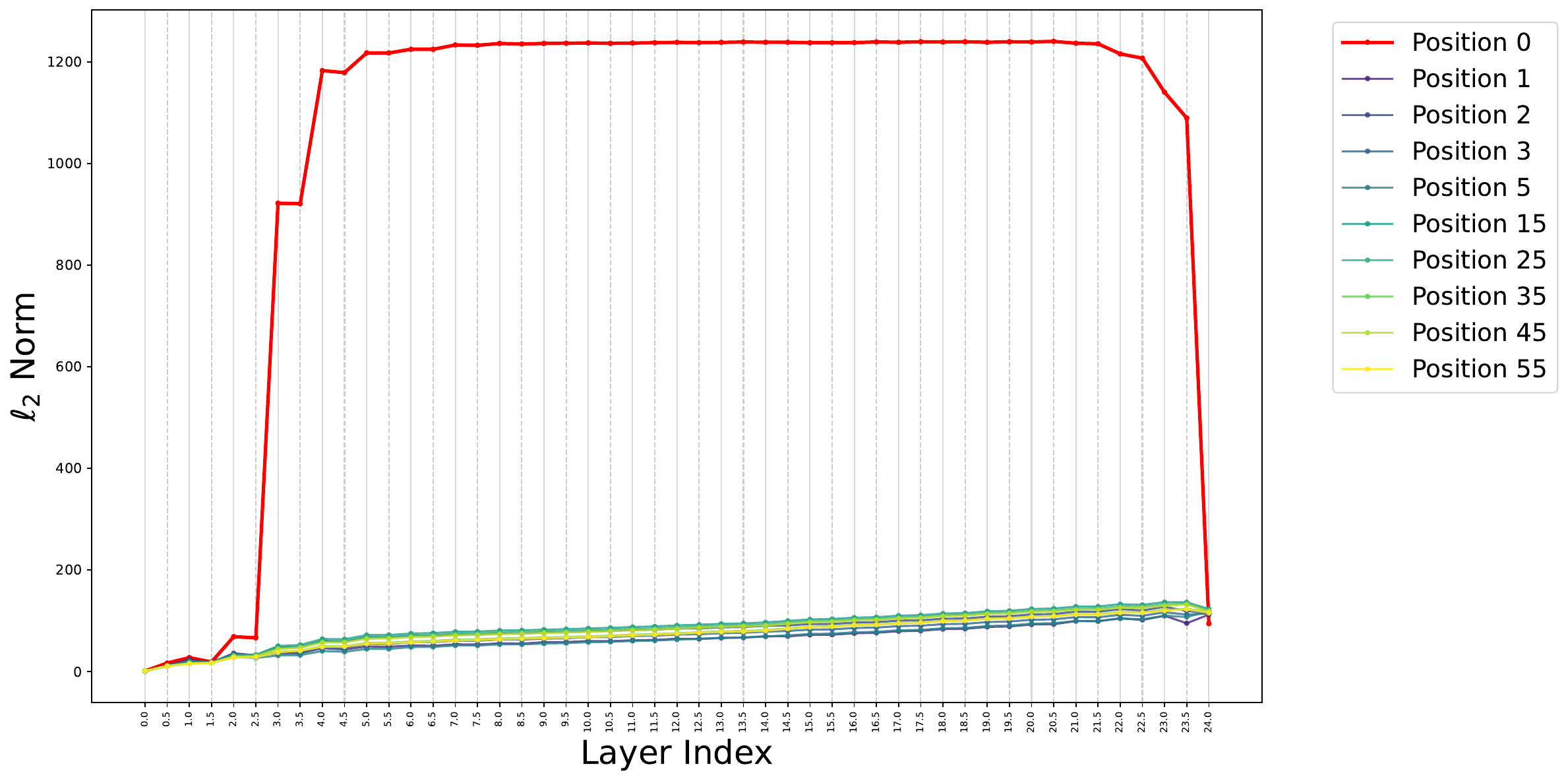}
        \end{subfigure}
    \end{minipage}
    \begin{minipage}[b]{0.48\linewidth} 
        \centering
        \begin{subfigure}[b]{0.95\linewidth} 
            \centering
            \caption{\tiny{$\ell_2$ Norm of Activations, $lr=4\times10^4$}}
            \includegraphics[width=0.95\columnwidth]{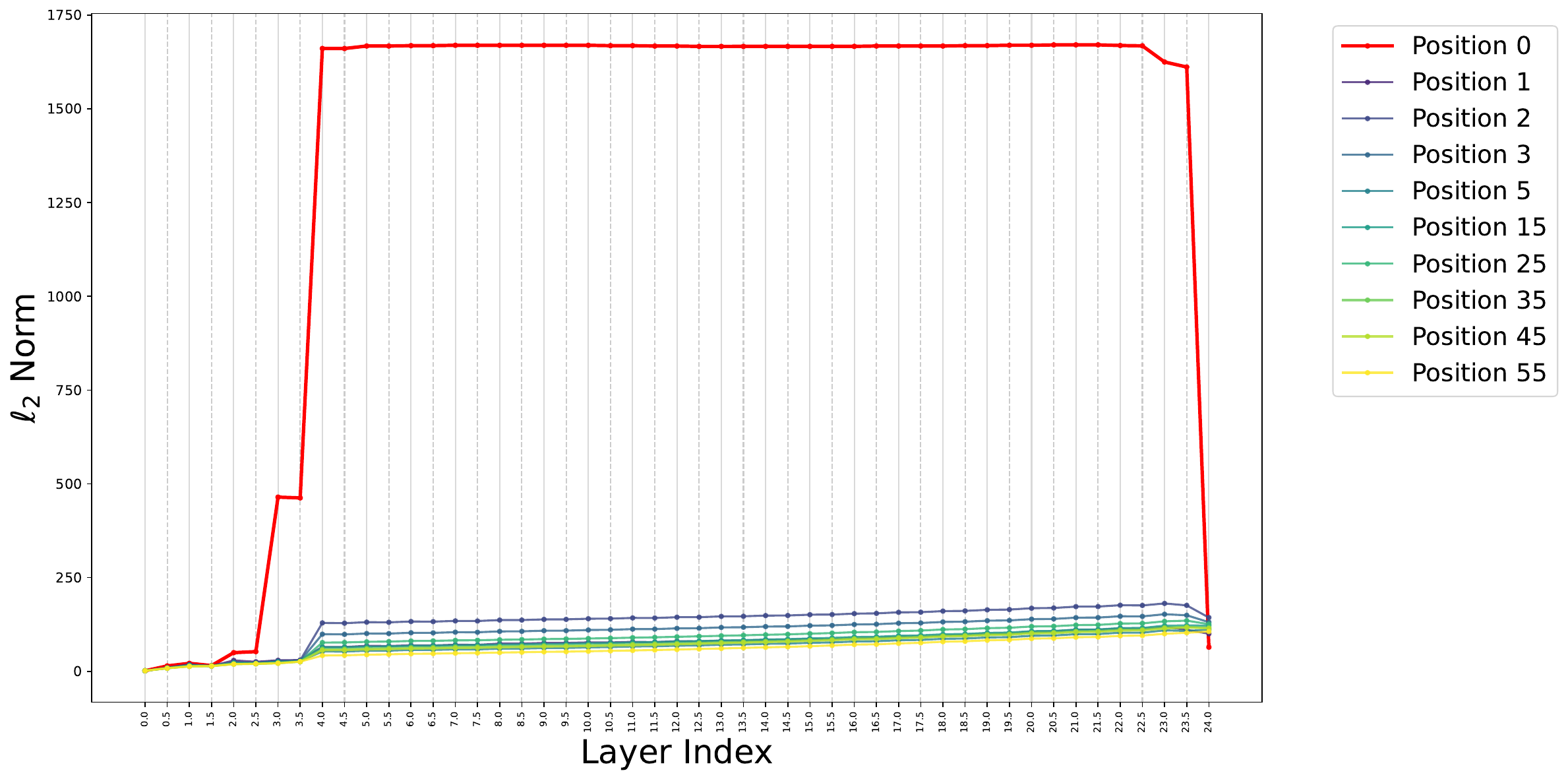}
        \end{subfigure}
    \end{minipage}
    \begin{minipage}[b]{0.48\linewidth} 
        \centering
        \begin{subfigure}[b]{0.95\linewidth} 
            \centering
            \caption{\tiny{$\ell_2$ Norm of Activations, $lr=5\times10^4$}}
            \includegraphics[width=0.95\columnwidth]{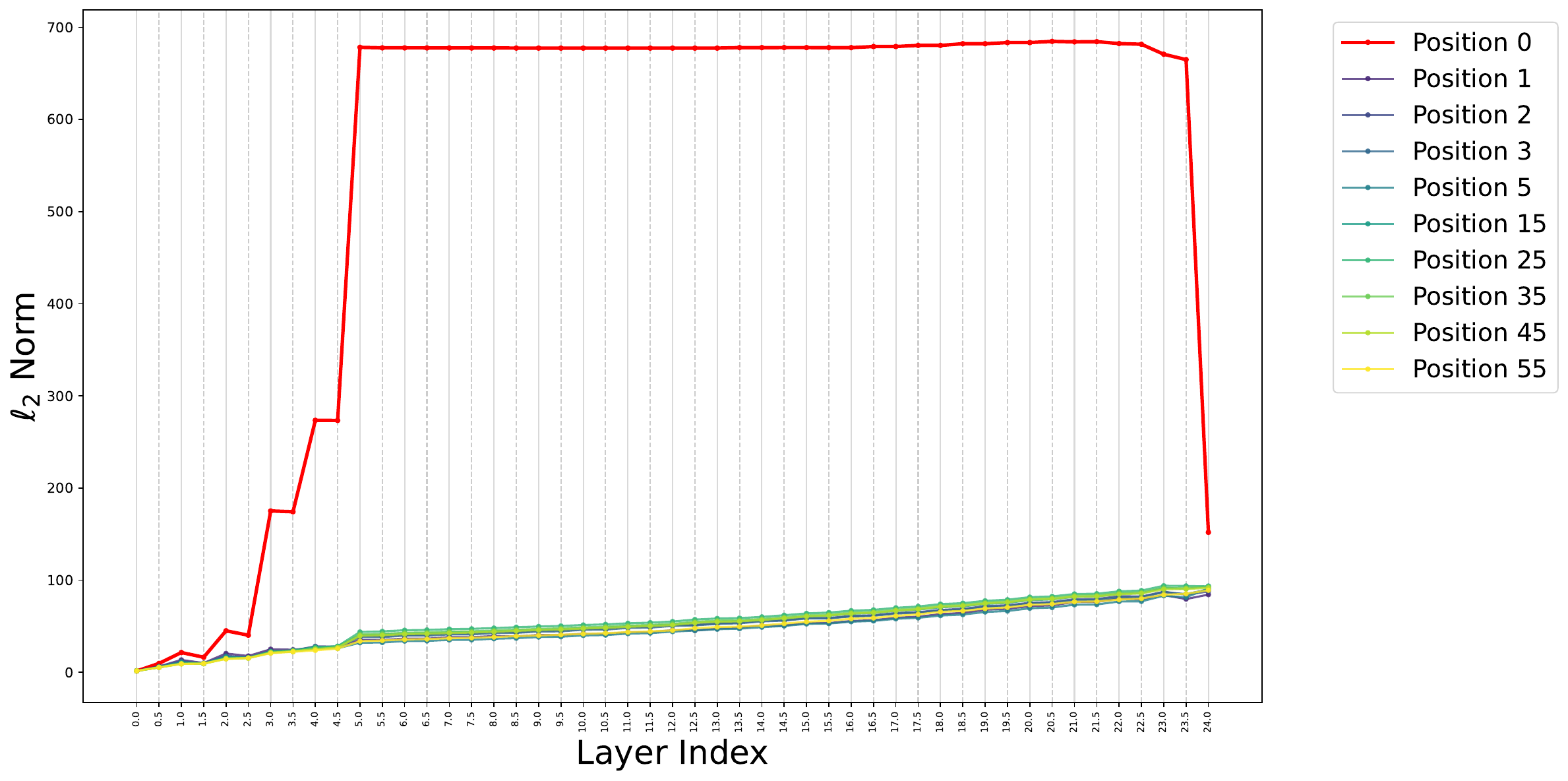}
        \end{subfigure}
    \end{minipage}
    \begin{minipage}[b]{0.48\linewidth} 
        \centering
        \begin{subfigure}[b]{0.95\linewidth} 
            \centering
            \caption{\tiny{$\ell_2$ Norm of Activations, data-saturated}}
            \includegraphics[width=0.95\columnwidth]{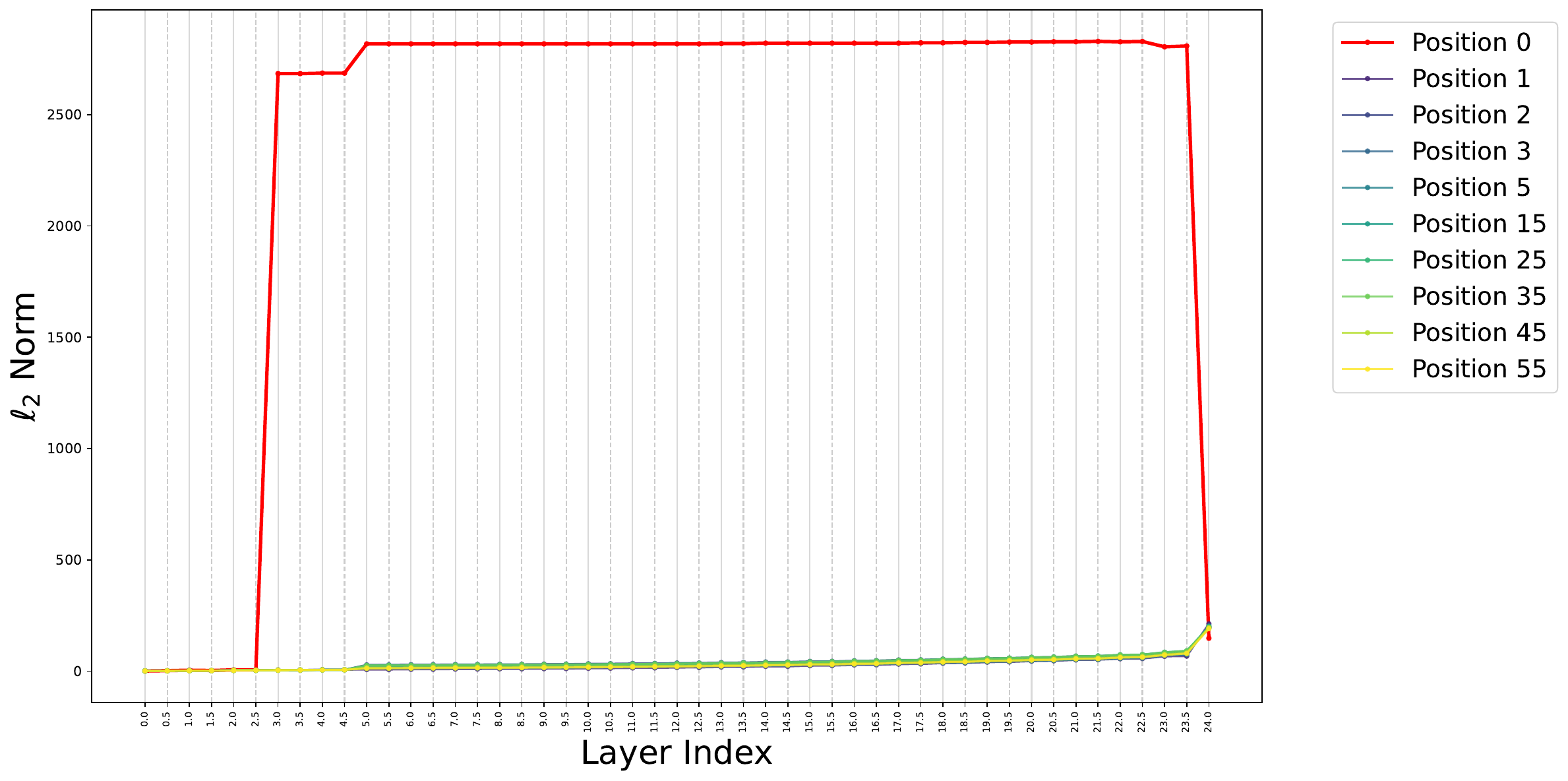}
        \end{subfigure}
    \end{minipage}
    \caption{Layer-wise $\ell_2$ norm of hidden states in baseline model. Half-integer indices correspond to attention module outputs after the residual connection. Metrics are computed over sequences of up to 64 tokens across 1024 validation samples.}
    \label{fig:C-baseline}
    \vspace{-1em}
\end{figure*}

\begin{figure*}[ht]
    \centering
    \begin{minipage}[b]{0.48\linewidth} 
        \centering
        \begin{subfigure}[b]{0.95\linewidth} 
            \centering
            \caption{\tiny{$\ell_2$ Norm of Activations, $lr=3\times10^4$ }}
            \includegraphics[width=0.95\columnwidth]{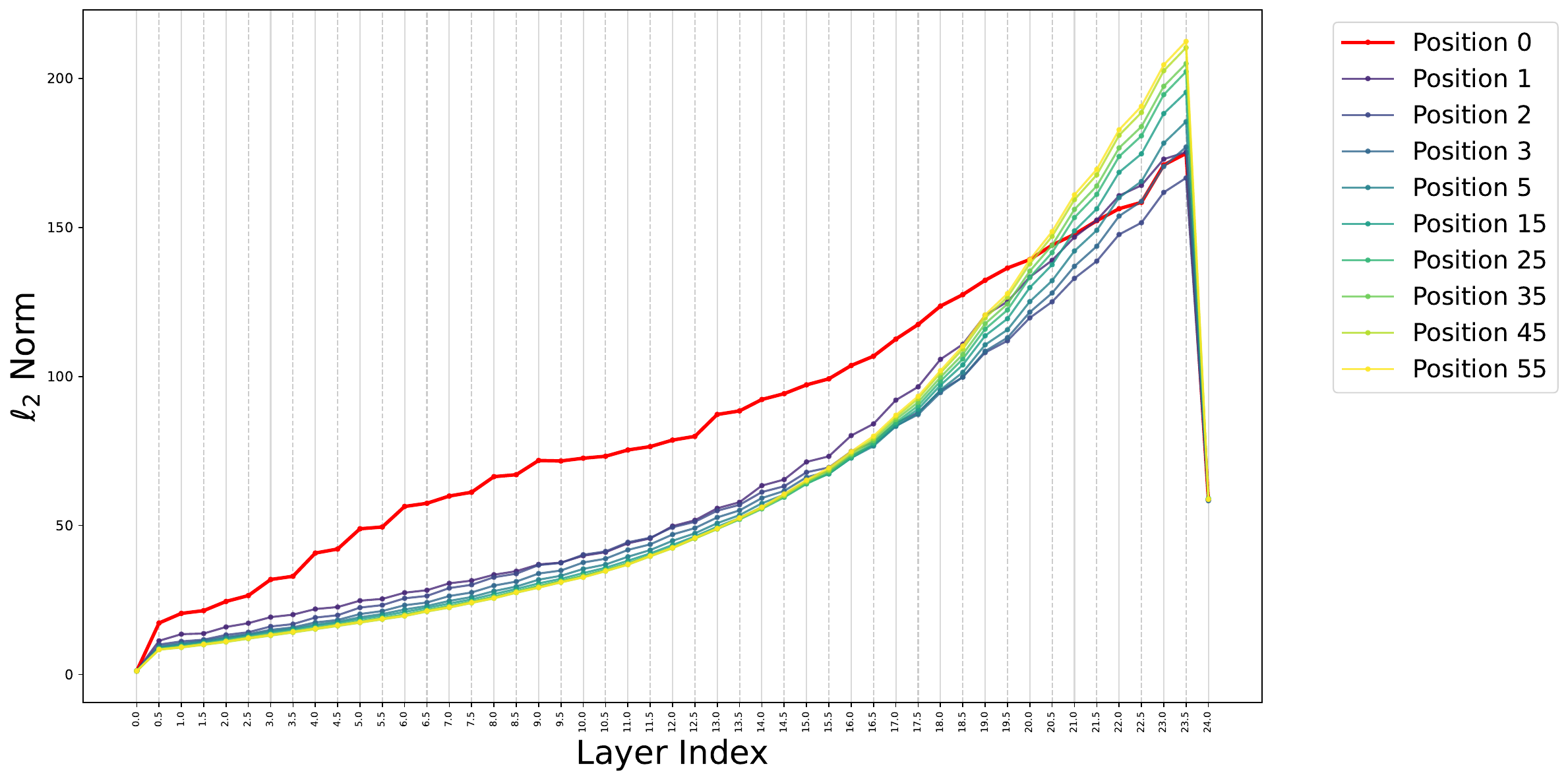}
        \end{subfigure}
    \end{minipage}
    \begin{minipage}[b]{0.48\linewidth} 
        \centering
        \begin{subfigure}[b]{0.95\linewidth} 
            \centering
            \caption{\tiny{$\ell_2$ Norm of Activations, $lr=4\times10^4$}}
            \includegraphics[width=0.95\columnwidth]{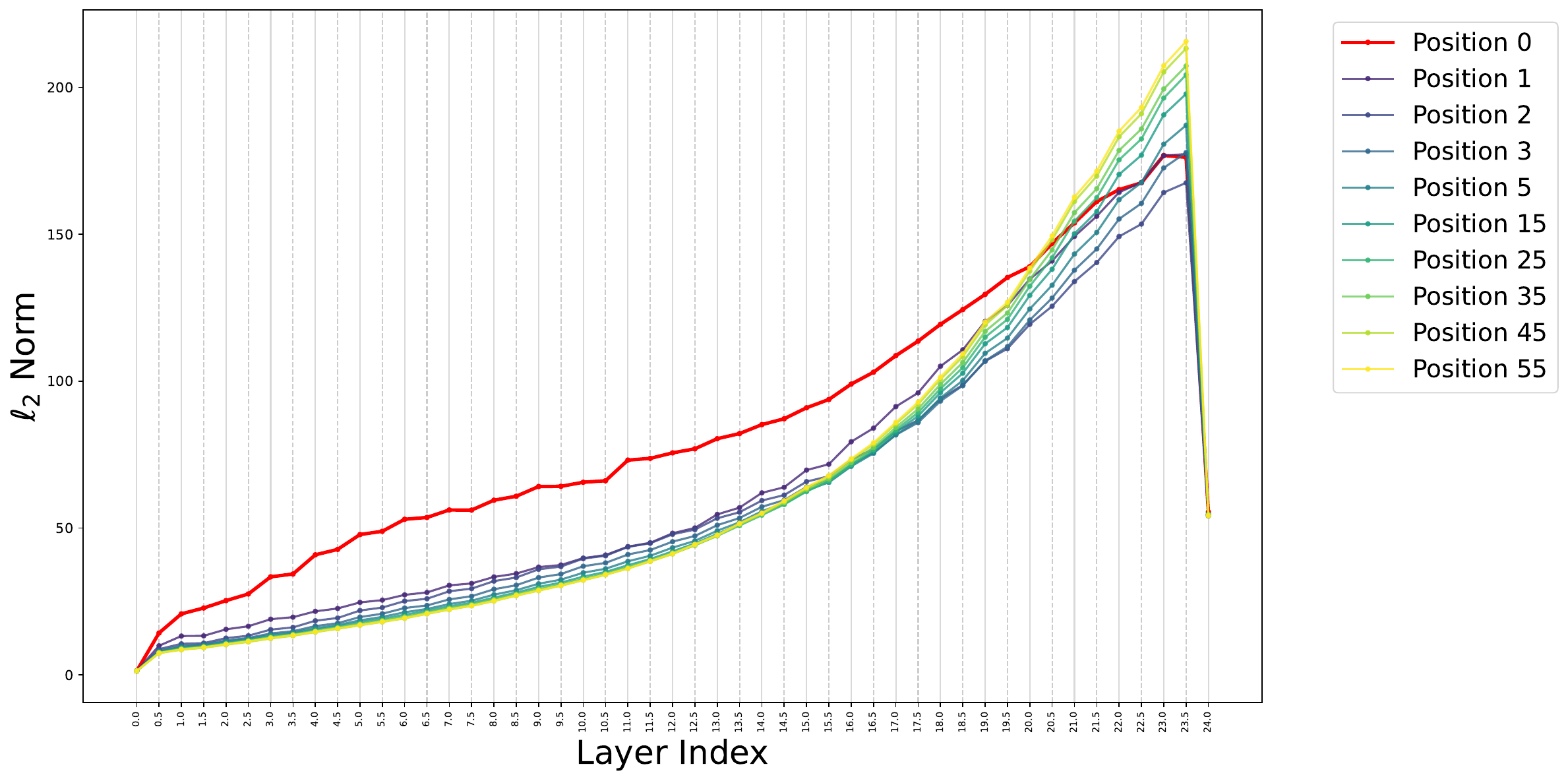}
        \end{subfigure}
    \end{minipage}
    \begin{minipage}[b]{0.48\linewidth} 
        \centering
        \begin{subfigure}[b]{0.95\linewidth} 
            \centering
            \caption{\tiny{$\ell_2$ Norm of Activations, $lr=5\times10^4$}}
            \includegraphics[width=0.95\columnwidth]{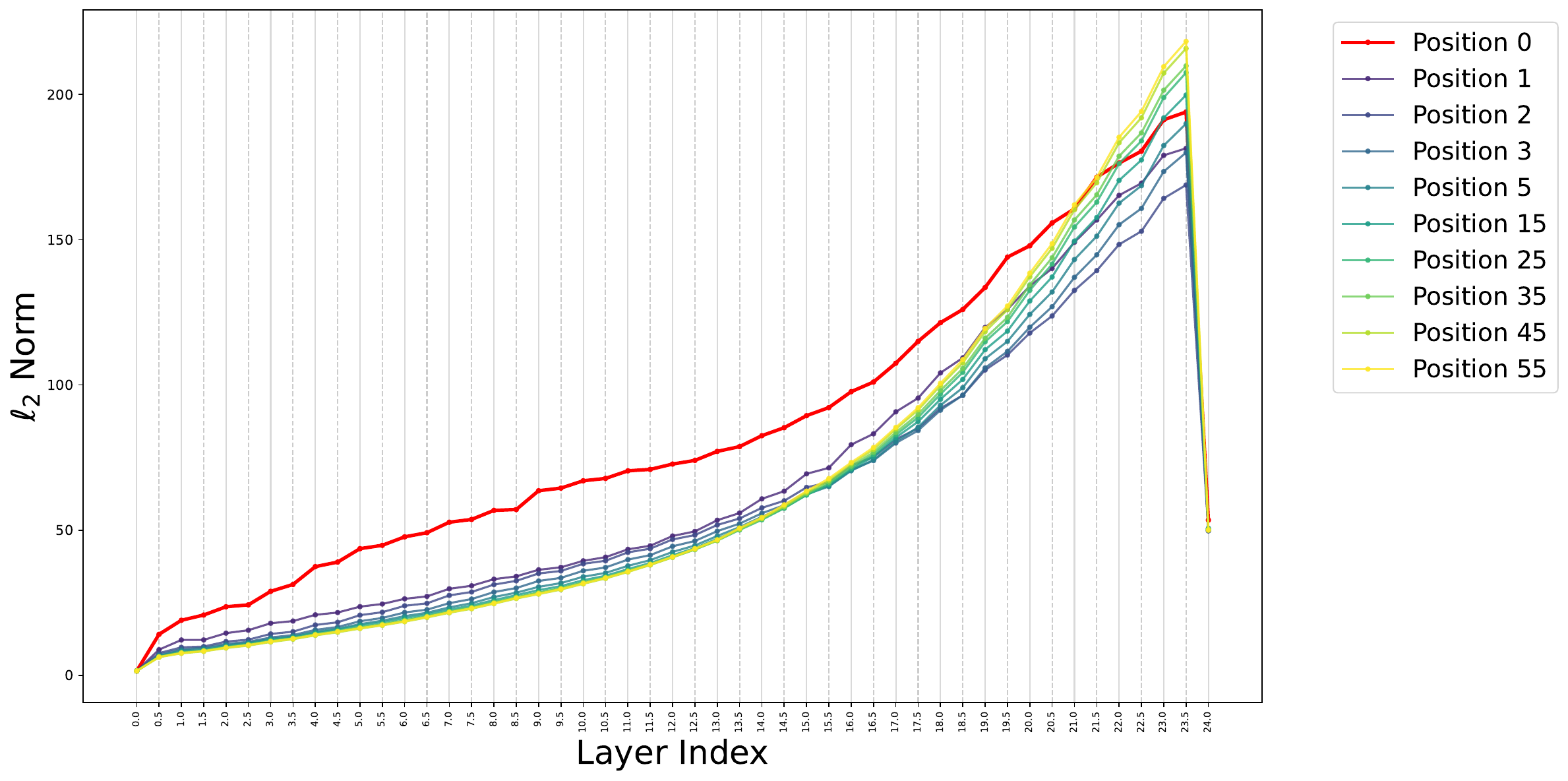}
        \end{subfigure}
    \end{minipage}
    \begin{minipage}[b]{0.48\linewidth} 
        \centering
        \begin{subfigure}[b]{0.95\linewidth} 
            \centering
            \caption{\tiny{$\ell_2$ Norm of Activations, data-saturated}}
            \includegraphics[width=0.95\columnwidth]{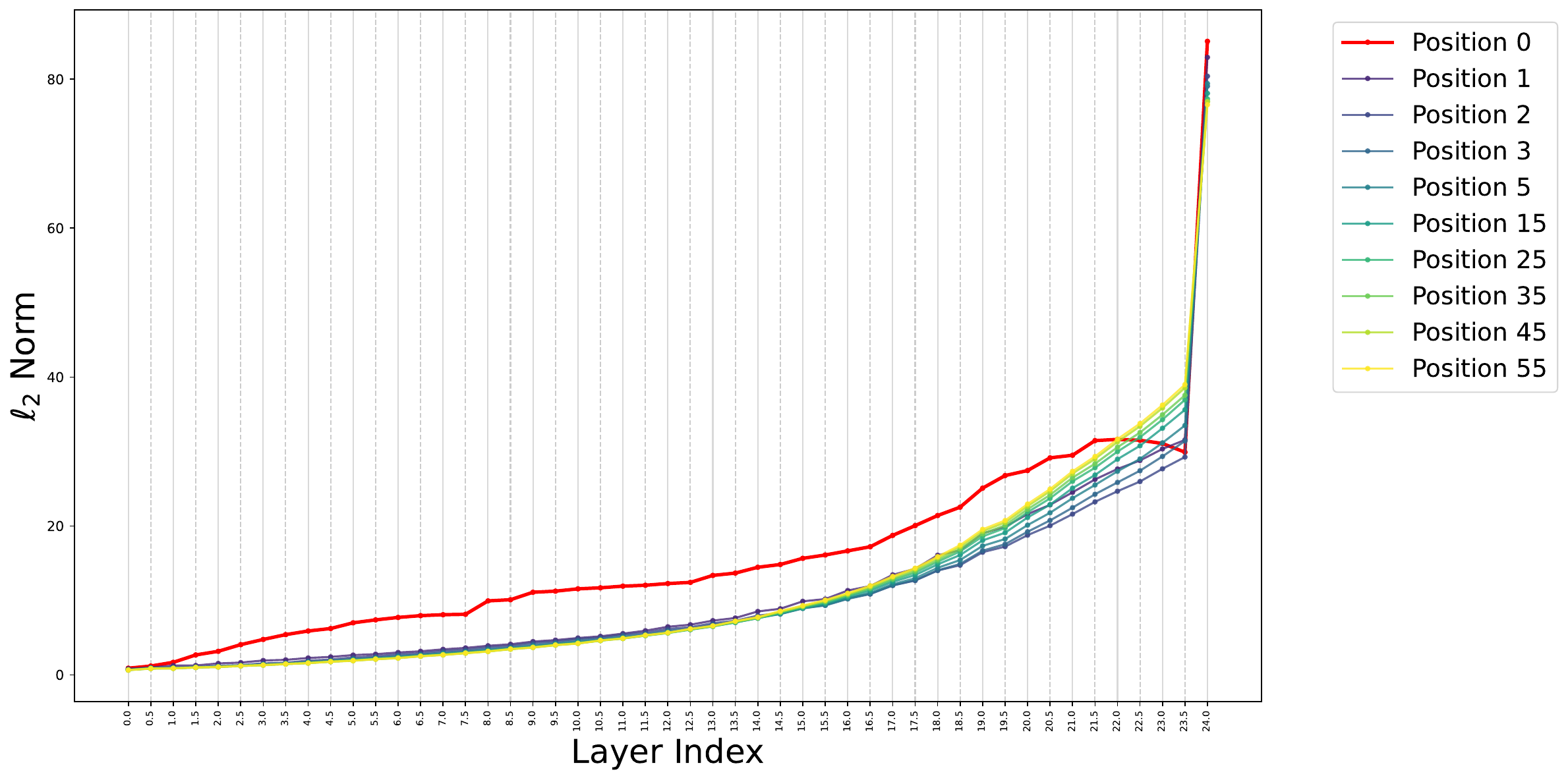}
        \end{subfigure}
    \end{minipage}
    \caption{Layer-wise $\ell_2$ norm of hidden states in gated attention model. Half-integer indices correspond to attention module outputs after the residual connection. Metrics are computed over sequences of up to 64 tokens across 1024 validation samples.}
    \label{fig:C-ga}
    \vspace{-1em}
\end{figure*}

\begin{figure*}[ht]
    \centering
    \begin{minipage}[b]{0.48\linewidth} 
        \centering
        \begin{subfigure}[b]{0.95\linewidth} 
            \centering
            \caption{\tiny{$\ell_2$ Norm of Activations, $lr=3\times10^4$ }}
            \includegraphics[width=0.95\columnwidth]{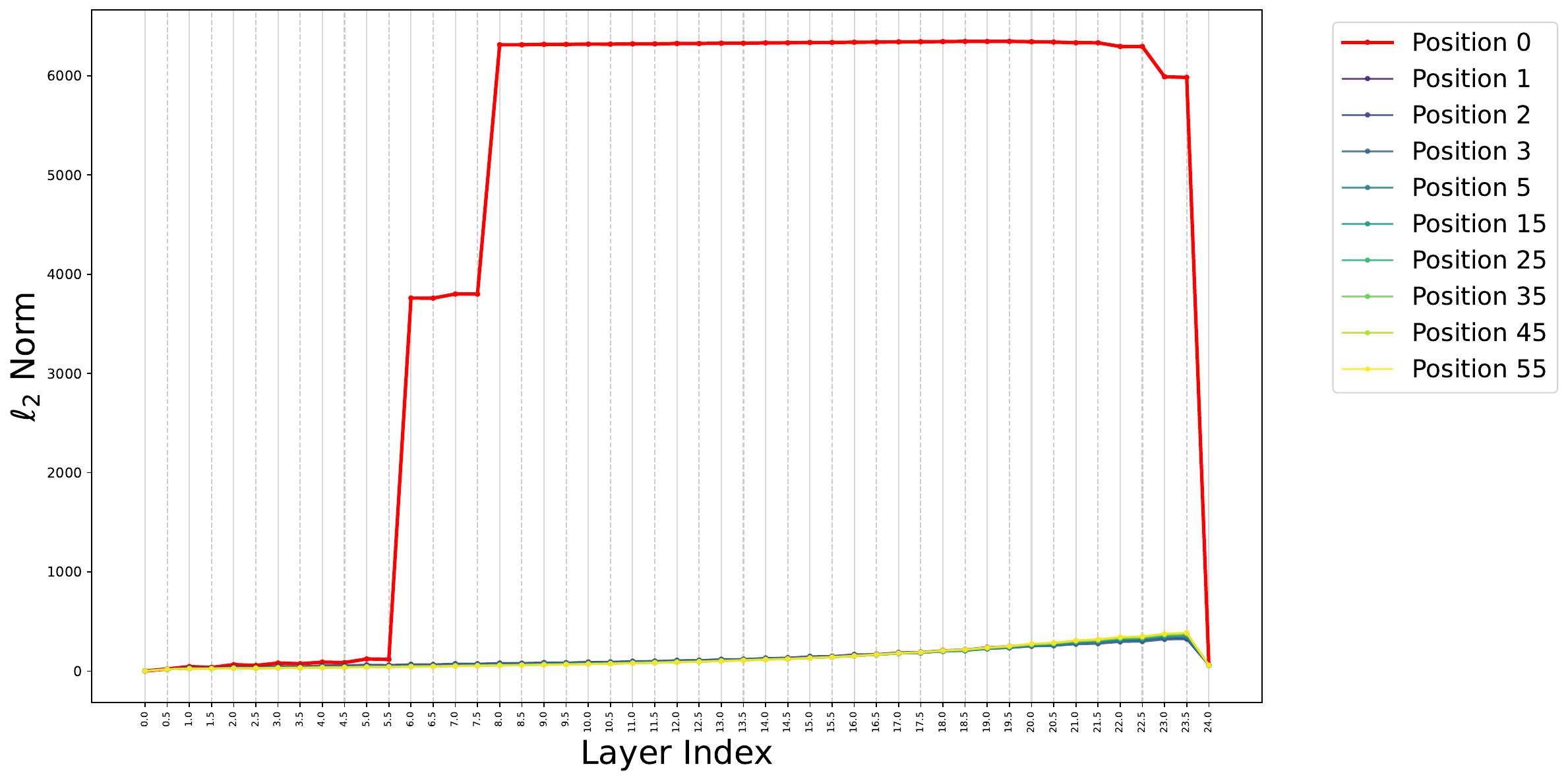}
        \end{subfigure}
    \end{minipage}
    \begin{minipage}[b]{0.48\linewidth} 
        \centering
        \begin{subfigure}[b]{0.95\linewidth} 
            \centering
            \caption{\tiny{$\ell_2$ Norm of Activations, $lr=4\times10^4$}}
            \includegraphics[width=0.95\columnwidth]{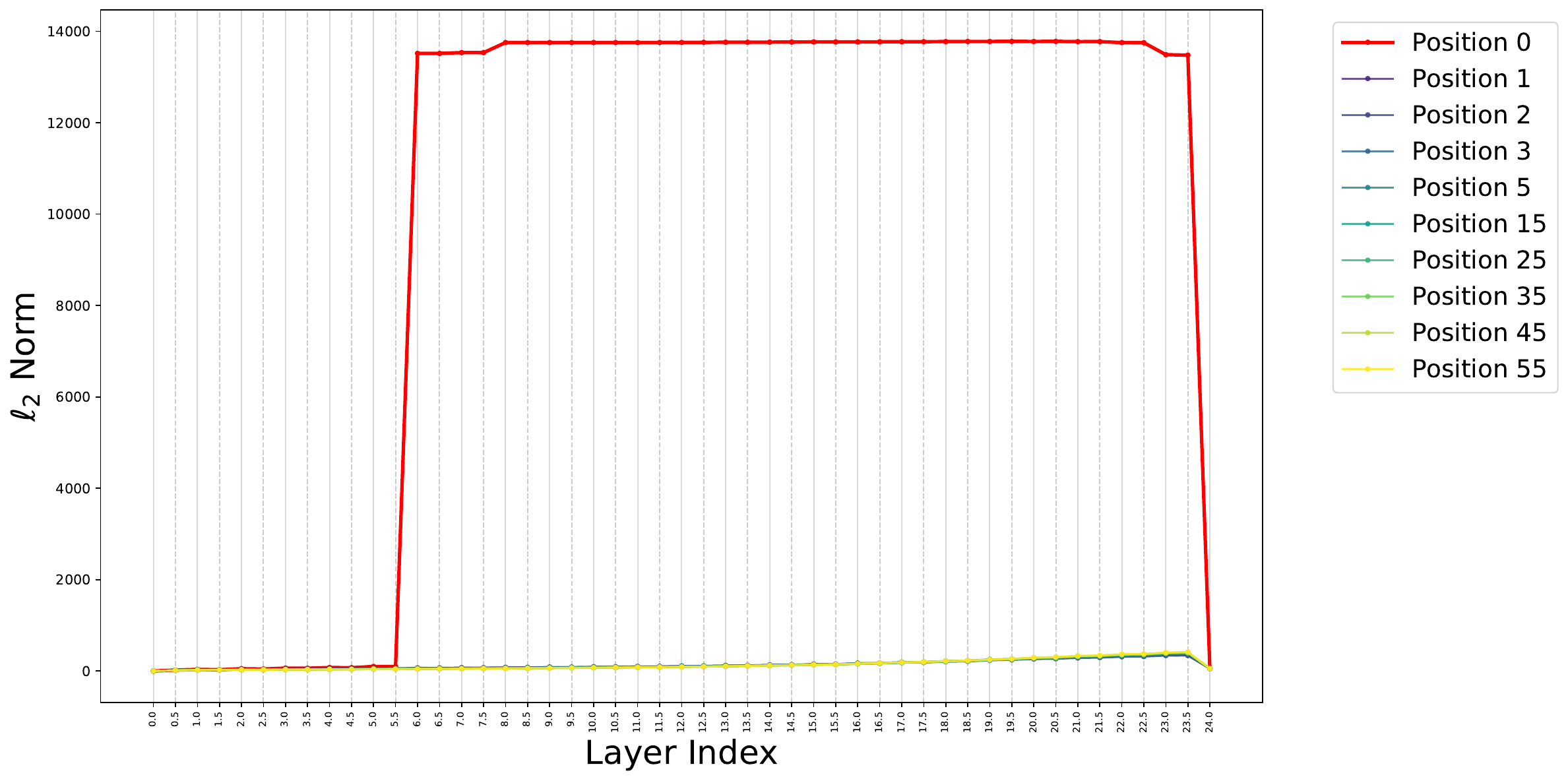}
        \end{subfigure}
    \end{minipage}
    \begin{minipage}[b]{0.48\linewidth} 
        \centering
        \begin{subfigure}[b]{0.95\linewidth} 
            \centering
            \caption{\tiny{$\ell_2$ Norm of Activations, $lr=5\times10^4$}}
            \includegraphics[width=0.95\columnwidth]{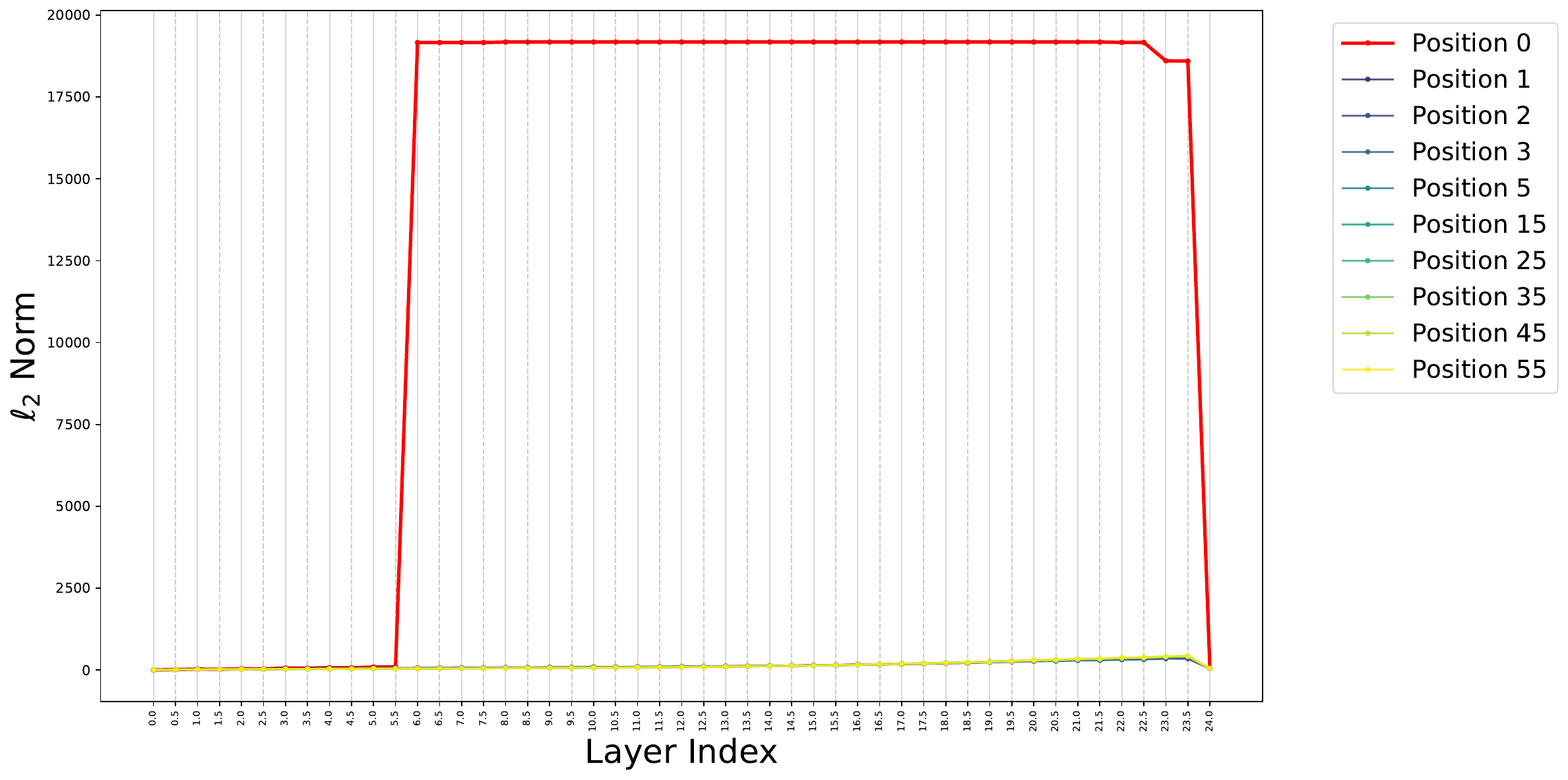}
        \end{subfigure}
    \end{minipage}
    \begin{minipage}[b]{0.48\linewidth} 
        \centering
        \begin{subfigure}[b]{0.95\linewidth} 
            \centering
            \caption{\tiny{$\ell_2$ Norm of Activations, data-saturated}}
            \includegraphics[width=0.95\columnwidth]{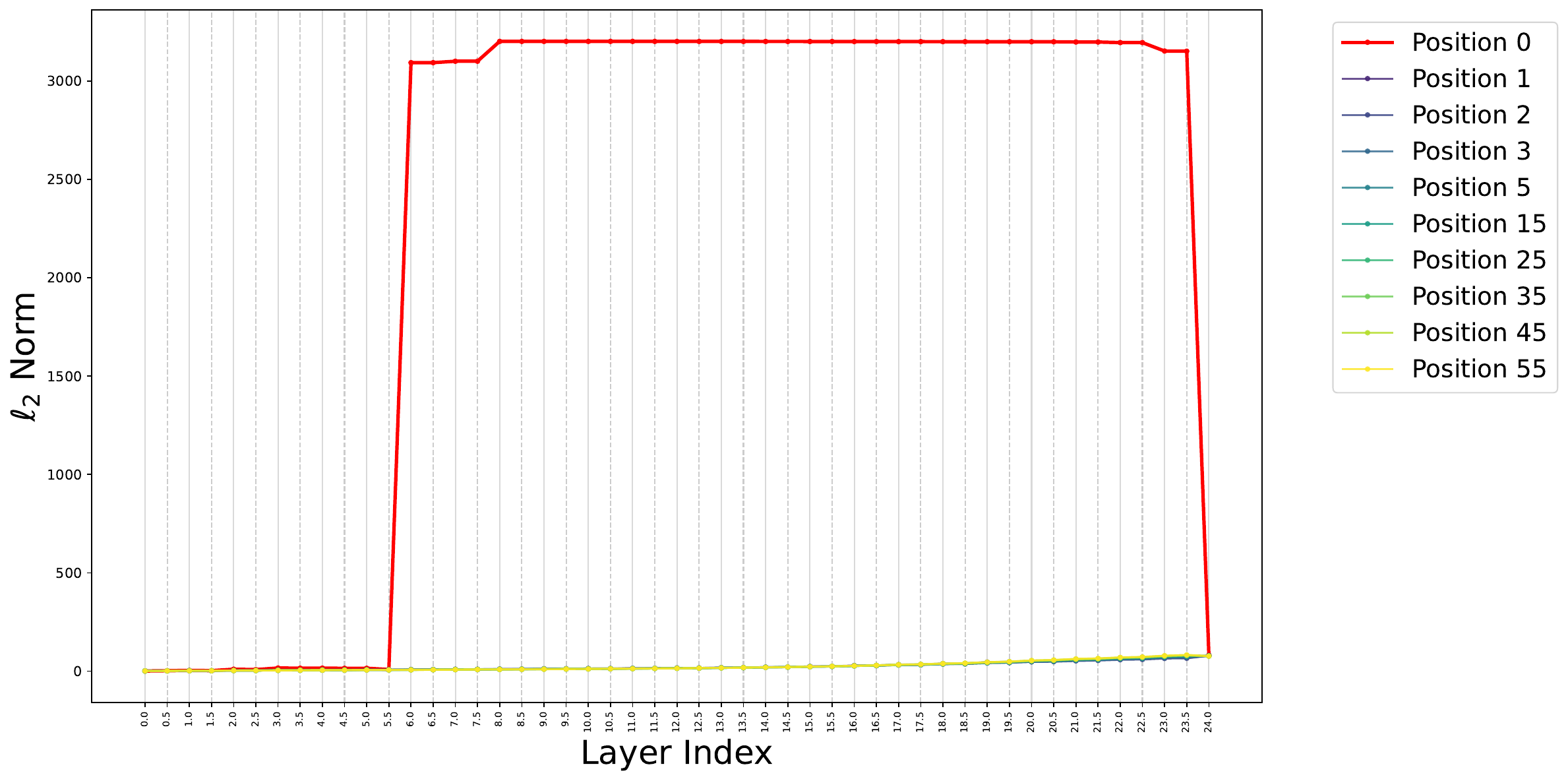}
        \end{subfigure}
    \end{minipage}
    \caption{Layer-wise $\ell_2$ norm of hidden states in transformer-native lm loss model. Half-integer indices correspond to attention module outputs after the residual connection. Metrics are computed over sequences of up to 64 tokens across 1024 validation samples.}
    \label{fig:C-tnlm}
    \vspace{-1em}
\end{figure*}

\begin{figure*}[ht]
    \centering
    \begin{minipage}[b]{0.48\linewidth} 
        \centering
        \begin{subfigure}[b]{0.95\linewidth} 
            \centering
            \caption{\tiny{$\ell_2$ Norm of Activations, $lr=3\times10^4$ }}
            \includegraphics[width=0.95\columnwidth]{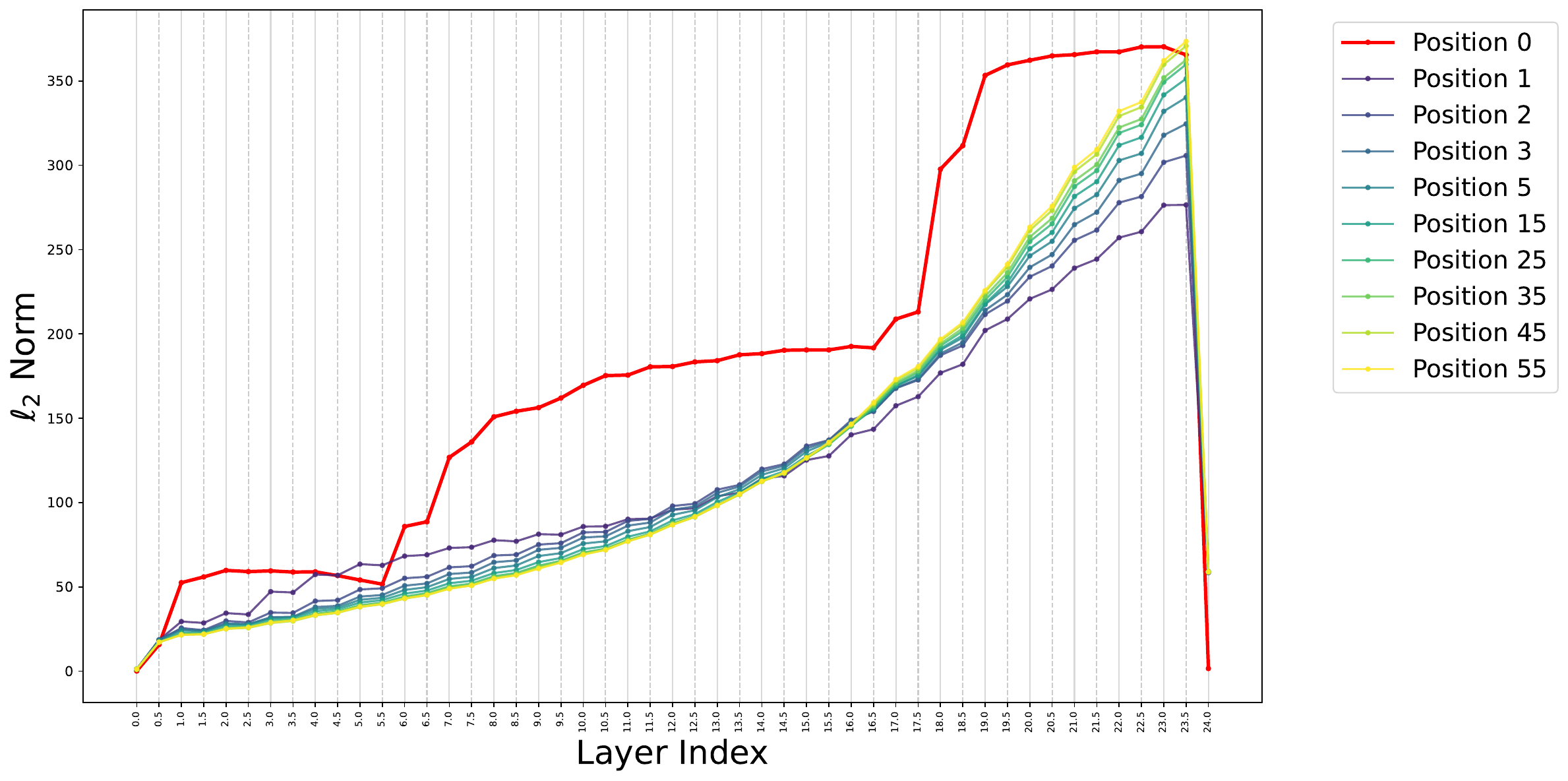}
        \end{subfigure}
    \end{minipage}
    \begin{minipage}[b]{0.48\linewidth} 
        \centering
        \begin{subfigure}[b]{0.95\linewidth} 
            \centering
            \caption{\tiny{$\ell_2$ Norm of Activations, $lr=4\times10^4$}}
            \includegraphics[width=0.95\columnwidth]{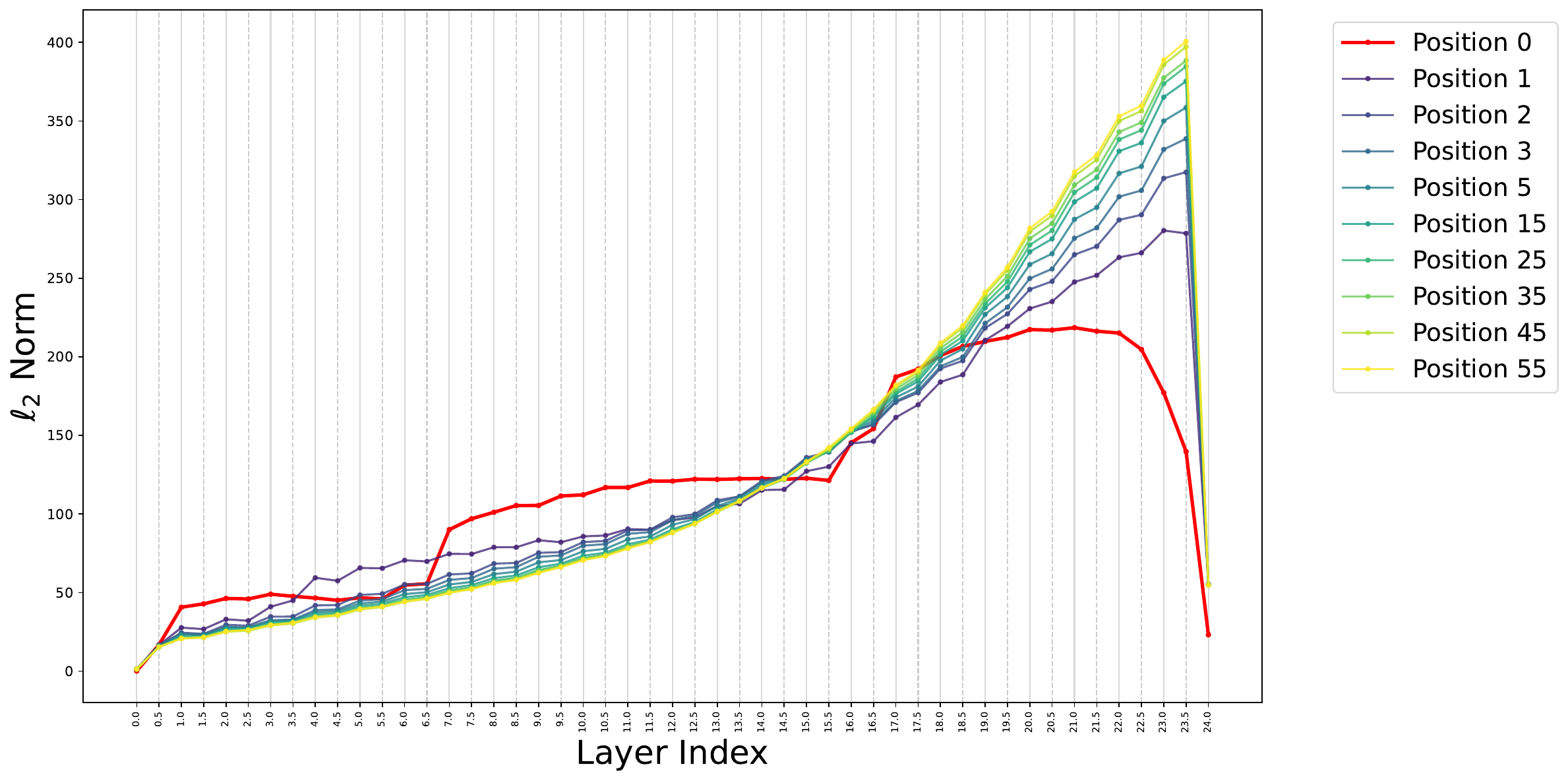}
        \end{subfigure}
    \end{minipage}
    \begin{minipage}[b]{0.48\linewidth} 
        \centering
        \begin{subfigure}[b]{0.95\linewidth} 
            \centering
            \caption{\tiny{$\ell_2$ Norm of Activations, $lr=5\times10^4$}}
            \includegraphics[width=0.95\columnwidth]{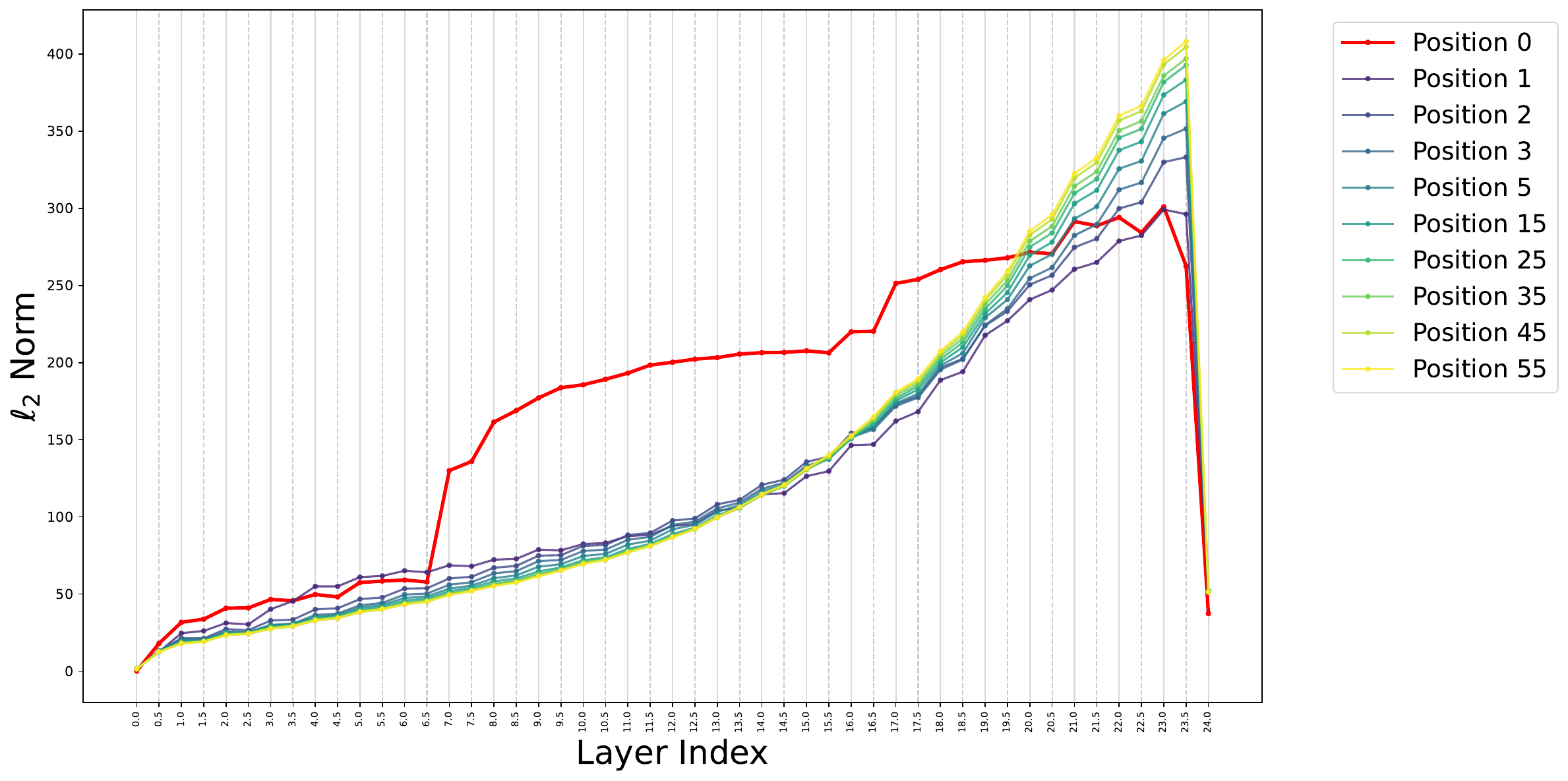}
        \end{subfigure}
    \end{minipage}
    \begin{minipage}[b]{0.48\linewidth} 
        \centering
        \begin{subfigure}[b]{0.95\linewidth} 
            \centering
            \caption{\tiny{$\ell_2$ Norm of Activations, data-saturated}}
            \includegraphics[width=0.95\columnwidth]{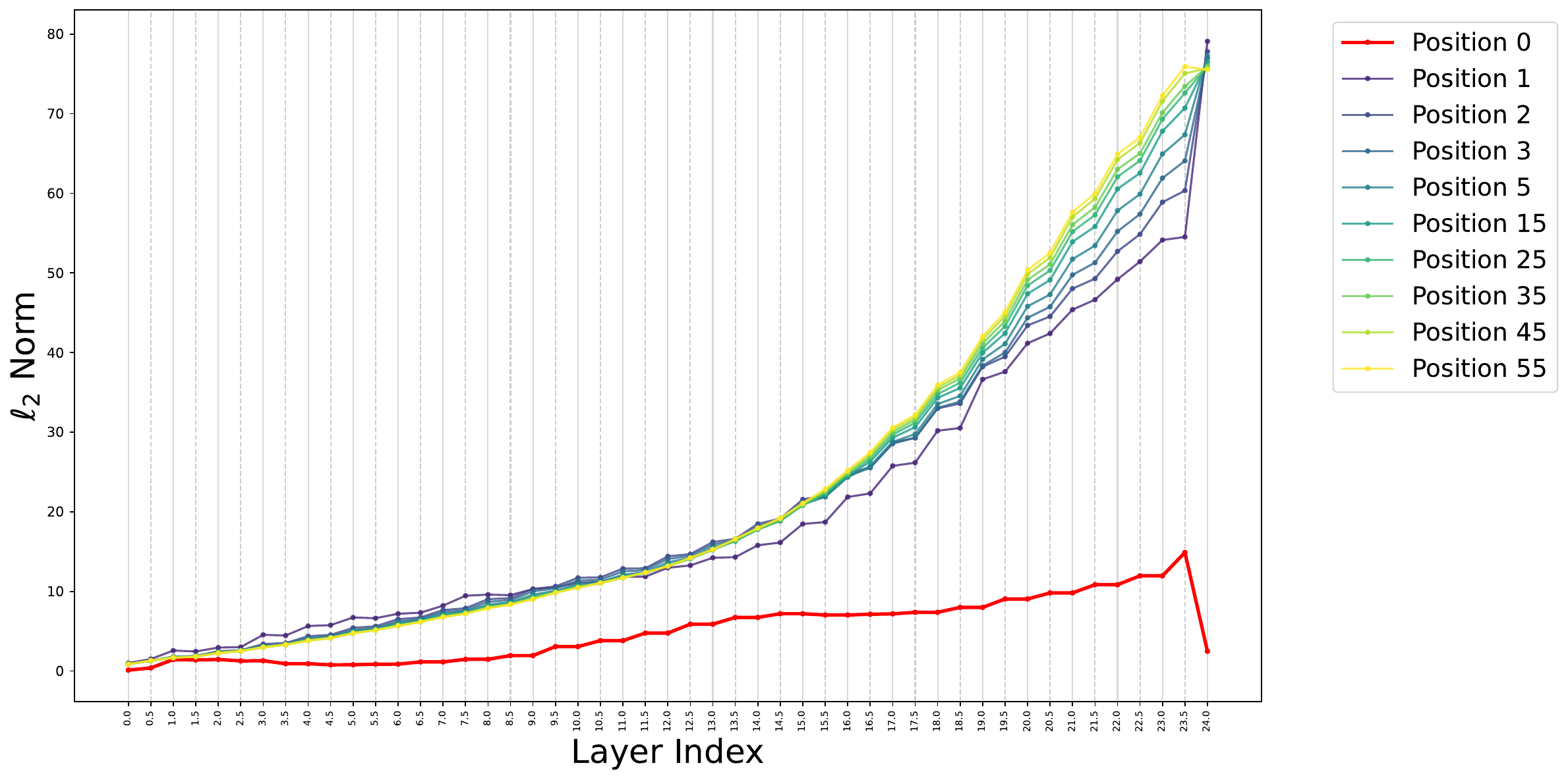}
        \end{subfigure}
    \end{minipage}
    \caption{Layer-wise $\ell_2$ norm of hidden states in force-sink masking model. Half-integer indices correspond to attention module outputs after the residual connection. Metrics are computed over sequences of up to 64 tokens across 1024 validation samples.}
    \label{fig:C-fsm}
    \vspace{-1em}
\end{figure*}

\begin{figure*}[ht]
    \centering
    \begin{minipage}[b]{0.48\linewidth} 
        \centering
        \begin{subfigure}[b]{0.95\linewidth} 
            \centering
            \caption{\tiny{$\ell_2$ Norm of Activations, $lr=3\times10^4$ }}
            \includegraphics[width=0.95\columnwidth]{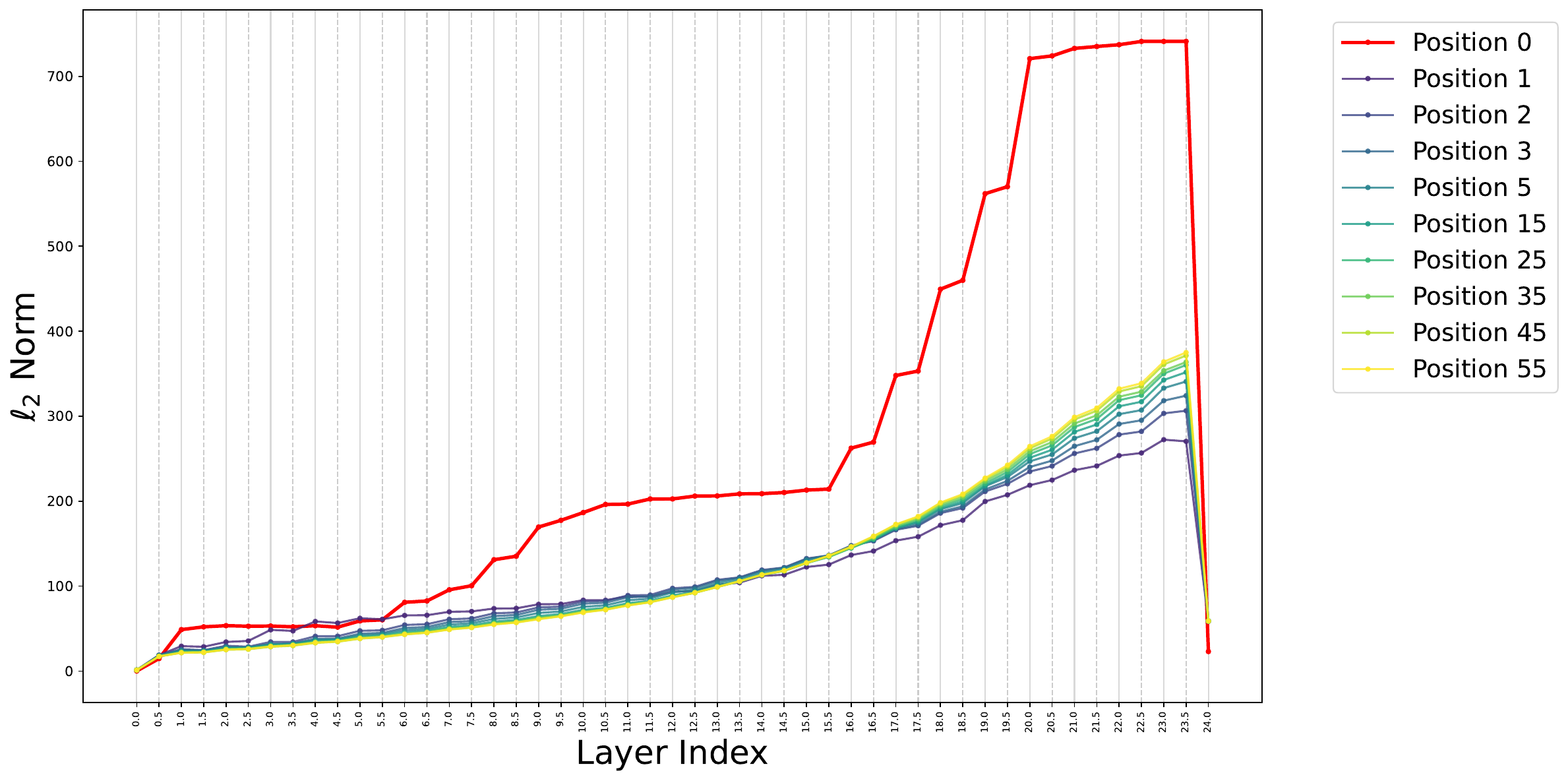}
        \end{subfigure}
    \end{minipage}
    \begin{minipage}[b]{0.48\linewidth} 
        \centering
        \begin{subfigure}[b]{0.95\linewidth} 
            \centering
            \caption{\tiny{$\ell_2$ Norm of Activations, $lr=4\times10^4$}}
            \includegraphics[width=0.95\columnwidth]{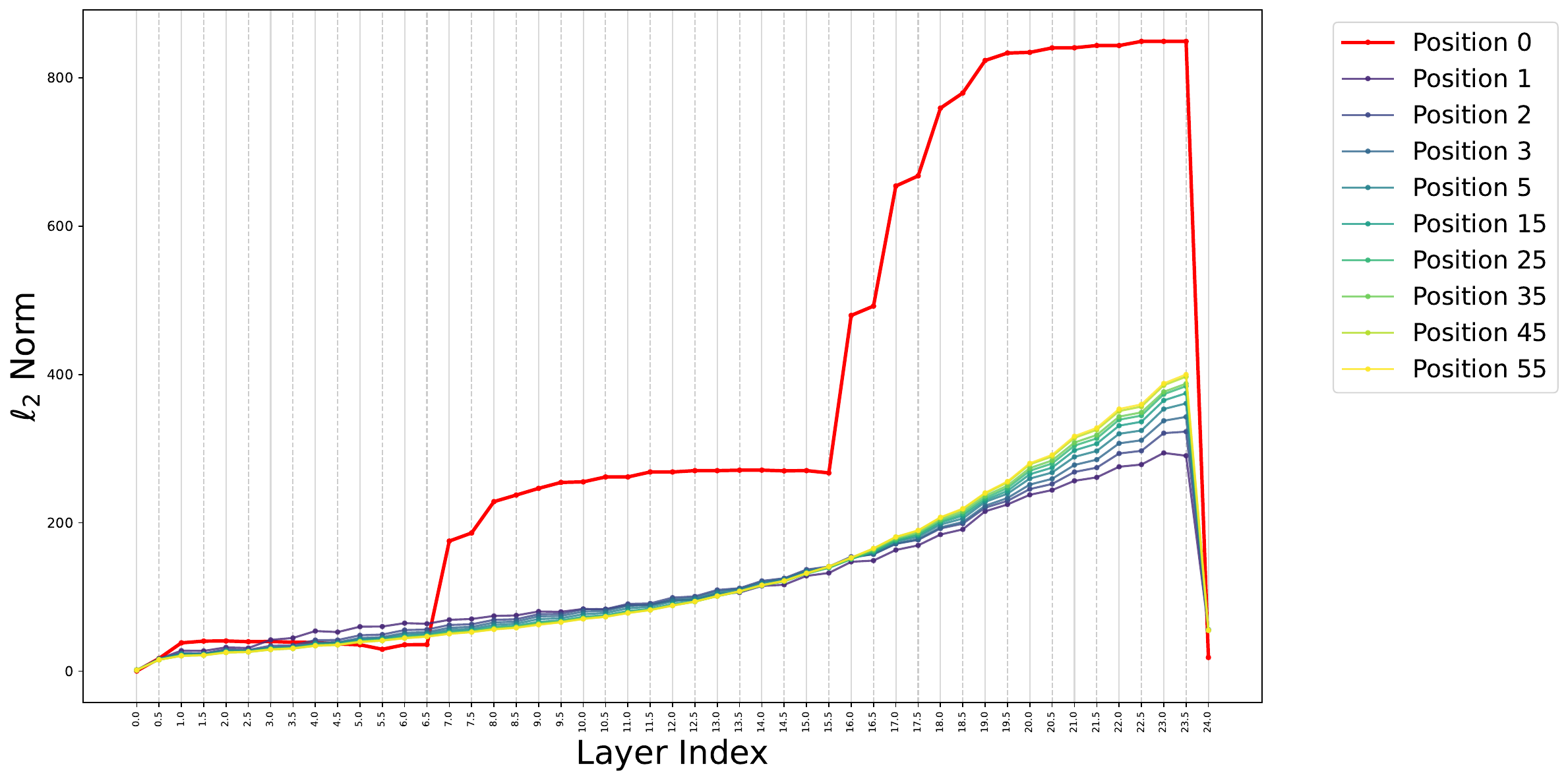}
        \end{subfigure}
    \end{minipage}
    \begin{minipage}[b]{0.48\linewidth} 
        \centering
        \begin{subfigure}[b]{0.95\linewidth} 
            \centering
            \caption{\tiny{$\ell_2$ Norm of Activations, $lr=5\times10^4$}}
            \includegraphics[width=0.95\columnwidth]{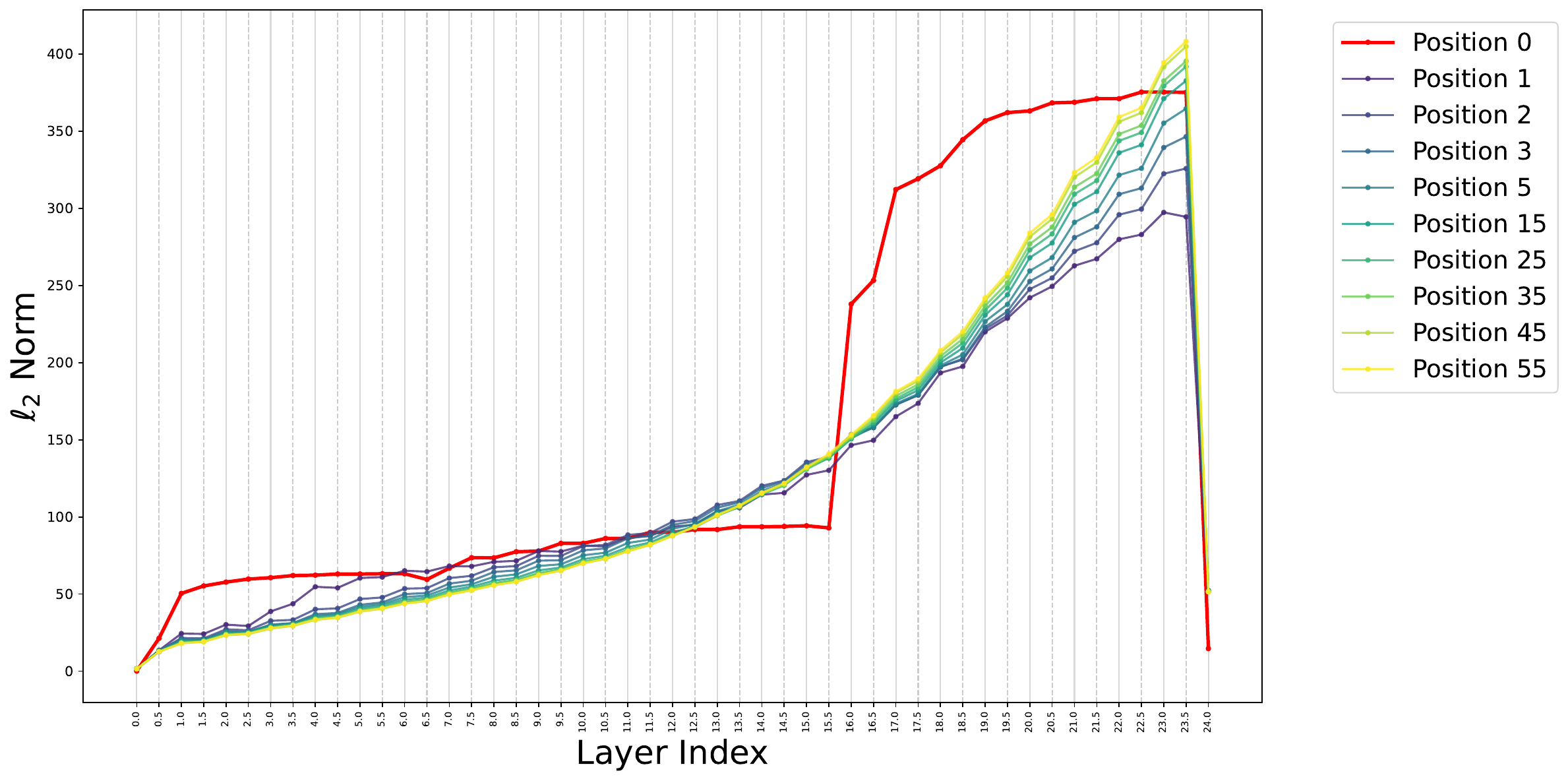}
        \end{subfigure}
    \end{minipage}
    \begin{minipage}[b]{0.48\linewidth} 
        \centering
        \begin{subfigure}[b]{0.95\linewidth} 
            \centering
            \caption{\tiny{$\ell_2$ Norm of Activations, data-saturated}}
            \includegraphics[width=0.95\columnwidth]{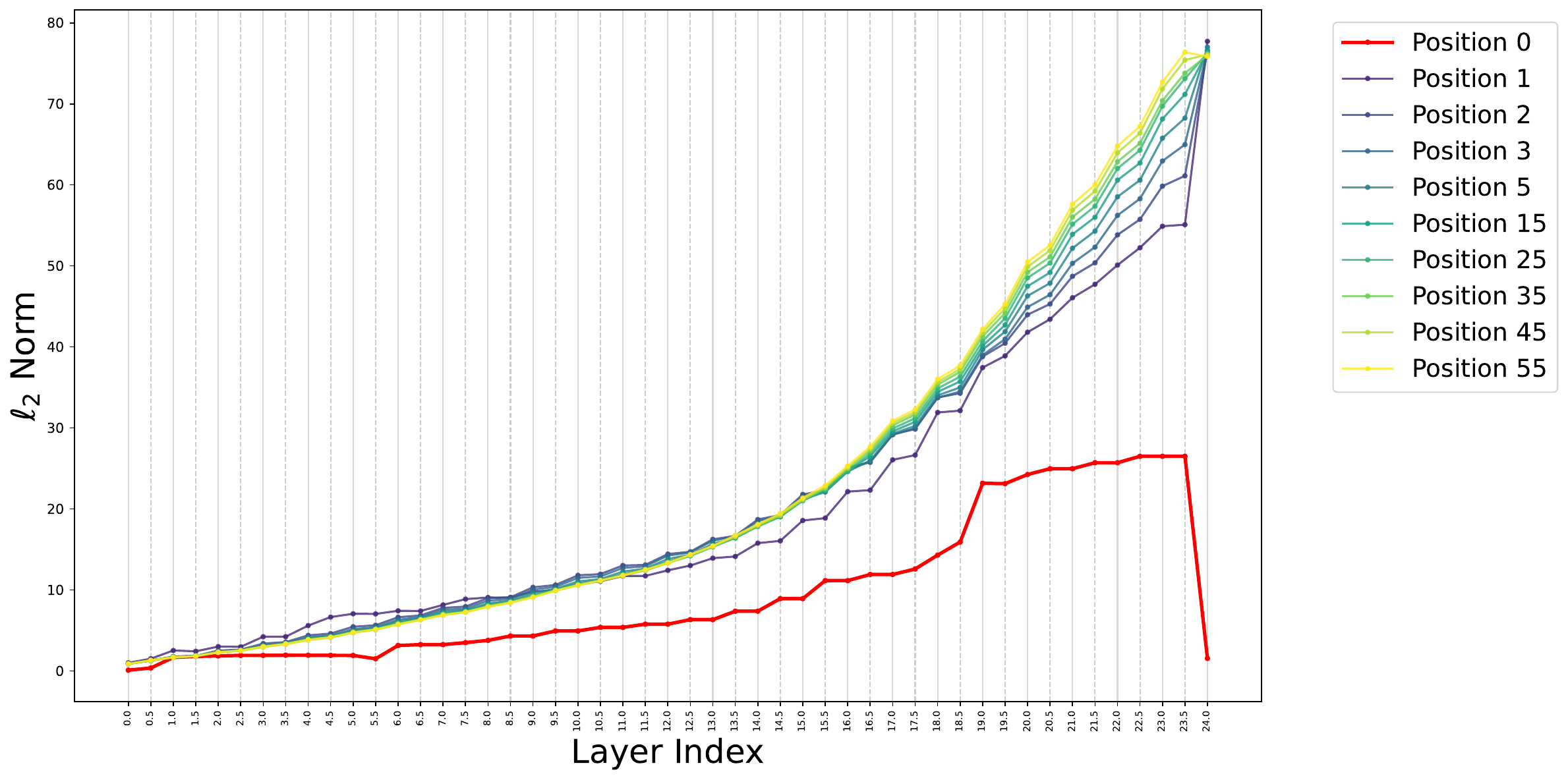}
        \end{subfigure}
    \end{minipage}
    \caption{Layer-wise $\ell_2$ norm of hidden states in force-sink masking + transformer-native lm loss model. Half-integer indices correspond to attention module outputs after the residual connection. Metrics are computed over sequences of up to 64 tokens across 1024 validation samples.}
    \label{fig:C-fsm+tnlm}
    \vspace{-1em}
\end{figure*}

\FloatBarrier

\section{Sink Rate Analysis for Main Experiment}
\label{appendix:sink-rate-on-main-exp}

\begin{figure}[ht]
    \centering
    \begin{subfigure}[c]{0.38\linewidth} 
        \centering
        \caption{\tiny{Qwen3-4B: $\ell_2$ Norm}}
        \includegraphics[width=\linewidth]{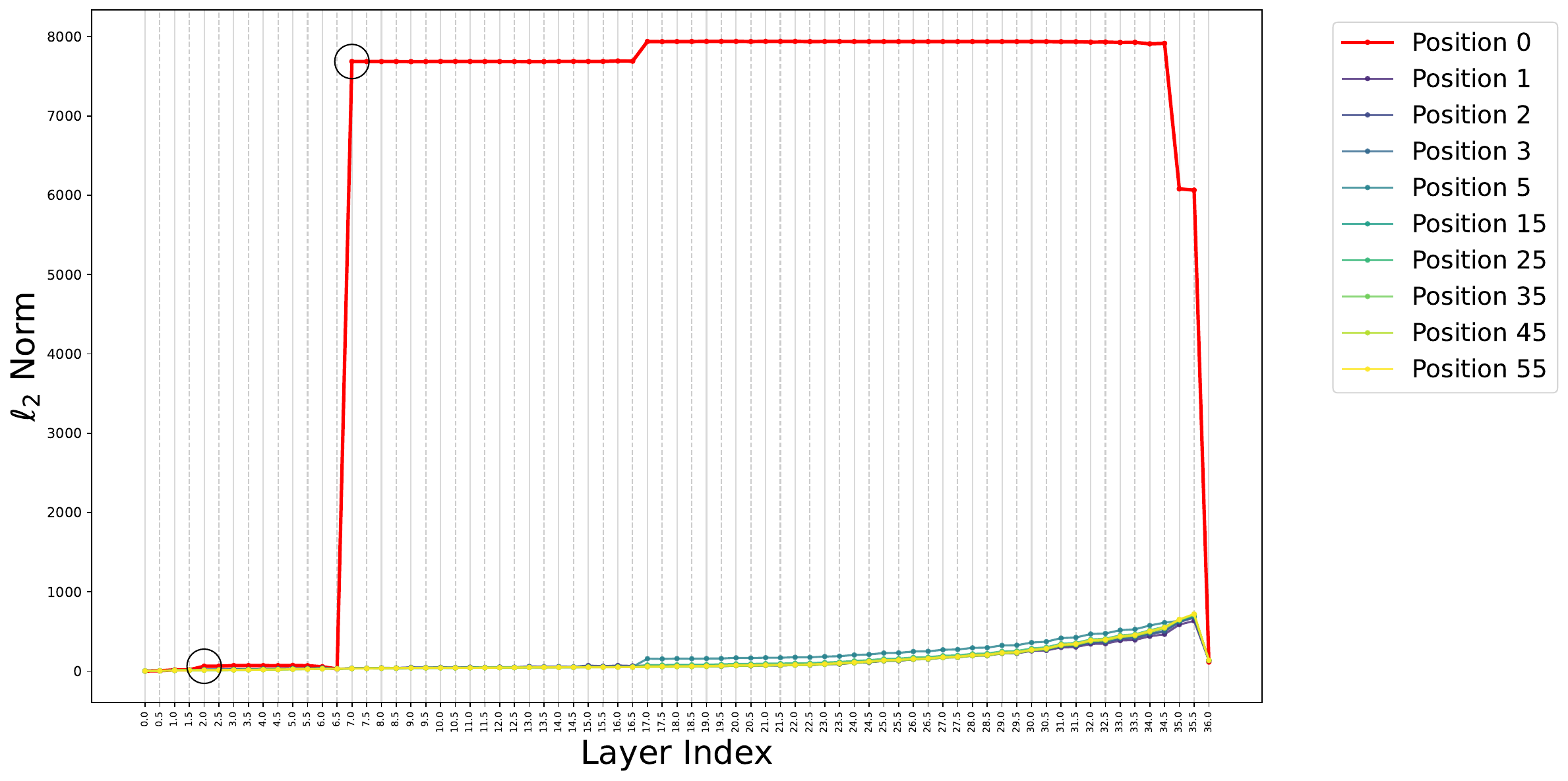}
    \end{subfigure}%
    \hspace{1.5em} 
    \begin{subfigure}[c]{0.26\linewidth} 
        \centering
        \caption{\tiny{Qwen3-4B: Layer 6 Score}} 
        \includegraphics[width=\linewidth]{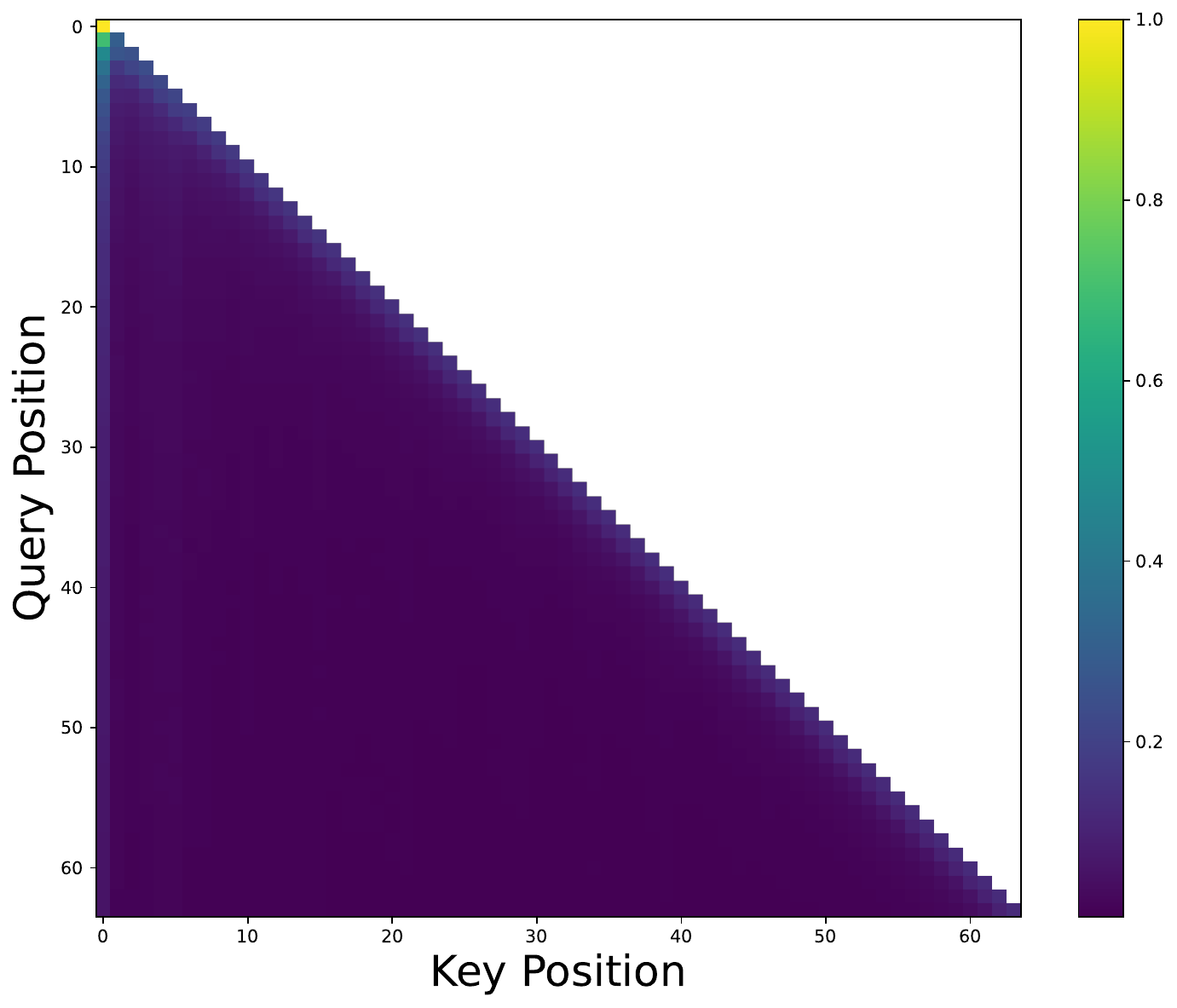}
    \end{subfigure}
    \begin{subfigure}[c]{0.26\linewidth} 
        \centering
        \caption{\tiny{Qwen3-4B: Layer 7 Score}} 
        \includegraphics[width=\linewidth]{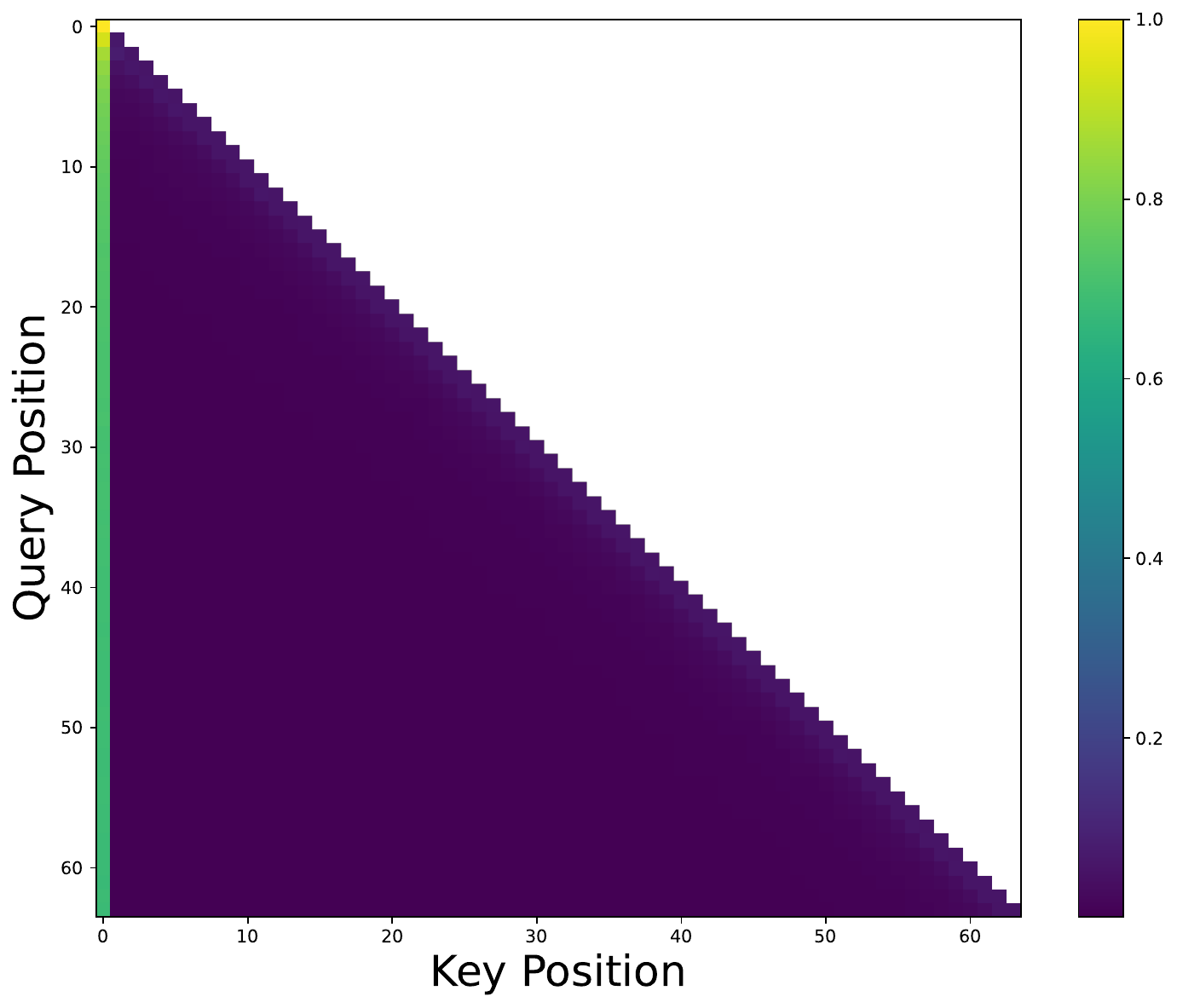}
    \end{subfigure}
    \caption{Layer-wise $\ell_2$ norm and attention sink visualization for Qwen3-4B. The attention sink becomes prominent after layer 7, coinciding with the sharp increase in the $\ell_2$ norm at position zero (P0).}
    \label{fig:appedixD-analysis-Qwen3-4B}
\end{figure}

\begin{figure*}[ht]
    \centering
    \begin{minipage}[b]{0.48\linewidth} 
        \centering
        \begin{subfigure}[b]{0.95\linewidth} 
            \centering
            \caption{\tiny{$\ell_2$ Norm of Activations, $lr=3\times10^4$ }}
            \includegraphics[width=0.95\columnwidth]{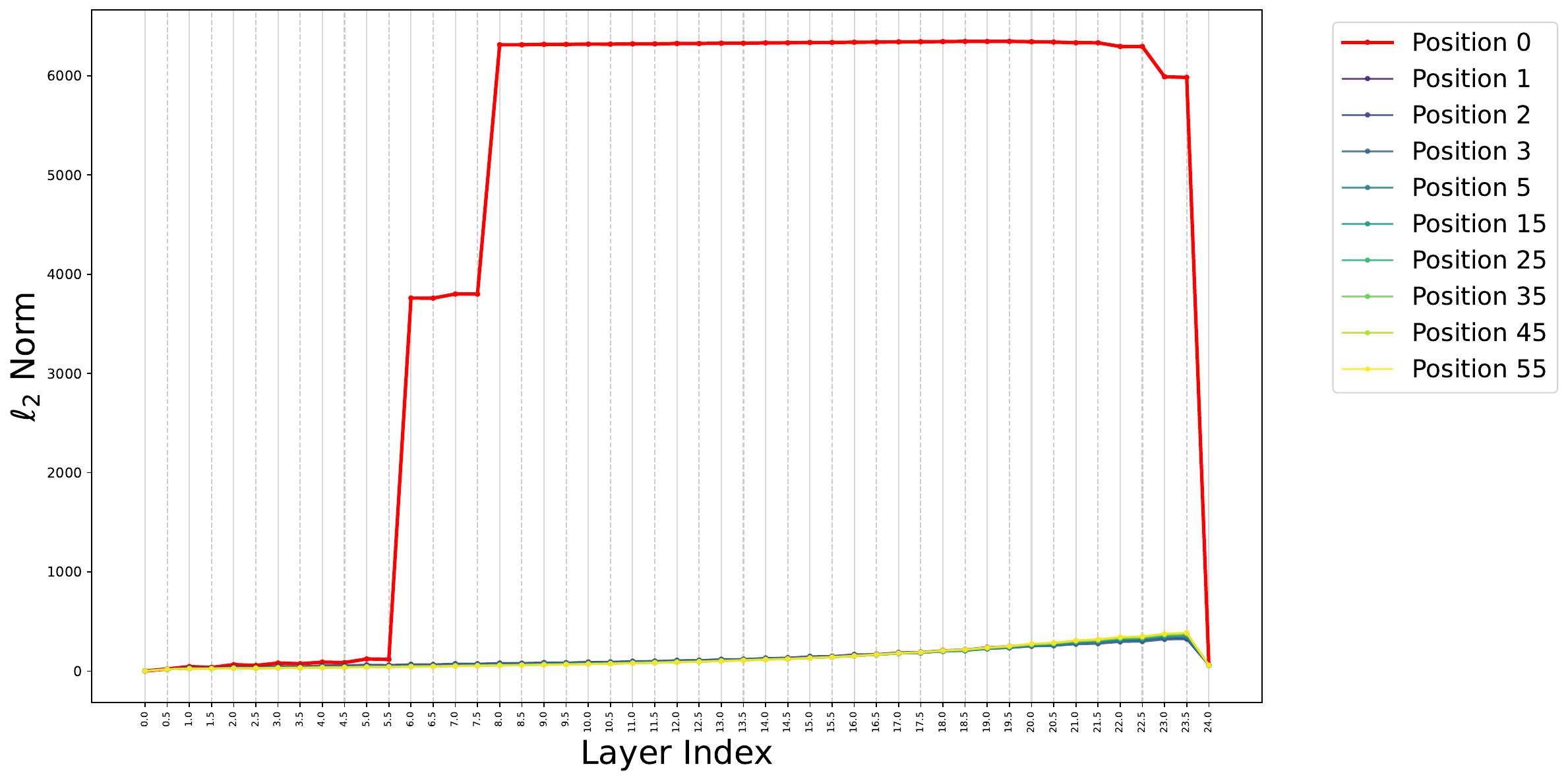}
        \end{subfigure}
    \end{minipage}
    \begin{minipage}[b]{0.48\linewidth} 
        \centering
        \begin{subfigure}[b]{0.95\linewidth} 
            \centering
            \caption{\tiny{$\ell_2$ Norm of Activations, $lr=4\times10^4$}}
            \includegraphics[width=0.95\columnwidth]{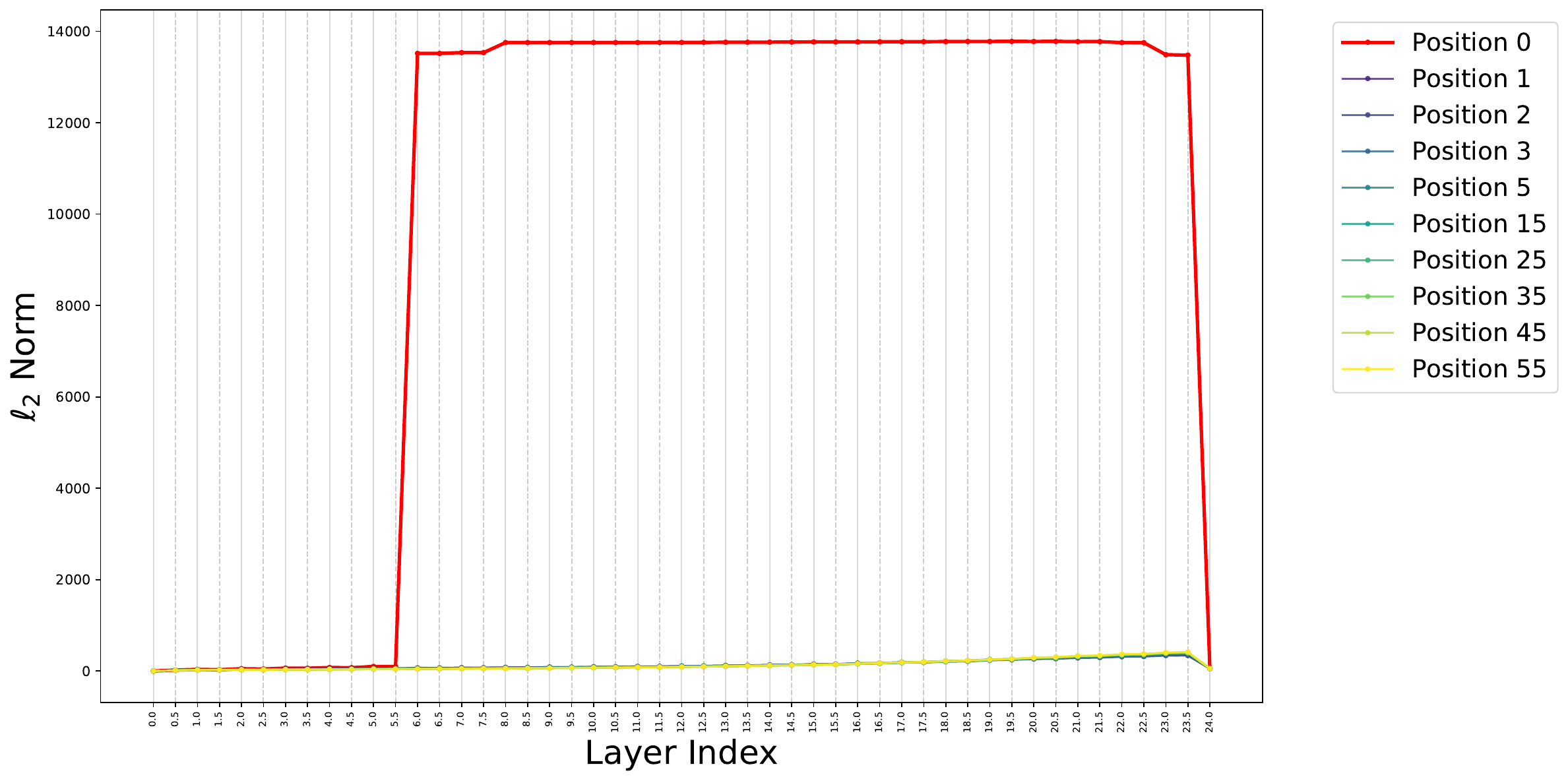}
        \end{subfigure}
    \end{minipage}
    \begin{minipage}[b]{0.48\linewidth} 
        \centering
        \begin{subfigure}[b]{0.95\linewidth} 
            \centering
            \caption{\tiny{$\ell_2$ Norm of Activations, $lr=5\times10^4$}}
            \includegraphics[width=0.95\columnwidth]{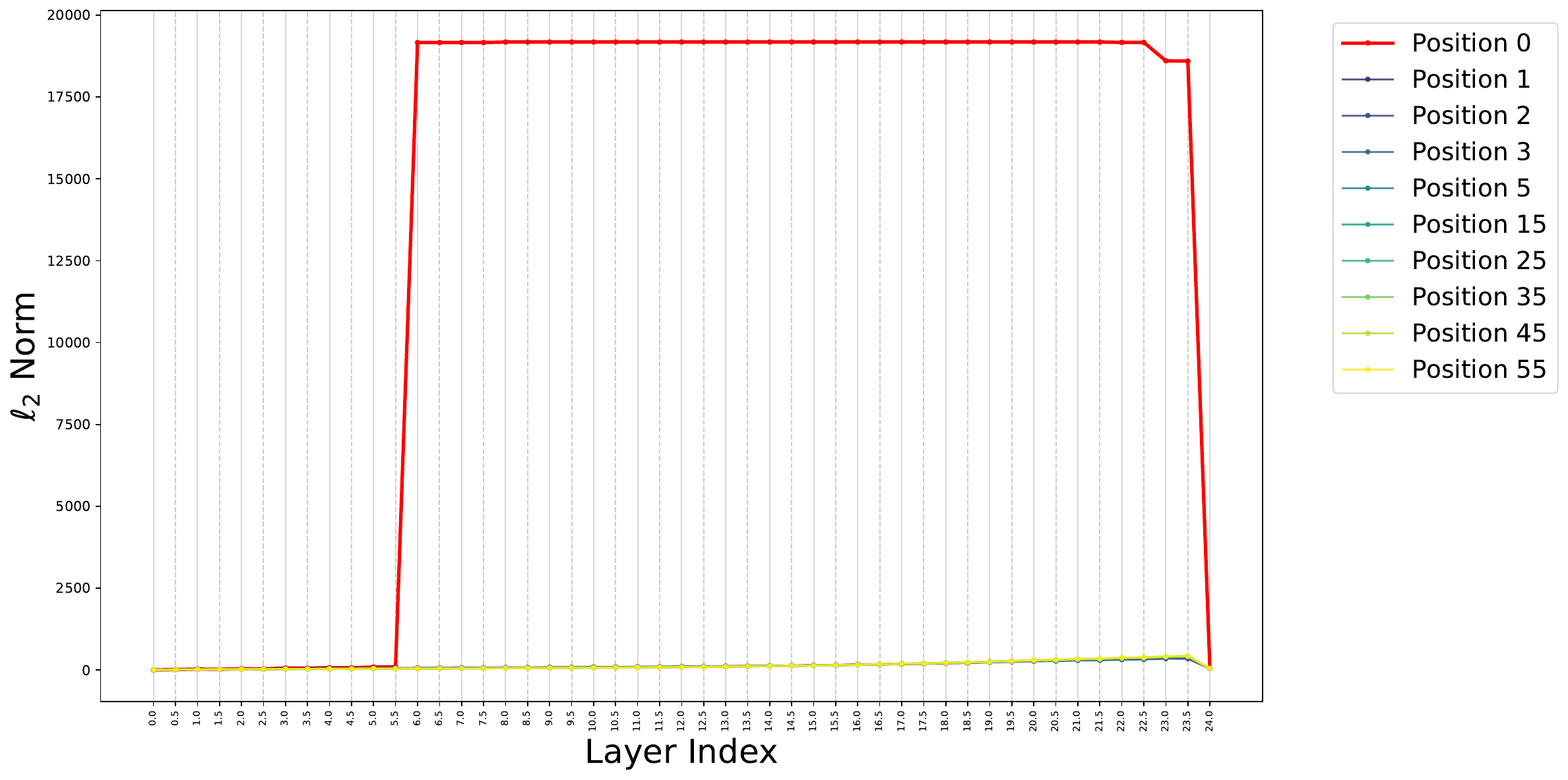}
        \end{subfigure}
    \end{minipage}
    \begin{minipage}[b]{0.48\linewidth} 
        \centering
        \begin{subfigure}[b]{0.95\linewidth} 
            \centering
            \caption{\tiny{$\ell_2$ Norm of Activations, data-saturated}}
            \includegraphics[width=0.95\columnwidth]{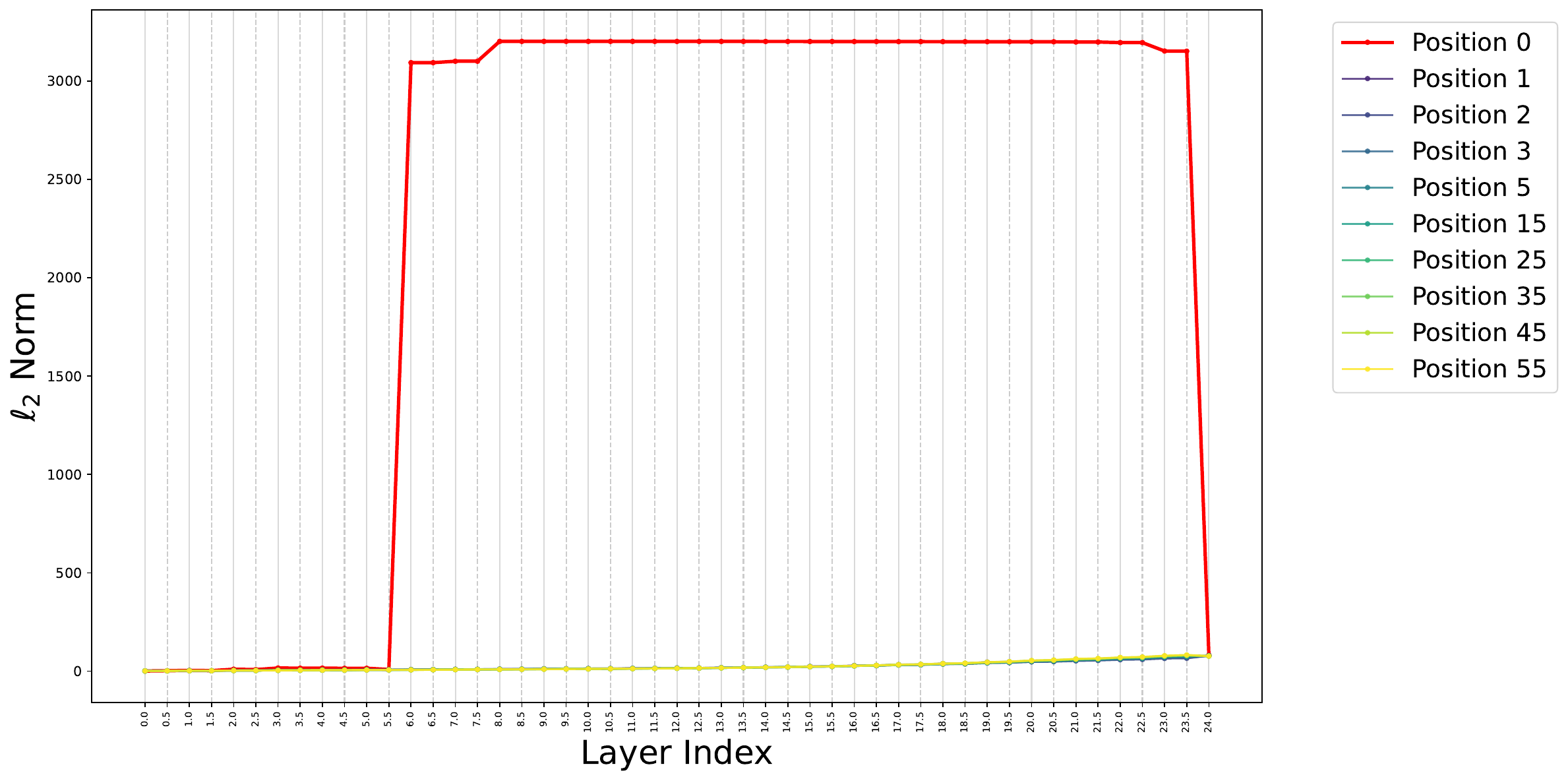}
        \end{subfigure}
    \end{minipage}
    \caption{Layer-wise $\ell_2$ norm of hidden states in transformer-native lm loss model. Half-integer indices correspond to attention module outputs after the residual connection. Metrics are computed over sequences of up to 64 tokens across 1024 validation samples.}
    \label{fig:C-tnlm-copy}
    \vspace{-1em}
\end{figure*}

As shown in Tables~\ref{fig:main-sink-0.3} and~\ref{fig:main-sink-0.4}, the TNLM Loss model does not form a P0 sink until layer 6\footnote{For brevity, only two exponent settings are presented in the table.}, where it emerges with more severely concentrated attention on the P0 token. This behavior closely resembles the sink pattern of Qwen3-4B, as shown in Table~\ref{tab:qwen-sink} and Figure~\ref{fig:appedixD-analysis-Qwen3-4B}. For convenience, we reproduce Figure~\ref{fig:C-tnlm} as Figure~\ref{fig:C-tnlm-copy}.

\begin{table}[ht]
\centering
\resizebox{\linewidth}{!}{
\begin{tabular}{l|c|cccccc}
\toprule
\textbf{Method} & \textbf{LR} & $\text{Sink}_{0}^{\epsilon}[0:]$ & $\text{Sink}_{0}^{\epsilon}[1:]$ & $\text{Sink}_{0}^{\epsilon}[2:]$ & $\text{Sink}_{0}^{\epsilon}[4:]$ & $\text{Sink}_{0}^{\epsilon}[8:]$ & $\text{Sink}_{0}^{\epsilon}[16:]$ \\
\midrule
Baseline*                     & \multirow{5}{*}{\textbf{$3\times10^{-4}$}} & 54.11 & 56.46 & 59.03 & 62.06 & 70.99 & 71.59 \\
Gated Attention              &                                             &  0.51 &  0.53 &  0.51 &  0.56 &  0.63 &  0.68 \\
Force-sink Masking           &                                             & 71.77 & 74.89 & 74.60 & 74.59 & 74.28 & 76.42 \\
TNLM Loss                    &                                             & 50.56 & 52.76 & 55.15 & 60.67 & 69.74 & 71.86 \\
Force-sink Masking+TNLM Loss &                                             & 71.27 & 74.36 & 74.05 & 74.06 & 73.84 & 75.71 \\
\midrule
Baseline*                     & \multirow{5}{*}{\textbf{$4\times10^{-4}$}} & 67.97 & 70.92 & 74.15 & 78.32 & 81.57 & 84.19 \\
Gated Attention              &                                             &  0.85 &  0.88 &  0.86 &  0.93 &  1.05 &  0.96 \\
Force-sink Masking           &                                             & 74.88 & 78.14 & 77.77 & 77.56 & 77.34 & 79.34 \\
TNLM Loss                    &                                             & 54.92 & 57.31 & 59.92 & 65.91 & 75.14 & 77.86 \\
Force-sink Masking+TNLM Loss &                                             & 74.68 & 77.93 & 77.49 & 77.42 & 77.69 & 80.13 \\
\midrule
Baseline*                     & \multirow{5}{*}{\textbf{$5\times10^{-4}$}} & 33.24 & 34.68 & 36.24 & 39.69 & 42.76 & 49.28 \\
Gated Attention              &                                             &  0.97 &  1.01 &  1.01 &  1.09 &  1.27 &  1.59 \\
Force-sink Masking           &                                             & 76.97 & 80.32 & 80.03 & 80.12 & 79.90 & 81.42 \\
TNLM Loss                    &                                             & 57.31 & 59.80 & 62.52 & 68.78 & 77.09 & 79.69 \\
Force-sink Masking+TNLM Loss &                                             & 76.38 & 79.70 & 79.35 & 79.52 & 80.26 & 82.05 \\
\bottomrule
\end{tabular}}
\caption{Sink scores at $\epsilon=0.3$ for the 3 learning rates under Chinchilla-optimal setting.}
\label{fig:main-sink-0.3}
\end{table}

\begin{table}[ht]
\centering
\resizebox{\linewidth}{!}{
\begin{tabular}{l|c|cccccc}
\toprule
\textbf{Method} & \textbf{LR} & $\text{Sink}_{0}^{\epsilon}[0:]$ & $\text{Sink}_{0}^{\epsilon}[1:]$ & $\text{Sink}_{0}^{\epsilon}[2:]$ & $\text{Sink}_{0}^{\epsilon}[4:]$ & $\text{Sink}_{0}^{\epsilon}[8:]$ & $\text{Sink}_{0}^{\epsilon}[16:]$ \\
\midrule
Baseline                      & \multirow{5}{*}{\textbf{$3\times10^{-4}$}} & 22.73 & 23.72 & 24.80 & 26.11 & 31.06 & 33.60 \\
Gated Attention               &                                             &  0.13 &  0.13 &  0.12 &  0.13 &  0.14 &  0.12 \\
Force-sink Masking            &                                             & 61.04 & 63.69 & 62.89 & 62.34 & 60.89 & 63.63 \\
TNLM Loss                     &                                             & 38.75 & 40.43 & 42.27 & 46.50 & 54.39 & 58.60 \\
Force-sink Masking+TNLM Loss  &                                             & 61.67 & 64.35 & 63.59 & 63.31 & 61.05 & 64.19 \\
\midrule
Baseline                      & \multirow{5}{*}{\textbf{$4\times10^{-4}$}} & 42.31 & 44.15 & 46.16 & 48.56 & 51.60 & 56.03 \\
Gated Attention               &                                             &  0.30 &  0.31 &  0.28 &  0.31 &  0.33 &  0.22 \\
Force-sink Masking            &                                             & 65.87 & 68.73 & 67.88 & 67.31 & 66.62 & 69.39 \\
TNLM Loss                     &                                             & 44.50 & 46.44 & 48.55 & 53.40 & 61.34 & 66.39 \\
Force-sink Masking+TNLM Loss  &                                             & 65.58 & 68.43 & 67.71 & 66.94 & 65.71 & 68.08 \\
\midrule
Baseline                      & \multirow{5}{*}{\textbf{$5\times10^{-4}$}} & 11.67 & 12.18 & 12.73 & 13.99 & 15.49 & 18.70 \\
Gated Attention               &                                             &  0.30 &  0.31 &  0.30 &  0.33 &  0.38 &  0.36 \\
Force-sink Masking            &                                             & 68.53 & 71.51 & 70.82 & 70.38 & 70.16 & 72.44 \\
TNLM Loss                     &                                             & 47.90 & 49.99 & 52.26 & 57.49 & 64.41 & 68.42 \\
Force-sink Masking+TNLM Loss  &                                             & 69.15 & 72.16 & 71.58 & 71.36 & 70.26 & 72.45 \\
\bottomrule
\end{tabular}}
\caption{Sink scores at $\epsilon=0.4$ for the 3 learning rates under Chinchilla-optimal setting.}
\label{fig:main-sink-0.4}
\end{table}

\begin{table}[ht]
\centering
\resizebox{\linewidth}{!}{
\begin{tabular}{l|c|cccccc}
\toprule
\textbf{Model} & \textbf{$\epsilon$} & $\text{Sink}_{0}^{\epsilon}[0:]$ & $\text{Sink}_{0}^{\epsilon}[1:]$ & $\text{Sink}_{0}^{\epsilon}[2:]$ & $\text{Sink}_{0}^{\epsilon}[4:]$ & $\text{Sink}_{0}^{\epsilon}[8:]$ & $\text{Sink}_{0}^{\epsilon}[16:]$ \\
\midrule
Qwen3-4B & $0.3$ & 68.47 & 70.43 & 72.50 & 76.92 & 84.54 & 84.66 \\
Qwen3-4B & $0.4$ & 59.51 & 61.21 & 63.01 & 66.92 & 73.37 & 75.40 \\
\bottomrule
\end{tabular}}
\caption{Sink scores for Qwen3-4B at $\epsilon \in \{0.3, 0.4\}$.}
\label{tab:qwen-sink}
\end{table}

\FloatBarrier

\section{Data-saturated Experiment}
\label{appendix:5xchinchilla}

As shown in Tables~\ref{tab:performance-bos-vs-nobos-pretrain} and~\ref{tab:sink-rate-bos-vs-non-bos}, baseline models in the Chinchilla-optimal setting have not yet completed P0 sink formation, making comparisons against methods that accelerate this process somewhat unfair, as the baseline still suffers from instability in the attention QK inner-product space. We therefore run data-saturated experiments at $5\times$ the Chinchilla-optimal budget. Under this setting, Gated Attention shows no advantage over the standard Transformer with an equivalent number of parameters added to the MLP. In contrast, TNLM Loss and Force-sink Masking$^+$ ($k=32$ Force-sink dimensions, Figure~\ref{fig:fsm}) consistently outperform the baseline with fewer FLOPs.

\begin{table}[ht]
    \centering
    \resizebox{\linewidth}{!}{
    \begin{tabular}{c|l|c|l|lllll|l}
    \toprule
    \textbf{Step} & \textbf{Method} & \textbf{Sink Rate} & \textbf{Lambada} & \textbf{PIQA} & \textbf{HellaSwag} & \textbf{WinoGrande} & \textbf{ARC\_e} & \textbf{ARC\_c} & \textbf{Avg.} \\
    \midrule
    \multirow{7}{*}{200704}
        & \textbf{Baseline} & 85.53 & 49.25 & 74.21 & 60.10 & 59.27 & \textbf{74.03} & \textbf{41.89} & 61.90 \\
        & \textbf{Gated Attention} & 2.33 & 51.31 (\textcolor{red}{+2.06}) & \textbf{75.03} (\textcolor{red}{+0.82}) & \textbf{60.69} (\textcolor{red}{+0.59}) & 60.93 (\textcolor{red}{+1.66}) & 71.25 (\textcolor{blue}{-2.78}) & 40.36 (\textcolor{blue}{-1.53}) & 61.65 (\textcolor{blue}{-0.25}) \\
        & \textbf{Force-sink Masking} & 92.65 & 51.56 (\textcolor{red}{+2.31}) & 73.94 (\textcolor{blue}{-0.27}) & 60.11 (\textcolor{red}{+0.01}) & 59.75 (\textcolor{red}{+0.48}) & 73.11 (\textcolor{blue}{-0.92}) & 40.02 (\textcolor{blue}{-1.87}) & 61.39 (\textcolor{blue}{-0.51}) \\
        & \textbf{Force-sink Masking$^+$} & 90.04 & 50.73 (\textcolor{red}{+1.48}) & 74.21 (\textcolor{red}{+0.00}) & 59.94 (\textcolor{blue}{-0.16}) & \textbf{61.01} (\textcolor{red}{+1.74}) & 73.61 (\textcolor{blue}{-0.42}) & 40.78 (\textcolor{blue}{-1.11}) & 61.91 (\textcolor{red}{+0.01}) \\
        & \textbf{TNLM Loss} & 72.27 & 51.10 (\textcolor{red}{+1.85}) & 74.86 (\textcolor{red}{+0.65}) & 60.26 (\textcolor{red}{+0.16}) & 60.30 (\textcolor{red}{+1.03}) & 73.53 (\textcolor{blue}{-0.50}) & 41.72 (\textcolor{blue}{-0.17}) & \textbf{62.13} (\textcolor{red}{+0.23}) \\
        & \textbf{Force-sink Masking+TNLM Loss} & 92.97 & 51.14 (\textcolor{red}{+1.89}) & 73.45 (\textcolor{blue}{-0.76}) & 60.34 (\textcolor{red}{+0.24}) & 59.12 (\textcolor{blue}{-0.15}) & 72.10 (\textcolor{blue}{-1.93}) & 39.51 (\textcolor{blue}{-2.38}) & 60.90 (\textcolor{blue}{-1.00}) \\
        & \textbf{Force-sink Masking$^+$+TNLM Loss} & 92.55 & \textbf{52.55} (\textcolor{red}{+3.30}) & 74.81 (\textcolor{red}{+0.60}) & 60.57 (\textcolor{red}{+0.47}) & 58.96 (\textcolor{blue}{-0.31}) & 73.36 (\textcolor{blue}{-0.67}) & 40.27 (\textcolor{blue}{-1.62}) & 61.59 (\textcolor{blue}{-0.31}) \\
    \midrule
    \multirow{7}{*}{202752}
        & \textbf{Baseline} & 85.61 & 48.11 & 74.10 & 59.70 & 59.12 & \textbf{73.36} & \textbf{40.87} & 61.43 \\
        & \textbf{Gated Attention} & 2.44 & \textbf{51.43} (\textcolor{red}{+3.32}) & \textbf{75.52} (\textcolor{red}{+1.42}) & \textbf{60.74} (\textcolor{red}{+1.04}) & \textbf{60.69} (\textcolor{red}{+1.57}) & 71.72 (\textcolor{blue}{-1.64}) & 39.42 (\textcolor{blue}{-1.45}) & \textbf{61.62} (\textcolor{red}{+0.19}) \\
        & \textbf{Force-sink Masking} & 92.65 & 49.23 (\textcolor{red}{+1.12}) & 73.56 (\textcolor{blue}{-0.54}) & 60.26 (\textcolor{red}{+0.56}) & 60.22 (\textcolor{red}{+1.10}) & 73.11 (\textcolor{blue}{-0.25}) & 39.85 (\textcolor{blue}{-1.02}) & 61.40 (\textcolor{blue}{-0.03}) \\
        & \textbf{Force-sink Masking$^+$} & 90.16 & 50.48 (\textcolor{red}{+2.37}) & 74.16 (\textcolor{red}{+0.06}) & 60.23 (\textcolor{red}{+0.53}) & 59.19 (\textcolor{red}{+0.07}) & 72.43 (\textcolor{blue}{-0.93}) & 40.78 (\textcolor{blue}{-0.09}) & 61.36 (\textcolor{blue}{-0.07}) \\
        & \textbf{TNLM Loss} & 72.30 & 50.94 (\textcolor{red}{+2.83}) & 74.37 (\textcolor{red}{+0.27}) & 60.15 (\textcolor{red}{+0.45}) & 60.30 (\textcolor{red}{+1.18}) & 73.02 (\textcolor{blue}{-0.34}) & 40.27 (\textcolor{blue}{-0.60}) & \textbf{61.62} (\textcolor{red}{+0.19}) \\
        & \textbf{Force-sink Masking+TNLM Loss} & 92.99 & 49.91 (\textcolor{red}{+1.80}) & 73.12 (\textcolor{blue}{-0.98}) & 60.27 (\textcolor{red}{+0.57}) & 60.62 (\textcolor{red}{+1.50}) & 71.25 (\textcolor{blue}{-2.11}) & 39.59 (\textcolor{blue}{-1.28}) & 60.97 (\textcolor{blue}{-0.46}) \\
        & \textbf{Force-sink Masking$^+$+TNLM Loss} & 92.59 & 50.94 (\textcolor{red}{+2.83}) & 74.48 (\textcolor{red}{+0.38}) & 60.49 (\textcolor{red}{+0.79}) & 59.19 (\textcolor{red}{+0.07}) & 73.11 (\textcolor{blue}{-0.25}) & 40.27 (\textcolor{blue}{-0.60}) & 61.51 (\textcolor{red}{+0.08}) \\
    \midrule
    \multirow{7}{*}{204800}
        & \textbf{Baseline} & 85.57 & 49.37 & 73.88 & 59.99 & 59.75 & \textbf{73.48} & 40.78 & 61.58 \\
        & \textbf{Gated Attention} & 2.37 & \textbf{51.64} (\textcolor{red}{+2.27}) & \textbf{74.97} (\textcolor{red}{+1.09}) & \textbf{60.74} (\textcolor{red}{+0.75}) & \textbf{59.83} (\textcolor{red}{+0.08}) & 72.01 (\textcolor{blue}{-1.47}) & 40.27 (\textcolor{blue}{-0.51}) & 61.56 (\textcolor{blue}{-0.01}) \\
        & \textbf{Force-sink Masking} & 92.65 & 50.38 (\textcolor{red}{+1.01}) & 73.56 (\textcolor{blue}{-0.32}) & 60.13 (\textcolor{red}{+0.14}) & 59.27 (\textcolor{blue}{-0.48}) & 72.47 (\textcolor{blue}{-1.01}) & 40.19 (\textcolor{blue}{-0.59}) & 61.12 (\textcolor{blue}{-0.45}) \\
        & \textbf{Force-sink Masking$^+$} & 90.10 & 49.74 (\textcolor{red}{+0.37}) & 74.86 (\textcolor{red}{+0.98}) & 60.40 (\textcolor{red}{+0.41}) & 59.12 (\textcolor{blue}{-0.63}) & 73.06 (\textcolor{blue}{-0.42}) & \textbf{42.06} (\textcolor{red}{+1.28}) & \textbf{61.90} (\textcolor{red}{+0.33}) \\
        & \textbf{TNLM Loss} & 72.29 & 51.10 (\textcolor{red}{+1.73}) & 74.05 (\textcolor{red}{+0.17}) & 60.46 (\textcolor{red}{+0.47}) & 59.35 (\textcolor{blue}{-0.40}) & 72.69 (\textcolor{blue}{-0.79}) & 39.33 (\textcolor{blue}{-1.45}) & 61.18 (\textcolor{blue}{-0.40}) \\
        & \textbf{Force-sink Masking+TNLM Loss} & 93.00 & 50.34 (\textcolor{red}{+0.97}) & 73.45 (\textcolor{blue}{-0.43}) & 60.48 (\textcolor{red}{+0.49}) & \textbf{59.83} (\textcolor{red}{+0.08}) & 71.80 (\textcolor{blue}{-1.68}) & 40.27 (\textcolor{blue}{-0.51}) & 61.17 (\textcolor{blue}{-0.41}) \\
        & \textbf{Force-sink Masking$^+$+TNLM Loss} & 92.59 & 51.48 (\textcolor{red}{+2.11}) & 74.43 (\textcolor{red}{+0.55}) & 60.58 (\textcolor{red}{+0.59}) & 59.51 (\textcolor{blue}{-0.24}) & 72.64 (\textcolor{blue}{-0.84}) & 40.19 (\textcolor{blue}{-0.59}) & 61.47 (\textcolor{blue}{-0.11}) \\
    \bottomrule
    \end{tabular}}
    \caption{Downstream Performance for the last 3 checkpoints under 5$\times$ Chinchilla-optimal setting.}
    \label{tab:placeholder}
\end{table}

\begin{table}[ht]
\centering
\resizebox{\linewidth}{!}{
\begin{tabular}{c|l|cccccc}
\toprule
\textbf{Step} & \textbf{Method} & $\text{Sink}_{0}^{\epsilon}[0:]$ & $\text{Sink}_{0}^{\epsilon}[1:]$ & $\text{Sink}_{0}^{\epsilon}[2:]$ & $\text{Sink}_{0}^{\epsilon}[4:]$ & $\text{Sink}_{0}^{\epsilon}[8:]$ & $\text{Sink}_{0}^{\epsilon}[16:]$ \\
\midrule
\multirow{7}{*}{200704} & Baseline                     & 85.53 & 89.25 & 93.31 & 97.64 & 97.36 & 97.72 \\
                                  & Gated Attention              &  2.33 &  2.39 &  2.42 &  2.66 &  3.15 &  4.76 \\
                                  & Force-sink Masking           & 92.65 & 96.68 & 96.53 & 96.19 & 95.49 & 94.97 \\
                                  & Force-sink Masking$^+$       & 92.35 & 96.37 & 96.20 & 95.92 & 95.41 & 94.64 \\
                                  & TNLM Loss                    & 72.27 & 75.41 & 78.84 & 86.72 & 96.26 & 94.73 \\
                                  & Force-sink Masking+TNLM Loss & 92.97 & 97.01 & 96.88 & 96.56 & 95.72 & 95.50 \\
                                  & Force-sink Masking$^+$+TNLM Loss & 92.55 & 96.57 & 96.42 & 96.06 & 95.40 & 95.05 \\
\midrule
\multirow{7}{*}{202752} & Baseline                     & 85.61 & 89.33 & 93.39 & 97.73 & 97.45 & 97.78 \\
                                  & Gated Attention              &  2.44 &  2.50 &  2.54 &  2.80 &  3.32 &  5.03 \\
                                  & Force-sink Masking           & 92.65 & 96.68 & 96.53 & 96.18 & 95.49 & 94.95 \\
                                  & Force-sink Masking$^+$       & 92.43 & 96.45 & 96.29 & 96.02 & 95.52 & 94.78 \\
                                  & TNLM Loss                    & 72.30 & 75.45 & 78.88 & 86.76 & 96.30 & 94.78 \\
                                  & Force-sink Masking+TNLM Loss & 92.99 & 97.03 & 96.89 & 96.58 & 95.75 & 95.57 \\
                                  & Force-sink Masking$^+$+TNLM Loss & 92.59 & 96.61 & 96.46 & 96.10 & 95.46 & 95.14 \\
\midrule
\multirow{7}{*}{204800} & Baseline                     & 85.57 & 89.29 & 93.35 & 97.68 & 97.42 & 97.73 \\
                                  & Gated Attention              &  2.37 &  2.42 &  2.46 &  2.70 &  3.22 &  4.78 \\
                                  & Force-sink Masking           & 92.67 & 96.70 & 96.55 & 96.21 & 95.53 & 95.05 \\
                                  & Force-sink Masking$^+$       & 92.40 & 96.42 & 96.25 & 95.97 & 95.47 & 94.70 \\
                                  & TNLM Loss                    & 72.29 & 75.44 & 78.86 & 86.74 & 96.26 & 94.72 \\
                                  & Force-sink Masking+TNLM Loss & 93.00 & 97.04 & 96.90 & 96.60 & 95.76 & 95.60 \\
                                  & Force-sink Masking$^+$+TNLM Loss & 92.59 & 96.62 & 96.47 & 96.11 & 95.44 & 95.11 \\
\bottomrule
\end{tabular}}
\caption{Sink scores at $\epsilon=0.3$ for the last 3 checkpoints under 5$\times$ Chinchilla-optimal setting.}
\end{table}

\begin{table}[ht]
\centering
\resizebox{\linewidth}{!}{
\begin{tabular}{c|l|cccccc}
\toprule
\textbf{Step} & \textbf{Method} & $\text{Sink}_{0}^{\epsilon}[0:]$ & $\text{Sink}_{0}^{\epsilon}[1:]$ & $\text{Sink}_{0}^{\epsilon}[2:]$ & $\text{Sink}_{0}^{\epsilon}[4:]$ & $\text{Sink}_{0}^{\epsilon}[8:]$ & $\text{Sink}_{0}^{\epsilon}[16:]$ \\
\midrule
\multirow{7}{*}{200704} & Baseline                     & 81.22 & 84.75 & 88.60 & 92.46 & 91.97 & 93.89 \\
                                  & Gated Attention              &  0.79 &  0.81 &  0.80 &  0.88 &  1.01 &  1.34 \\
                                  & Force-sink Masking           & 90.42 & 94.35 & 94.09 & 93.50 & 92.31 & 91.94 \\
                                  & Force-sink Masking$^+$       & 90.04 & 93.95 & 93.68 & 93.18 & 92.08 & 91.47 \\
                                  & TNLM Loss                    & 69.20 & 72.21 & 75.49 & 83.04 & 91.89 & 90.34 \\
                                  & Force-sink Masking+TNLM Loss & 90.72 & 94.66 & 94.42 & 93.86 & 92.57 & 92.36 \\
                                  & Force-sink Masking$^+$+TNLM Loss & 89.83 & 93.74 & 93.45 & 92.80 & 91.87 & 92.09 \\
\midrule
\multirow{7}{*}{202752} & Baseline                     & 81.53 & 85.07 & 88.94 & 92.83 & 92.36 & 94.30 \\
                                  & Gated Attention              &  0.85 &  0.87 &  0.88 &  0.97 &  1.11 &  1.51 \\
                                  & Force-sink Masking           & 90.45 & 94.39 & 94.13 & 93.54 & 92.37 & 92.10 \\
                                  & Force-sink Masking$^+$       & 90.16 & 94.08 & 93.81 & 93.32 & 92.25 & 91.73 \\
                                  & TNLM Loss                    & 69.29 & 72.30 & 75.59 & 83.15 & 92.05 & 90.44 \\
                                  & Force-sink Masking+TNLM Loss & 90.76 & 94.71 & 94.47 & 93.91 & 92.65 & 92.54 \\
                                  & Force-sink Masking$^+$+TNLM Loss & 89.87 & 93.78 & 93.50 & 92.85 & 91.97 & 92.31 \\
\midrule
\multirow{7}{*}{204800} & Baseline                     & 81.34 & 84.88 & 88.74 & 92.61 & 92.15 & 94.03 \\
                                  & Gated Attention              &  0.80 &  0.82 &  0.82 &  0.91 &  1.04 &  1.34 \\
                                  & Force-sink Masking           & 90.44 & 94.37 & 94.12 & 93.53 & 92.35 & 92.01 \\
                                  & Force-sink Masking$^+$       & 90.10 & 94.02 & 93.75 & 93.25 & 92.17 & 91.59 \\
                                  & TNLM Loss                    & 69.21 & 72.22 & 75.50 & 83.05 & 91.97 & 90.25 \\
                                  & Force-sink Masking+TNLM Loss & 90.75 & 94.69 & 94.45 & 93.90 & 92.63 & 92.44 \\
                                  & Force-sink Masking$^+$+TNLM Loss & 89.89 & 93.80 & 93.52 & 92.87 & 91.98 & 92.27 \\
\bottomrule
\end{tabular}}
\caption{Sink scores at $\epsilon=0.4$ for the last 3 checkpoints under 5$\times$ Chinchilla-optimal setting.}
\end{table}

\end{document}